\pdfoutput=1

\documentclass[11pt]{article}

\usepackage{EMNLP2022}

\usepackage{times}
\usepackage{latexsym}

\usepackage[T1]{fontenc}

\usepackage[utf8]{inputenc}

\usepackage{microtype}

\usepackage{inconsolata}
\usepackage{multirow}
\usepackage{bm}

\usepackage{comment}
\includecomment{skeleton}
\excludecomment{braindump}
\includecomment{notarxiv}
\excludecomment{arxiv}
\excludecomment{cutout}
\excludecomment{toappendix}

\usepackage{booktabs,hhline,array}
\usepackage{amsmath}
\newcommand\norm[1]{\lVert#1\rVert}

\usepackage{graphicx}
\usepackage{wrapfig}
\usepackage{caption}
\usepackage{subcaption}

\graphicspath{ {Figures/} }

\title{Train Flat, Then Compress: \\Sharpness-Aware Minimization Learns More Compressible Models}

\author{Clara Na \quad Sanket Vaibhav Mehta \quad Emma Strubell \\ 
Language Technologies Institute \\
School of Computer Science \\
Carnegie Mellon University \\
\texttt{\{csna, svmehta, estrubel\}@cs.cmu.edu}}

\begin{document}
\maketitle
\begin{abstract}

Model compression by way of parameter pruning, quantization, or distillation has recently gained popularity as an approach for reducing the computational requirements of modern deep neural network models for NLP. Inspired by prior works suggesting a connection between simpler, more generalizable models and those that lie within wider loss basins, we hypothesize that optimizing for flat minima should lead to simpler parameterizations and thus more compressible models. We propose to combine sharpness-aware minimization (SAM) with various task-specific model compression methods, including iterative magnitude pruning (IMP), structured pruning with a distillation objective, and post-training dynamic quantization. Empirically, we show that optimizing for flatter minima consistently leads to greater compressibility of parameters compared to vanilla Adam when fine-tuning BERT models, with little to no loss in accuracy on the GLUE text classification and SQuAD question answering benchmarks. Moreover, SAM \textit{finds superior winning tickets} during IMP that 1) are amenable to vanilla Adam optimization, and 2) transfer more effectively across tasks.\footnote{Code is available at \url{https://github.com/clarana/train-flat-compress}}
\end{abstract}

\section{Introduction}
\label{sec:introduction}

\begin{figure}[ht!]
    \centering
    \includegraphics[scale=0.5]{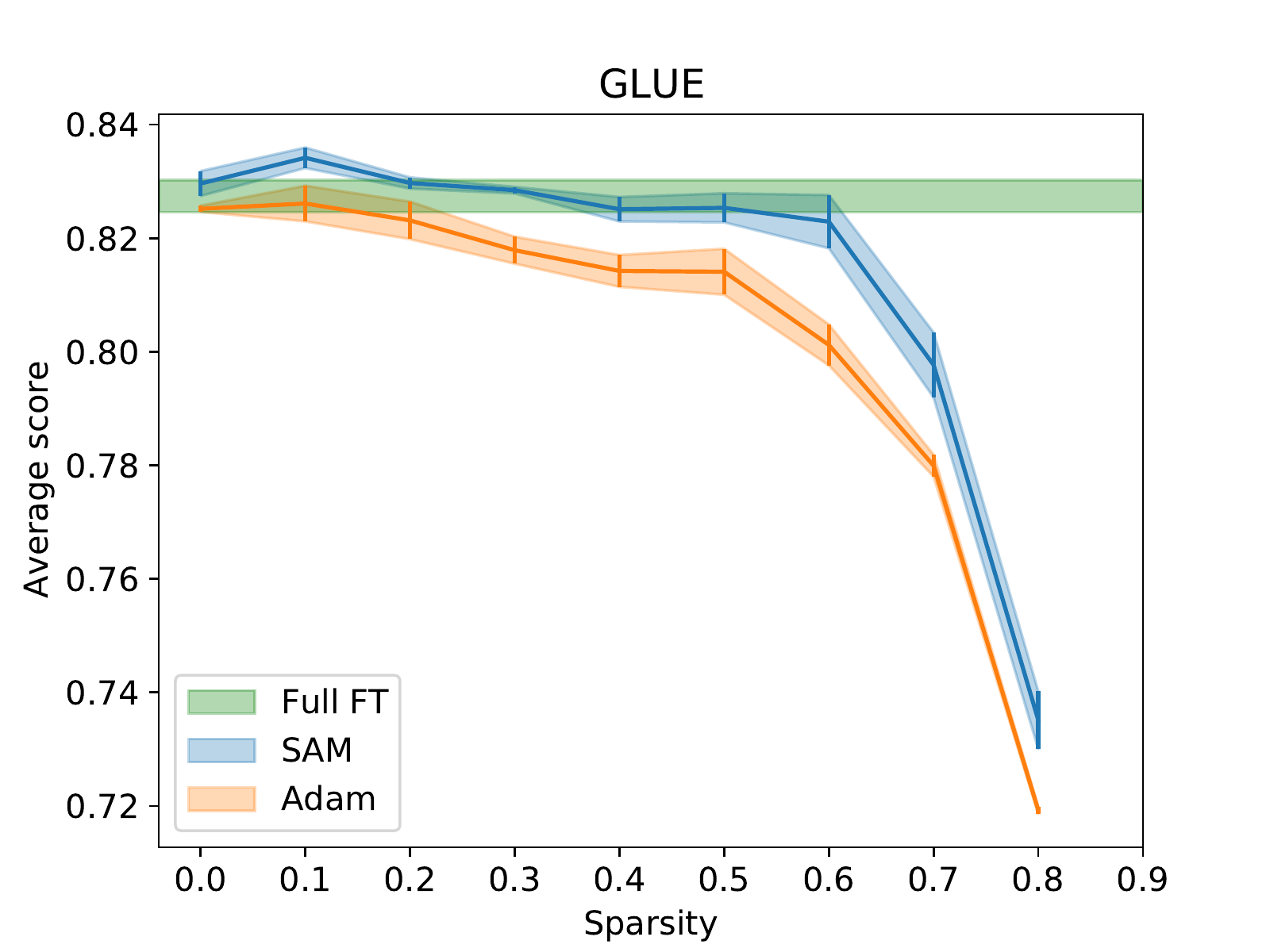}
    \caption{Average score over all GLUE tasks as a function of sparsity (\% of parameters pruned) of BERT$_{base}$ through unstructured iterative magnitude pruning. We compare the baseline Adam optimized model's performance to our model, optimized to prefer flat minima via SAM. The green horizontal bands mark initial performance of our full fine-tuned (FT) models. SAM outperforms baseline Adam across all sparsity values. }
    \label{fig:1}
\end{figure}

Recent advances in hardware, modeling, and optimization for deep neural networks have enabled training of substantially larger models on massive amounts of unlabeled data, leading to corresponding improvements in accuracy across a variety of tasks in NLP \citep{devlin-etal-2019-bert, brown2020gpt3, JMLR:t5raffel}. Unfortunately, this sudden increase in scale of state-of-the-art models also has adverse consequences, such as reducing equity of access %
\citep{yu2020algorithmic,nur2020dedemocratization} and increasing computational and energy requirements \citep{strubell-etal-2019-energy, dodge2022measuring}. 

In response, %
\textit{model compression} has emerged as a dominant approach to improving memory and inference efficiency in neural network models, including approaches such as knowledge distillation \citep{bucilua2006compression, hinton2014distilling, jiao-etal-2020-tinybert}, model quantization \citep{vanhoucke2011improving, shen2020qbert} and pruning \citep{lecun1989optimal, chen2020BertLT, xia2022structuredCoFi}. 

The vast majority of work on model compression focuses on methods for selecting how and where to reduce (via quantization or distillation) or remove (by pruning) model parameters without sacrificing end-task accuracy. While these approaches are usually simple to implement and work relatively well for maintaining overall accuracy on considered benchmarks, recent work has revealed that these commonly used compression methods can result in negative impacts on model behavior that are not necessarily captured by current performance metrics. For example, in the field of computer vision, pruning has been shown to sacrifice performance on long-tail examples in order to preserve accuracy on more frequent phenomena \citep{hooker2020characterising} and reduce robustness to out-of-distribution examples \citep{liebenwein2021lost}. At the same time, it has also been shown that compression can sometimes have a regularizing effect, improving generalization in some settings \citep{ahia-etal-2021-low-resource}. Clearly, more work is needed better understand the relationship between compression and generalizability, and to develop improved compression methods that are informed by this knowledge.

Meanwhile, there is a growing body of work investigating curvature of the loss landscape and its relation to generalization in deep neural models. \citet{hochreiter1997flat} first defined \textit{flat minima} as regions in parameter space where error remains relatively stable despite perturbations in parameter values, arguing that models in flat minima should correspond to simpler, more generalizable functions. More recently, \citet{wu2017understanding} further showed that wide loss basins correspond to low-complexity solutions,\footnote{\protect{\citet{wu2017understanding} demonstrate this theoretically for two layer feed-forward networks, and empirically for larger networks applied to computer vision.}} and it has been shown empirically that directly optimizing for solutions in flat minima leads to improved generalization on a wide range of supervised learning tasks in both vision \citep{foret2021sharpnessaware} and language \citep{bahri22acl} modalities. 

Inspired by the above results connecting wider loss regions to simpler, more generalizable models, in this work we aim to advance our understanding of model compression by examining the relationship between flat minima and compressibility in fine-tuned language models. Intuitively, model parameters in neighborhoods having uniformly low loss values should be more robust to perturbations, such as those resulting from model compression, compared to models in sharper regions, since changes to parameter values should lead to minimal change in loss with respect to the main training objective. Empirically and theoretically, previous work has also linked generalizability with compressibility, finding that neural networks whose weights reflect simpler, more general hypotheses may be more robust to compression, and compression itself can act as a regularization mechanism \citep{zhou2018nonvacuous, liang-etal-2021-super, kuhn2021robustnessGeneralizationPruning}. \citet{li2020train} showed that larger pre-trained language models, which are known to genereralize better, are also more compressible. Further,  \citet{bartoldson2020generalizationCompression} connect flatness to generalization in pruned models, showing that pruning noise correlates positively with measures of flatness in a CNN for computer vision.

We investigate the relationship between flat minima and compression in large pre-trained language models by directly optimizing for flat minima during language model fine-tuning using \textit{sharpness-aware minimization} (SAM; \citet{foret2021sharpnessaware}). %
Through extensive experiments on the GLUE text classification %
and SQuAD question answering benchmarks, we show that fine-tuning BERT models with SAM leads to optima in flatter basins, and compressing those models consistently results in higher model accuracy at the same level of compression compared to standard Adam-optimized baselines. Our results hold across multiple BERT variants and sizes \cite{devlin-etal-2019-bert, Liu2019RoBERTaAR} and a wide variety of compression methods: Iterative magnitude pruning with and without rewinding, a structured pruning procedure that also employs knowledge distillation, %
and an off-the-shelf method for post-training quantization. We also show that sparse subnetworks (\textit{winning tickets}; \citet{frankle2018lotteryticket}) discovered by SAM are better transferable across tasks%
, suggesting improved generalizability. Our findings shed light on a promising new avenue for obtaining practical improvements in model compression.%

\begin{cutout}
Distillation is a popular approach that has provided an effective tradeoff between accuracy and inference efficiency. Quantization of floating point parameters below half precision (16 bits) can provide impressive compression rates and modest inference improvements that can be amplified with specialized hardware and instruction sets. Model pruning is a promising approach that can be compatible with a wide variety of hardware without substantial specialized training in addition to typical fine-tuning. Approaches for \textit{structured pruning}, where entire parameter matrices are pruned at once in order for sparse networks to run efficiently on hardware designed to accelerate large tensor operations, can provide superior inference efficiency compared to unstructured pruning \citep{lagunas-etal-2021-block, xia2022structuredCoFi}, and have recently been shown to out-perform distillation alone in terms of accuracy-efficiency trade-off for task-specific fine-tuning \citep{xia2022structuredCoFi}. Pruning is also a theoretically attractive approach in light of recent work suggesting that large, dense neural networks naturally contain sparse subnetworks that can be trained to match the accuracy of the full model \citep{frankle2018lotteryticket, chen2020BertLT}.

Approaches for model pruning typically take a large, accurate model as input, then incrementally remove increasingly more parameters until meeting a desired network sparsity or end-task accuracy. Methods vary primarily in the way that they identify which parameters to remove and when. In \textit{iterative magnitude pruning}, a percent of parameters with magnitude closest to zero are incrementally removed, then the model re-trained with the new substructure \citep{han2015learning}. Other methods such as \textit{movement pruning} \citep{sanh2020movement} remove parameters according to how much they grow in magnitude throughout training, and earlier methods have used higher-order statistics of the gradient to serve a similar purpose \citep{lecun1989optimal}. %
\end{cutout}

\begin{braindump}
[Why model compression? save storage space, potentially faster inference, can use on device if small enough, "train large then compress" in general for good generalization]

[introduce distinct "goals" we have with compression: 1) Lottery ticket-style NAS (Can SAM help us find sparse subnetworks more easily?), 2) useful training and inference speedups (Can SAM help models converge to more , 3) ]

[]

(Varying degrees of real-world usefulness, and exploring the different effects of SAM on them hints at different mechanisms / identify different points in a training pipeline where sharpness "matters" (and how much it matters))

[why do we think sam might help us get more compressible models? below are just bullet points ! not a coherent narrative, will try to figure out how to organize and frame later]
\begin{itemize}
    \item Simplest: there may be a link between compressibility and generalization in general. (and there's a link between flatness and generalization). If SAM helps us find models with good generalization performance, does it also help us find good compressible models? (Can we connect the third side of the flatness/generalization/compressibility triangle)
    \item idea (cite probably the old flatness paper): sam -> flatness -> simpler hypotheses -> more robust to compression because it's \textit{extra} over-parameterized
    \item train large then compress – for good generalization performance, train a large model (for fewer steps) and then compress it. These larger models tend to be more robust to compression techniques (quantization, pruning) than smaller models, so much so that large models retain superior performance when trained and then compressed to match the size of full-size smaller models. (cite train large then compress paper + also tinybert probably)
    \item $\uparrow$ they don't propose a specific hypothesis for why exactly it works this way, but they do point to the lottery ticket hypothesis paper and claim that "for certain accuracies, as models become increasingly large, they contain increasingly small subnetworks which achieve that accuracy."
    \begin{itemize}
        \item this can be our justification for the lottery ticket style pruning
    \end{itemize}
    \item Practical side: do we "need" to start with much larger models (which may not be as feasible to train in the first place given limited compute etc.)? (don't want to speculate too much and open ourselves up to being asked for more experiments right away lol but) If we can squeeze better compressibility out of a "normal" sized model just by pushing it into a flatter place in its loss landscape, as an alternative to having to start with a very large model, that would be kind of a big deal
    
\end{itemize}

our contributions : (list)

\begin{skeleton}
\begin{enumerate}
    \item 
\end{enumerate}
\end{skeleton}

\end{braindump}

\section{Methods}
\label{sec:preliminaries}
Broadly, we are interested in understanding: 1) Are models in flatter minima more compressible? 2) If so, why? 3) Beyond task-specific accuracy, what properties do flat, compressed models have? 

To provide empirical answers to these questions, our experimental setup is as follows. We fine-tune pre-trained language models on standard benchmarks using both ``vanilla'' Adam optimization and Sharpness-Aware Minimization (\S\ref{sec:sam}). We then experiment with a variety of strategies for compressing those fine-tuned models: iterative magnitude pruning (unstructured) with and without rewinding (\S\ref{sec:imp}), structured pruning using $\ell_0$ regularization and a distillation obejctive (\S\ref{sec:structured}), and post-training dynamic quantization (\S\ref{sec:ptq}). We evaluate model end-task accuracy at different compression rates, and compare that accuracy to the full (uncompressed) model and across experimental settings, such as varying the pre-trained language model, and transferring initializations across tasks. %

\subsection{Flat Minima}
\label{sec:sam}

\paragraph{Sharpness-Aware Minimization (SAM).}
To explicitly encourage flatter loss basins, we employ the SAM \citep{foret2021sharpnessaware} procedure. %
Given a loss function $f(w)$, SAM strives to find parameters that lie in the neighborhood with uniformly low loss by optimizing the following minimax objective:
\begin{align}
    \min_w \max_{||\epsilon||_2 \leq \rho} f(w+\epsilon) \label{eq:sam_minimax}
\end{align}
where the maximization (or neighborhood) region is an $\ell^p$ ball with radius $\rho$ for $p=2$ in Equation 
\eqref{eq:sam_minimax}. 
The gradient of the result of the above (inner) maximization problem can be approximated as:
\begin{align}
\small
    \approx \nabla_w f(w) \Big\rvert_{w+\mathbf{\hat{\epsilon}(w)}} + \frac{\partial\mathbf{\hat{\epsilon}(w)}}{w} \nabla_w f(w) \Big\rvert_{w+\mathbf{\hat{\epsilon}(w)}} \nonumber \\
    \text{where, } \mathbf{\hat{\epsilon}(w)} = \rho \nabla_w f(w)/\norm{\nabla_w f(w)}_2 \nonumber
\end{align}
\citet{foret2021sharpnessaware} showcase that one can simplify the optimization problem without compromising the algorithm's effectiveness by dropping the second order term in the gradient, leaving us with:
\begin{align}
    \nabla_w \max_{||\epsilon||_2 \leq \rho} f(w+\epsilon) \approx \nabla_w f(w) \Big\rvert_{w+\mathbf{\hat{\epsilon}(w)}} 
\end{align}
Intuitively, SAM takes a gradient step at each iteration based on the gradient estimated at the parameters yielding the highest loss ($w + \hat{\epsilon}(w)$) in an $\ell^p$ neighborhood around the current parameters ($w$).

\paragraph{Stochastic Weight Averaging (SWA).}
Although, we use SAM to optimize for flatness, there exist other alternatives to promote flatness like Stochastic Weight Averaging \citep{izmailov2018averaging}. SWA performs an equal average of model checkpoints along the optimization trajectory to find flatter solutions compared to vanilla optimization. We also consider SWA for our experimentation (\S\ref{sec:swa_expts}).

\paragraph{Sharpness Metric.} 
To verify that SAM and SWA indeed leads to flatter minima as compared to Adam, we compute a \textit{sharpness metric} \citep{keskar2016large}, which estimates the flatness by computing the maximum value of $f(w)$ within a neighborhood region controlled by the hyperparameter $\epsilon$. Following \citep{keskar2016large, mehta2021empirical}, the neighborhood region is defined as:
\begin{align}
C_\epsilon = &\{z \in R^p : -\epsilon(|(A^+w)_i| + 1) \leq \nonumber \\
&z_i \leq \epsilon(|(A^+w)_i| + 1) \forall i \in \{1\dots p\}\} 
\label{eq:sharpness_bounds}
\end{align}
where $R^p$ is a random subspace of the entire parameter space $R^n$ constructed using a projection matrix $A \in R^{n \times p}$, and $A^+$ is the pseudo inverse of $A$. Concretely, the sharpness metric (lower corresponds to flatter minima) is computed as follows:
\begin{align}
\phi_{w,f} := \frac{\max_{z\in C_\epsilon} f(w + Az) - f(w)}{1 + f(w)} \times 100 \label{eq:sharpness_value}
\end{align}
To qualitatively verify for flatness, we also visualize loss contours (see Appendix \ref{sec:sharpnessmetric}).

\subsection{Pruning}
\label{sec:pruning}

We investigate compressibility primarily in an unstructured pruning setting. Given a network $\mathcal{N}$ and weights $\mathbf{w}$, we wish to prune a subset of individual weights to leave only a subset, $\mathbf{w'}$. A successfully pruned model has $|\mathbf{w'}| \ll |\mathbf{w}|$ while retaining good performance on the task(s) of interest.

\subsubsection{Iterative Magnitude Pruning}
\label{sec:imp}

Typically, $\mathbf{w'}$ is found through an iterative process where $\mathcal{N}$ is trained and pruned incrementally, either until some target sparsity is reached or until some larger-than-desired performance drop is observed, and a common criterion selects weights with the smallest absolute magnitude to be pruned at each iteration. In the standard pruning scenario \citep{Renda2020ComparingStdPruning2rewinding, han2015StdPruning1}, training simply resumes with the remaining weights after each iteration of pruning. Previous work \citep{Renda2020ComparingStdPruning2rewinding} presents evidence that rewinding remaining weights to earlier learned values may be beneficial for compressibility. 

\citet{frankle2018lotteryticket} present a formulation of iterative magnitude pruning (IMP) as a way to obtain sparse ``winning tickets'' $\mathbf{w'}$ that can be trained from initialization to reach performance to match that of the original full network $\mathcal{N}$ while using significantly fewer parameters. In IMP, model parameters are repeatedly reset to their original initialized values after pruning, before the next iteration of training. Reverting weights to values from an earlier point during training is also known as \textit{rewinding} \cite{frankle2020linear}. 
Following \citet{chen2020BertLT}, we consider both standard (no rewinding) and lottery ticket-style IMP (with rewinding) settings. In alignment with the paradigm of pre-training and fine-tuning, we treat the pre-trained model's weights as the initial weights to which parameters are reset at each iteration.

\subsubsection{Structured Pruning}
\label{sec:structured}
We also explore flat minima in a recently proposed structured pruning setting: \citet{xia2022structuredCoFi} incorporate a layerwise distillation objective into their structured pruning process, which dynamically maps layers between teacher and student models as structured units of varying granularity are incrementally pruned in the student model via an $\ell_0$ regularizer. In our experiments, we vary only optimizer used to fine-tune the \textit{teacher model} and compare downstream compression performance.

\subsection{Post-Training Quantization \label{sec:ptq}}

We compare performance of Int8 quantized BERT$_{base}$ models fine-tuned with Adam- and SAM-optimized models. Using a standard PyTorch implementation\footnote{\url{https://pytorch.org/docs/stable/quantization.html}}, we perform \textit{post-training dynamic quantization}, where full-size (32-bit) floating point model weights are statically mapped to a lower precision (in our case, 8-bit integer) representation after training, and activations are dynamically reduced in precision during inference.

\section{Experimentation}
\label{sec:experimentation}

\subsection{Research Questions}
\label{sec:RQs}
In this section, we describe a series of experiments and analyses aimed at answering the following research questions:

\begin{enumerate}
    \item [\textbf{Q0}] Does SAM help make models more robust to compression? (\S\ref{sec:imp_expts}, %
    \S\ref{sec:cofi_structured_expts}, \S\ref{sec:quantization})
    \item [\textbf{Q1}] Does SAM benefit compressibility across different model initializations and sizes? (\S\ref{sec:large_and_roberta})
    \item [\textbf{Q2}] \textit{How} does SAM influence model compressibility? (\S\ref{sec:analysis})
    \item [\textbf{Q3}] How does SAM affect compressed model properties beyond single-task accuracy? (\S\ref{sec:analysis}, \S\ref{sec:transfer_expts})%
    \item [\textbf{Q4}] How does flatness in general, beyond SAM specifically, influence compressibility? (\S\ref{sec:swa_expts})
\end{enumerate}

\subsection{Datasets and Metrics}

We consider eight tasks from the standard GLUE \citep{wang2018glue} benchmark for our experimentation, as well as SQuAD v1.1 \citep{rajpurkar2016squad}. The GLUE datasets include MNLI \citep{williams2018broad}, QQP\footnote{\url{https://quoradata.quora.com/First-Quora-Dataset-Release-Question-Pairs}}, STS-B \citep{cer2017semeval}, QNLI \citep{wang2018glue}, MRPC \citep{dolan2005automatically}, RTE \citep{wang2018glue}, SST-2 \citep{socher2013recursive}, and CoLA \citep{warstadt2019neural}. For all experiments unless otherwise noted, we follow prior work \citep{chen2020BertLT} and report validation set accuracy for QQP, QNLI, MRPC, RTE, SST-2, matched accuracy for MNLI, Matthew's correlation for CoLA, Pearson correlation for STS-B, and F1 score for SQuAD. We also make use of a sharpness metric (\S\ref{sec:sharpnessmetric}) to quantify the flatness of the basins that our models lie in.

\subsection{Implementation Details}
For all experiments described in this section, we fine-tune publicly available pre-trained BERT model weights \citep{wolf2019huggingface}. 
We use the uncased BERT$_{base}$ model for all experiments except when otherwise noted. For our iterative magnitude pruning experiments, we mainly set hyperparameters as mentioned by \citet{chen2020BertLT} and follow a similar general procedure for iterative magnitude pruning, pruning an absolute $10\%$ of prunable weights over $9$ iterations to reach $90\%$ sparsity in order to facilitate direct comparisons. Appendix~\ref{subsec:imp_reprod} contains further implementation details including hyperparameters used and explanations of when our methods differ. For our SAM optimizer, we use Adam \citep{kingma2014adam} as our base optimizer, and following \citep{mehta2021empirical, bahri22acl} set $\rho$ to $0.05$. Appendix \ref{sec:sharpnessmetric} contains implementation details for SWA.

\begin{toappendix}
the maximum sequence length of $128$, batch size of $32$, learning rate to $2e-5$, linear decay of learning rate from initial value to zero with no warmup period. For tasks with smaller datasets (RTE, MRPC, CoLA, STS-B), we fine-tune models for $10$ epochs, evaluate them after every epoch and retain the checkpoint yielding best task-specific performance on the hold-out validation set. For tasks with comparatively larger datasets (MNLI, QQP, QNLI, SST-2), we fine-tune models for $3$ epochs.   %
We set Adam's weight decay to $\epsilon=0$ in order to remove the potential confound of regularization on models' amenability to magnitude pruning. This differs from \citet{chen2020BertLT}'s $\epsilon=1\times 10^{-8}$.

We follow \citet{chen2020BertLT}'s formulation of iterative magnitude pruning and compare pruned Adam and SAM models' performance at their reported reference sparsity levels on GLUE tasks and SQuAD v1.1
\end{toappendix}

\subsection{Results}

\begin{table*}[h]
\scriptsize
    \centering
    \begin{tabular}{m{0.01\textwidth} r|c c c c c c c c c}%
    \toprule
         & Dataset & MNLI & QQP & STS-B & QNLI & MRPC & RTE & SST-2 & CoLA & SQuAD \\
         & Sparsity & $70\%$ & $90\%$ & $50\%$ & $70\%$ & $50\%$ & $60\%$  & $60\%$ & $50\%$ & $40\%$ \\
         & Metric & Match/Mismatch acc. & Acc. & Pearson Cor. & Acc. & Acc. & Acc. & Acc. & Matthew's Cor. & F1 \\
         \midrule
         Full & Adam & $84.6_{0.1}$/$83.6_{0.3}$ & $89.1_{0.1}$ & $84.0_{0.4}$ & $90.8_{0.2}$ & $82.6_{1.4}$ & $67.0_{1.4}$ & $93.2_{0.6}$ & $53.3_{1.2}$  & $88.5_{0.2}$ \\
         & SAM & $85.0_{0.1}/84.5_{0.1}$ & $89.2_{0.2}$ & $84.7_{0.1}$ & $90.8_{0.4}$ & $82.9_{1.3}$ & $65.5_{0.8}$ & $93.6_{0.1}$ & $54.1_{1.3}$ & $89.2_{0.1}$ \\
         \midrule
         \textbf{IMP} & Adam & $82.3_{0.3}/81.4_{0.2}$ & $83.0_{0.0}$ & $83.3_{0.3}$ & $88.6_{0.3}$ & $81.5_{1.3}$ & $63.3_{2.8}$ & $92.2_{0.4}$ & $47.7_{1.5}$ & $86.9_{0.3}$ \\
         & SAM & $\mathbf{83.2_{0.2}/82.5_{0.4}}$ & $\mathbf{85.4_{0.1}}$ & $\mathbf{84.1_{0.1}}$ & $\mathbf{89.4_{0.2}}$ & $\mathbf{83.6_{0.2}}$ & $\mathbf{65.0_{1.5}}$ & $\mathbf{92.9_{0.5}}$ & $\mathbf{49.5_{2.0}}$ & $\mathbf{87.8_{0.2}}$ \\
         & *SWA & $83.1_{0.1}/82.1_{0.1}$ & $\bm{\mathit{87.7_{0.2}}}$ & $\bm{\mathit{85.3_{0.4}}}$ & $89.0_{0.2}$ & $\bm{\mathit{83.7_{0.1}}}$ & $\bm{\mathit{67.4_{0.4}}}$ & $92.7_{0.3}$ & $\bm{\mathit{50.9_{1.0}}}$ & $-$ \\
         \midrule
         \textbf{Std} & Adam & $82.4_{0.3}$/$81.1_{0.5}$ & $87.2_{0.1}$ & $83.9_{0.1}$ & $88.8_{0.1}$ & $82.9_{1.0}$ & $64.4_{1.6}$ & $92.3_{0.2}$ & $50.9_{0.2}$  & $86.3_{0.2}$ \\
         & SAM & $83.2_{0.1}$/$81.8_{0.1}$ & $87.4_{0.4}$ & $84.6_{0.1}$ & $89.4_{0.1}$ & $81.9_{1.2}$ & $64.3_{0.8}$ & $93.0_{0.6}$ & $51.3_{2.0}$  & $87.1_{0.2}$ \\
    \bottomrule
    \end{tabular}
    \caption{At full size and at \citet{chen2020BertLT}'s reference sparsities, we report task-specific metrics for Adam and SAM-optimized BERT$_{base}$ models in their (1) Iterative Magnitude Pruning (\textbf{IMP}) and (2) \textbf{Std} pruning settings. We report mean and standard deviation calculated over 3 random seeds. All GLUE results are reported for test sets. We report results on the development set for SQuAD as test set evaluation is unavailable for v1.1.
     Table~\ref{tab:recreateBERTLT_DEV} in Appendix~\ref{subsec:imp_devset} contains comparison with reference (\textbf{Ref}) values using development sets. *Additionally, we report test accuracy metrics in IMP models trained with stochastic weight averaging (SWA), another optimization method which empirically leads to flatter minima. We observe that optimizing with SAM or SWA throughout iterative magnitude pruning allows pruned models to retain higher performance at reference sparsity levels when compared to models trained with vanilla Adam optimizer.}
    \label{tab:recreateBERTLT}
\end{table*}

\begin{figure*}[h]
    \centering
    \begin{subfigure}{.245\textwidth}
      \centering
      \includegraphics[width=\textwidth]{"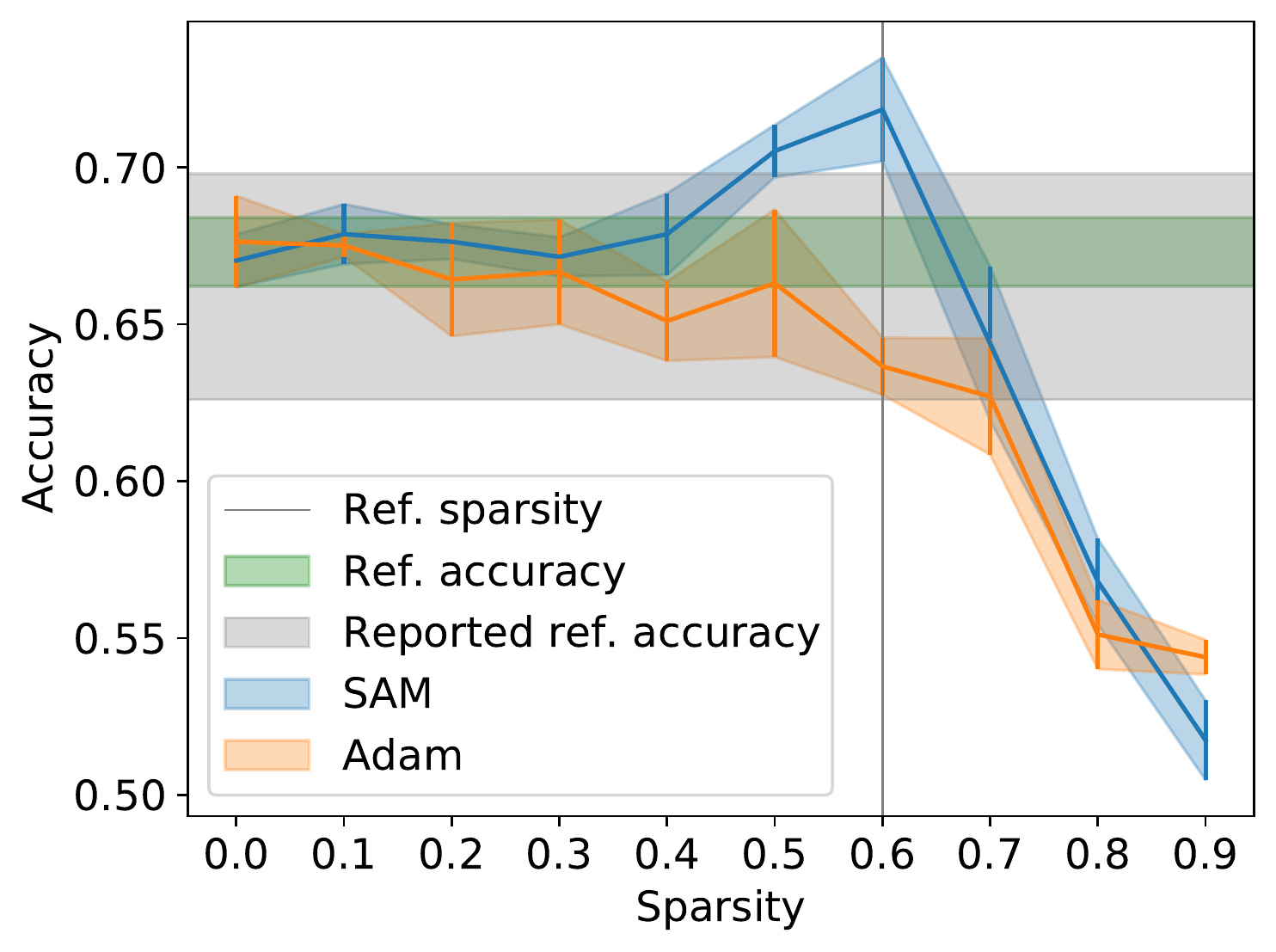"}
      \subcaption{RTE}
      \label{fig:rte_plot}
    \end{subfigure}\hspace{\fill}%
    \begin{subfigure}{.245\textwidth}
      \centering
      \includegraphics[width=\textwidth]{"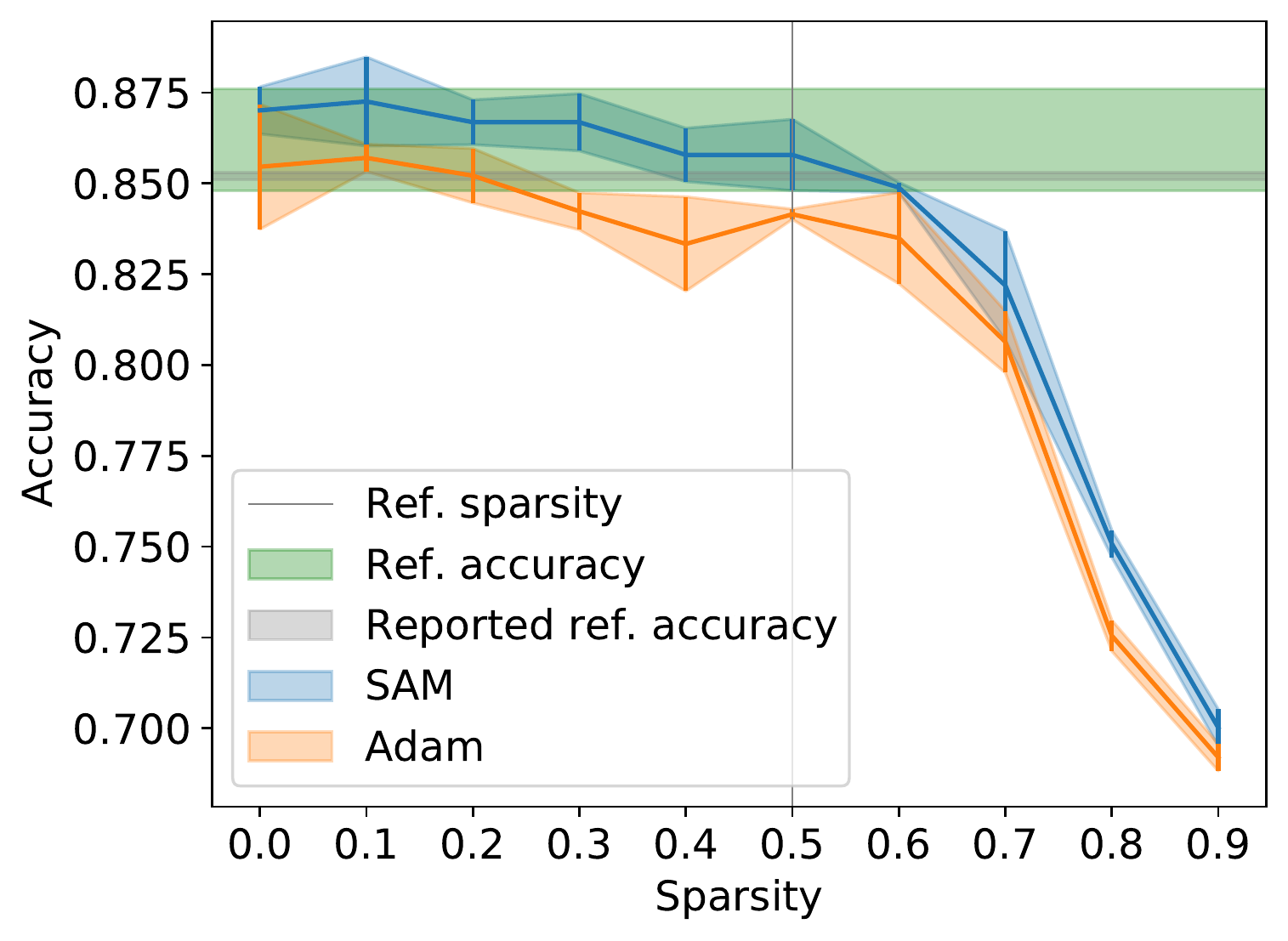"}
      \caption{MRPC}
      \label{fig:mrpc_plot}
    \end{subfigure}\hspace{\fill}%
    \begin{subfigure}{.245\textwidth}
      \centering
      \includegraphics[width=\textwidth]{"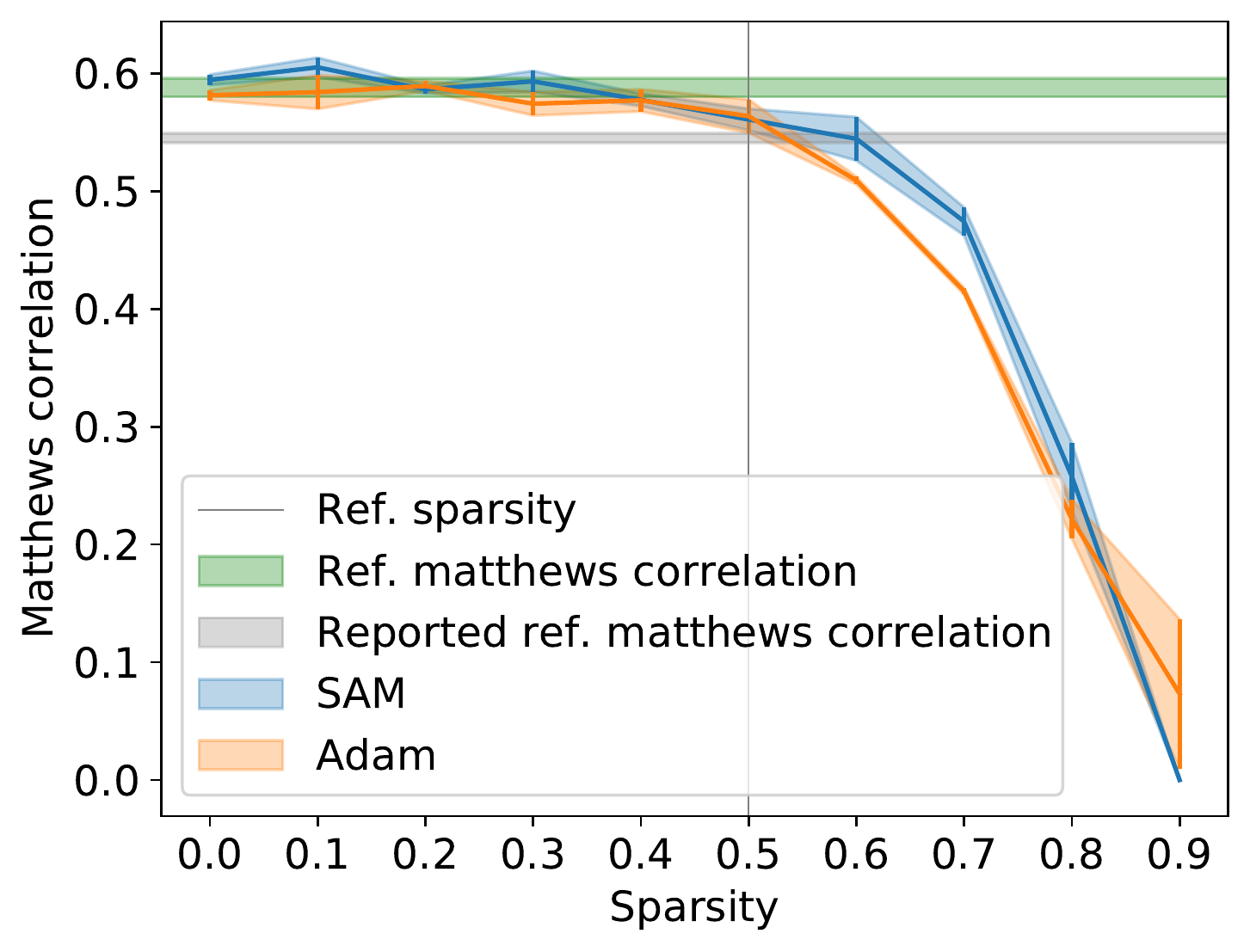"}
      \caption{CoLA}
      \label{fig:cola_plot}
    \end{subfigure}\hspace{\fill}%
    \begin{subfigure}{.245\textwidth}
      \centering
      \includegraphics[width=\textwidth]{"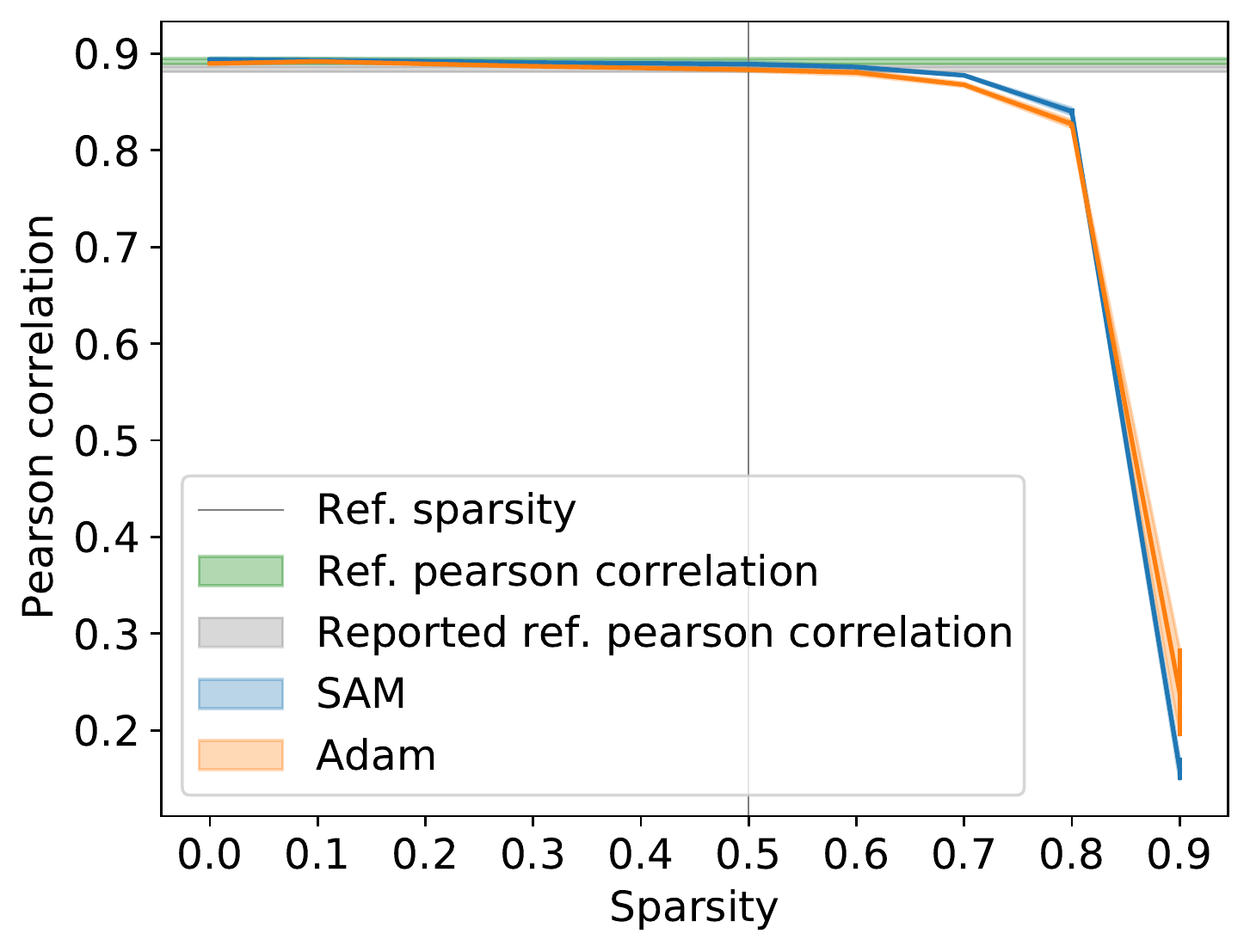"}
      \caption{STS-B}
      \label{fig:stsb_plot}
    \end{subfigure}\hspace{\fill}%
    \bigskip
    \begin{subfigure}{.245\textwidth}
      \centering
      \includegraphics[width=\textwidth]{"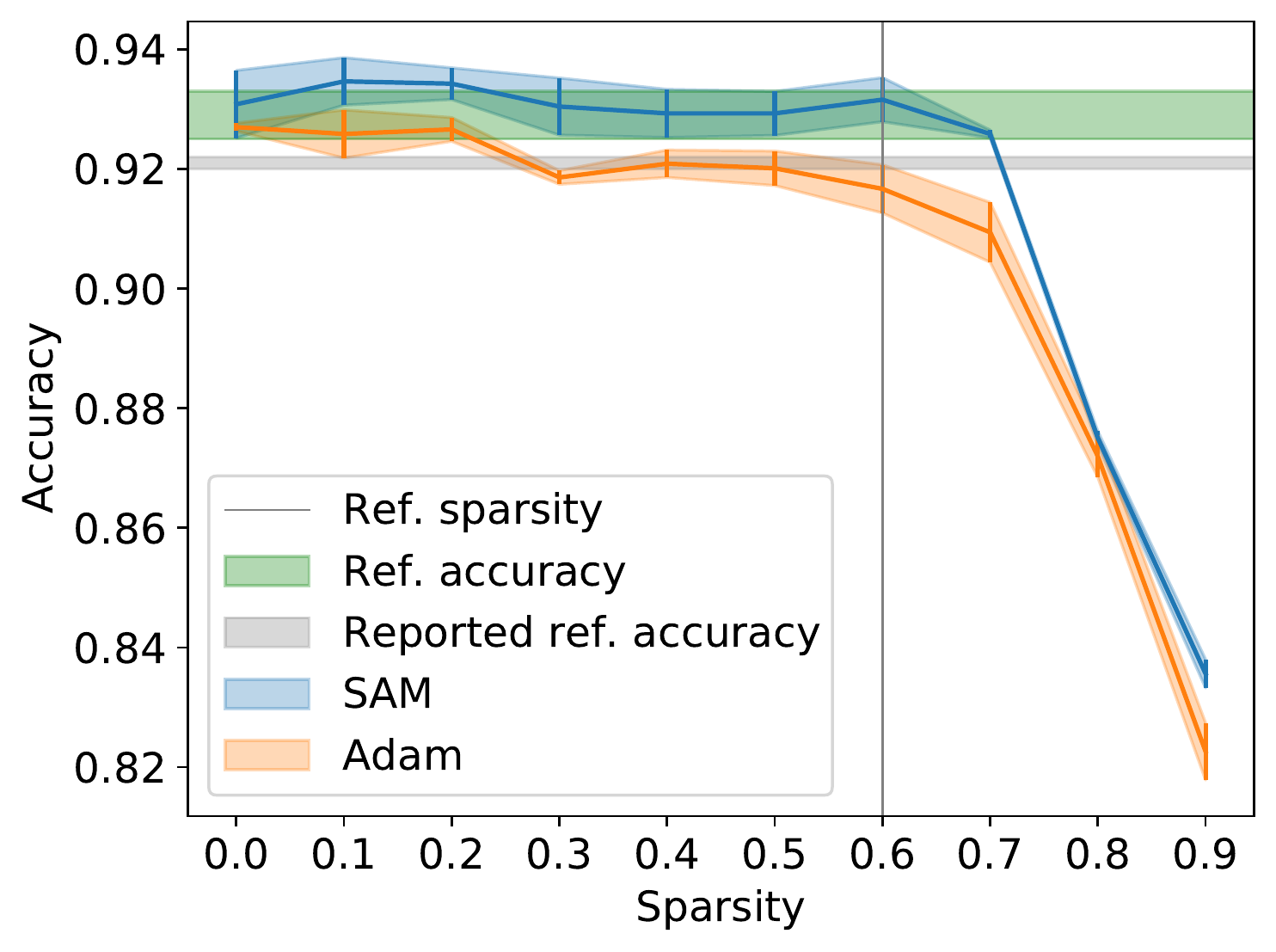"}
      \subcaption{SST-2}
      \label{fig:sst2_plot}
    \end{subfigure}\hspace{\fill}%
    \begin{subfigure}{.245\textwidth}
      \centering
      \includegraphics[width=\textwidth]{"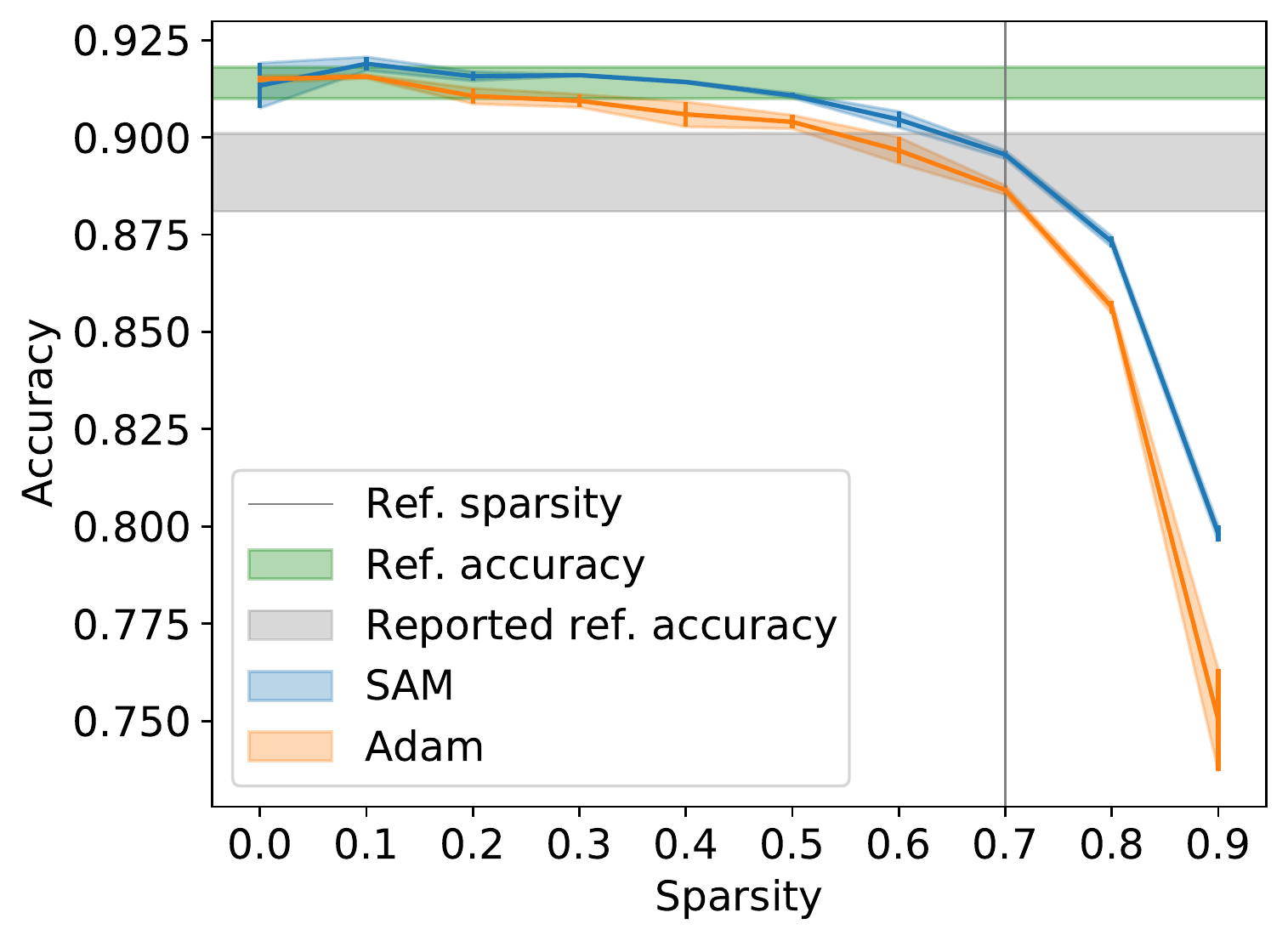"}
      \caption{QNLI}
      \label{fig:qnli_plot}
    \end{subfigure}\hspace{\fill}%
    \begin{subfigure}{.245\textwidth}
      \centering
      \includegraphics[width=\textwidth]{"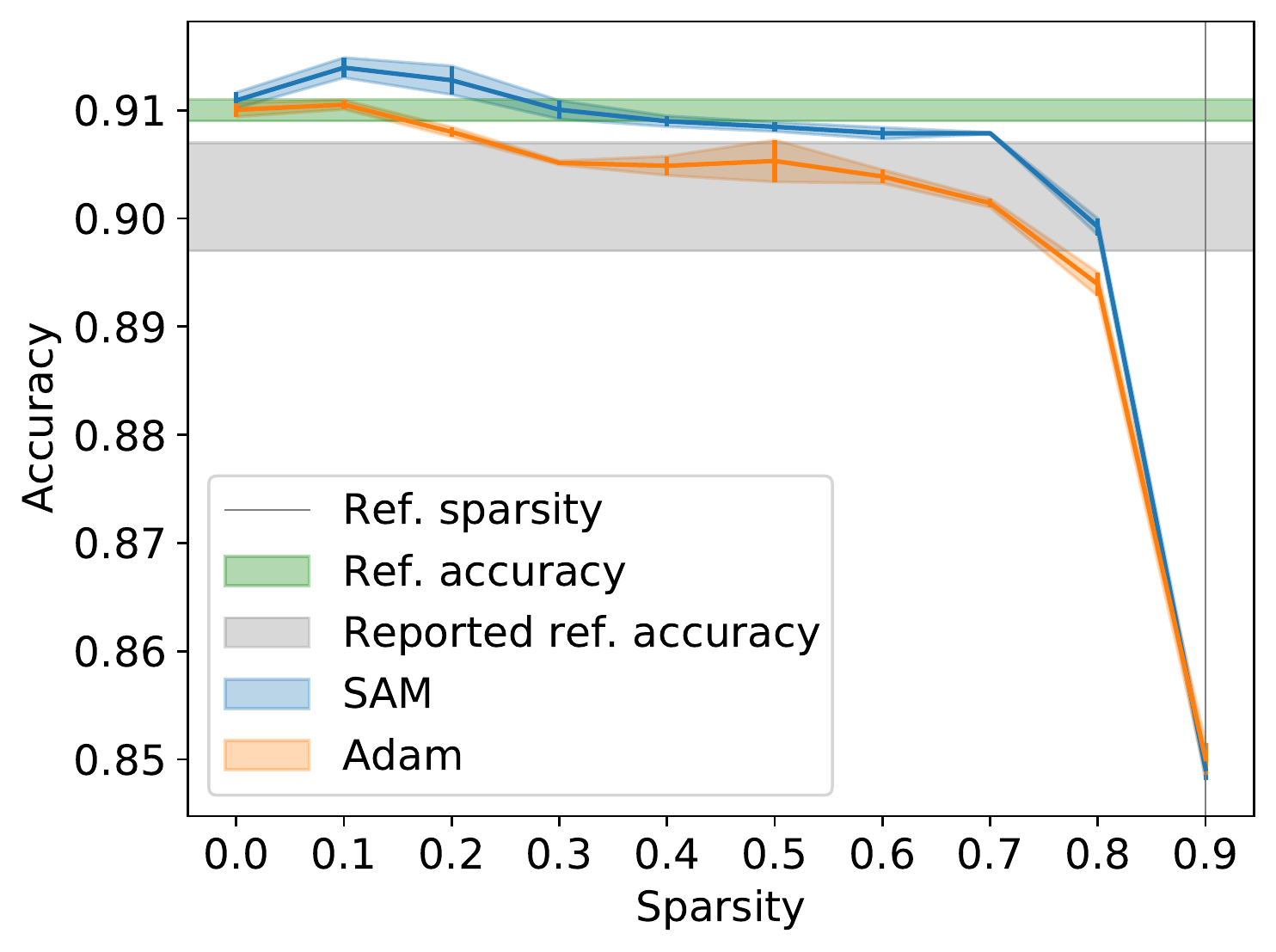"}
      \caption{QQP}
      \label{fig:qqp_plot}
    \end{subfigure}\hspace{\fill}%
    \begin{subfigure}{.245\textwidth}
      \centering
      \includegraphics[width=\textwidth]{"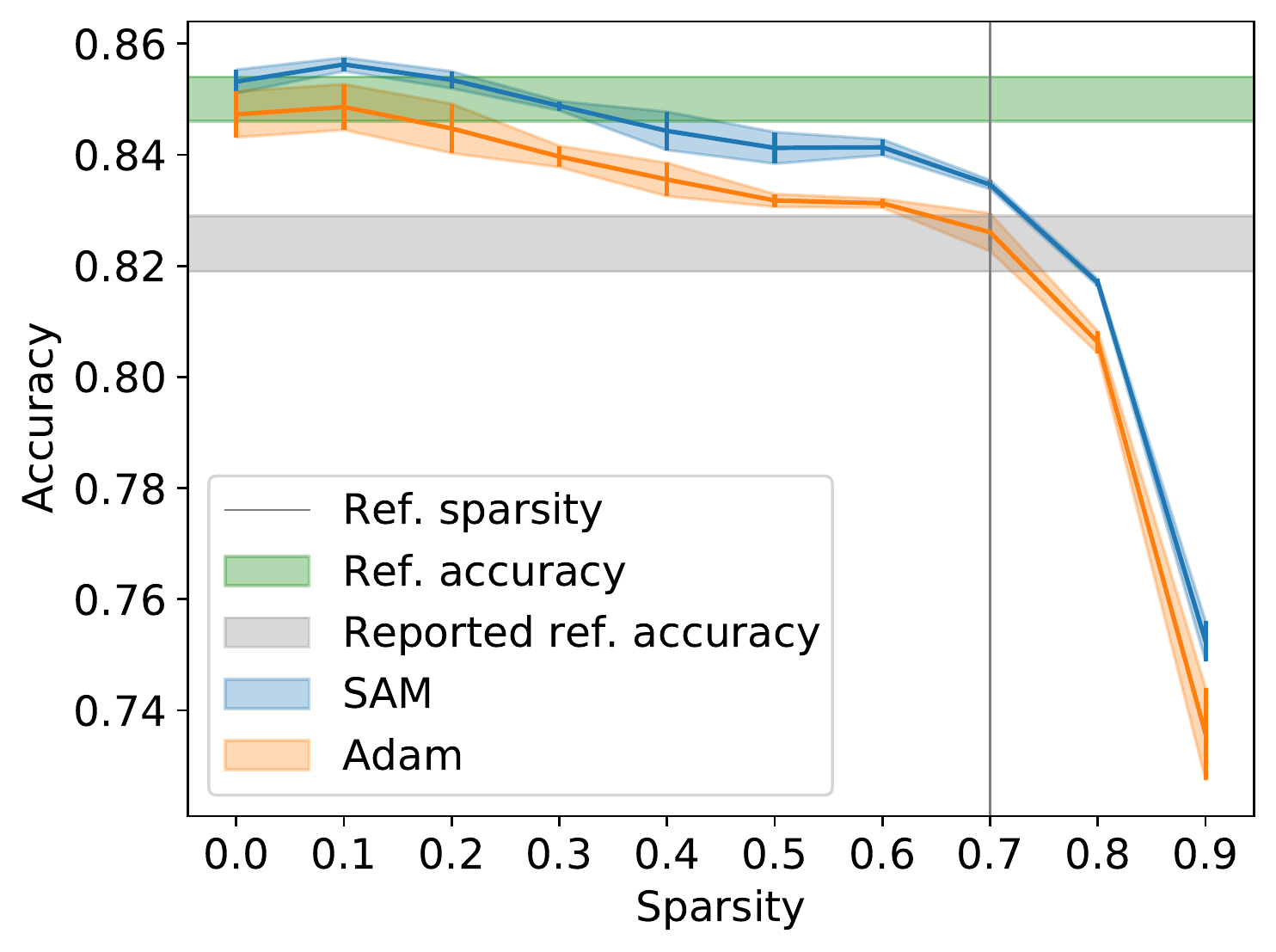"}
      \caption{MNLI}
      \label{fig:mnli_plot}
    \end{subfigure}\hspace{\fill}
    \caption{Individual plots showing sparsity vs. task metrics (validation set) for GLUE throughout IMP. The vertical lines and gray horizontal bands mark reference sparsity and "winning ticket" evaluation metric values that were obtained by \citet{chen2020BertLT}. The green horizontal bands mark the initial performance of our full fine-tuned (uncompressed) models.}
    \label{fig:task_IMP_plots}
\end{figure*}

\subsubsection{Iterative Magnitude Pruning (Q0)}
\label{sec:imp_expts}

\paragraph{With rewind to BERT$_{base}$} We investigate the SAM procedure's effectiveness in uncovering winning tickets \cite{frankle2018lotteryticket}. 
The \textbf{IMP} section of Table~\ref{tab:recreateBERTLT} shows that \textbf{optimizing with SAM throughout iterative magnitude pruning allows pruned models to retain higher performance} at reference sparsity levels when compared to models trained with vanilla Adam optimizer.

The plots in Figure~\ref{fig:task_IMP_plots} show evaluation metrics over successive IMP iterations for individual GLUE tasks. We see that although initial performance of Adam- and SAM-optimized models is usually comparable, promoting flat minima during IMP with SAM leads to either 1) higher performance compared to Adam at \citet{chen2020BertLT}'s reference sparsity level\footnote{Our Adam optimized models often differ in performance from \citet{chen2020BertLT}'s at iteration $0$ -- the fairest comparison is between our implementation of Adam and SAM optimized models, and our reference performance bands in Figure~\ref{fig:task_IMP_plots} are often much higher than originally reported.} or 2) performance comparable to the full sized model at higher sparsity levels, if not both, for all but one smaller GLUE task (CoLA).

Moreover, for some tasks (RTE, SST-2), \textbf{the final pruned model's accuracy tends to be higher than the full BERT$_{base}$ fine-tuned model's}. This is an especially striking result given that we reset remaining weights to the BERT$_{base}$ initialization after each successive iteration of pruning; there is no progressive learning of weights from iteration to iteration. Instead, we reach higher accuracy simply by optimizing over the learned substructures rather than the full network.

Both SAM and vanilla Adam induce marginal improvements compared to the full fine-tuned model with $10\%$ pruning for some tasks (e.g. \ref{fig:mrpc_plot}, \ref{fig:qnli_plot}, \ref{fig:qqp_plot}), which we might attribute to slight pruning acting as a structure regularizer. However, only SAM-optimized models sometimes continue to exhibit improvements in much later stages of pruning.

\paragraph{"Standard pruning" without rewind}%
We also investigate SAM and vanilla Adam optimizers in the standard pruning setting, where we continue training immediately after pruning in each iteration, without resetting remaining weights to pre-trained BERT$_{base}$ initialization. The \textbf{Std} section of Table~\ref{tab:recreateBERTLT} shows these results.

SAM-optimized models retain more of the full sized model's performance at \citet{chen2020BertLT}'s reference sparsity levels. However, the trend is not more stark in this setting, which hints at the structure of winning tickets found by SAM playing an important role. 

We more directly investigate \textit{how} SAM benefits model compressibility in Section~\ref{sec:analysis}, but we also take care to rule out the possibility that SAM is simply acting as an implicit $\ell_1$ regularizer (\ref{subsec:l1reg_appendix}).

\subsubsection{Analysis: Answering the Structure vs. Optimization Question (Q2, Q3)}
\label{sec:analysis}
\begin{figure*}[h!]
    \centering
    \includegraphics[width=\textwidth]{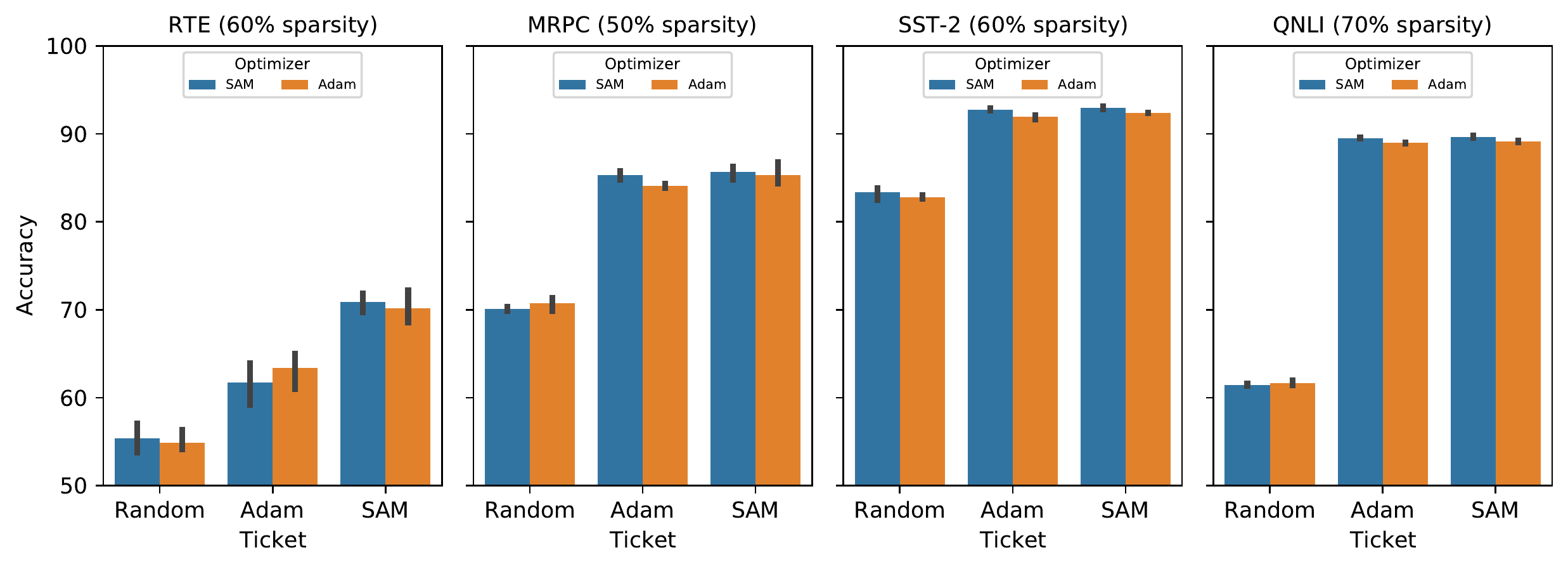}
    \caption{We compare optimizers' \textit{learned tickets}, as well as their performance in \textit{optimization over a given ticket}. For select GLUE tasks at their reference sparsity values, we fine-tune pruned subnetworks of pre-trained BERT$_{base}$ initializations networks based on 1) random masks, 2) Adam-learned masks, and 3) SAM-learned masks, using a) SAM and b) Adam optimizers.  \textcolor{RoyalBlue}{SAM} optimization over SAM- and Adam-learned winning tickets tends to yield marginal improvements compared to \textcolor{YellowOrange}{Adam} optimization. Comparing bar heights from left to right within each figure allows us to see that, at least when \textcolor{YellowOrange}{Adam} is used for final fine-tuning, a Random- $\ll$ Adam- $<$ SAM-learned masks. Exact values are reported in Table~\ref{tab:OptimizerSwapAnalysis}, and Figure~\ref{fig:OptimizerSwapAltView} in Appendix shows an alternative view of the data.}
    \label{fig:OptimizerSwap}
\end{figure*}

In this experimental setting, we aim to disentangle the effects of the \textit{structure}\footnote{In this work, we refer to "winning tickets" and "structures" thereof interchangeably, although, as observed by \citet{frankle2018lotteryticket}, winning tickets are conditioned on models' (in our case, pre-trained) initializations.} of the pruning masks learned using different optimizers, from the \textit{optimization} over given substructures using different optimizers. Figure~\ref{fig:OptimizerSwap} displays our results. 

We observe that, in general, subnetworks learned through IMP greatly outperform random subnetworks of the same sparsity when trained. %
Subnetworks found with SAM tend to outperform those found with vanilla Adam, especially with subsequent optimization with Adam. Furthermore, training any given IMP-learned subnetwork with SAM yields modest improvements in accuracy compared to fine-tuning with vanilla Adam.

\subsubsection{Analysis: Comparing Transferability of Winning Tickets (Q3)}
\label{sec:transfer_expts}

\begin{figure}[h]
\centering
        \begin{subfigure}{.45\textwidth}
      \centering
      \includegraphics[width=\textwidth]{"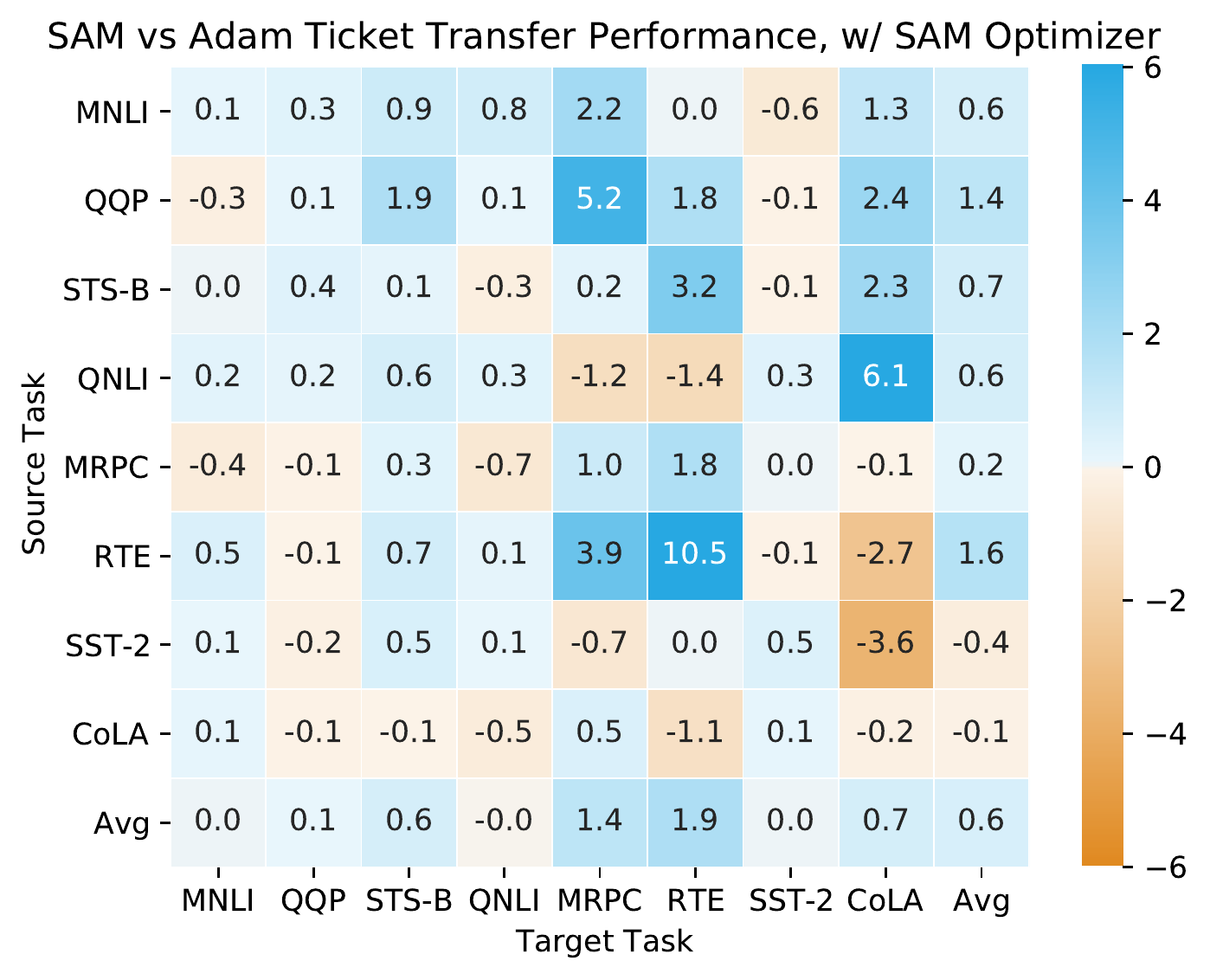"}
    \end{subfigure}
    \begin{subfigure}{.45\textwidth}
      \centering
      \includegraphics[width=\textwidth]{"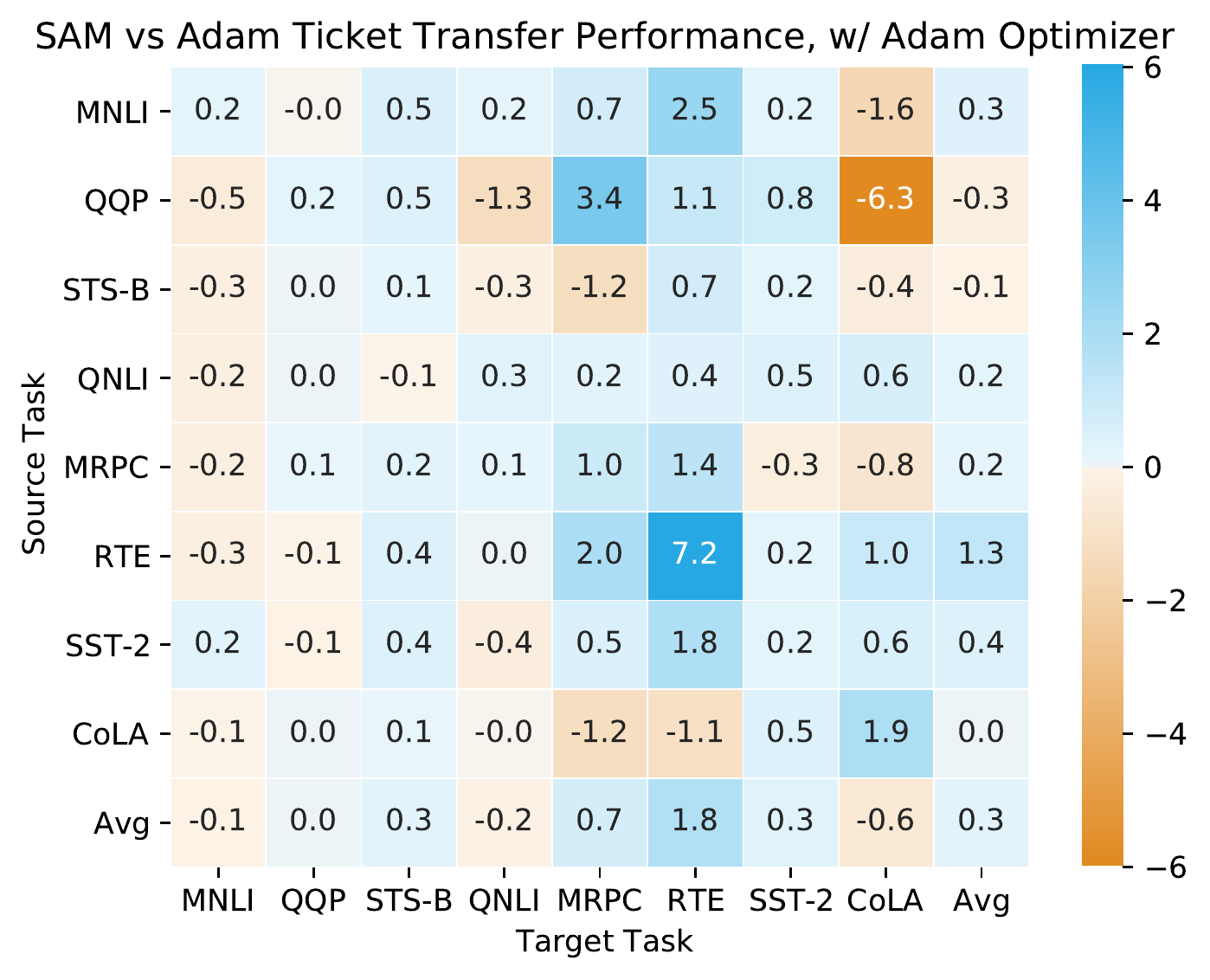"}
    \end{subfigure}\hspace{\fill}%
\caption{Heatmaps indicating the \textit{difference} in target task performance between $70\%$ sparsity SAM and Adam tickets when transferring tickets across tasks, fine-tuned with either SAM (top) or Adam (bottom) optimizers during IMP. Values $>$ 0 indicate the extent to which \textcolor{RoyalBlue}{SAM tickets transferred better} than Adam tickets; Values $<$ 0 indicate where \textcolor{YellowOrange}{Adam tickets transferred better} than SAM. Overall, SAM tickets transfer better regardless of the final fine-tuning optimizer. Note that the positive values along the diagonal indicate superior SAM ticket performance in the single task setting, even with ``transfer'' between optimizers. \label{fig:transfer-heatmaps}}
\end{figure}

We explore the extent to which SAM tickets are more or less transferable across tasks compared to tickets discovered by Adam (Figure~\ref{fig:transfer-heatmaps}). For consistency, we use $70\%$ sparsity tickets for all tasks evaluated. SAM-learned tickets tend to transfer better across tasks than Adam-learned tickets. This complements our findings in \S\ref{sec:analysis}, which can be interpreted as a study on transferability of winning tickets between optimizers instead of across tasks.

We also compare SAM versus Adam as an optimizer for fine-tuning in this setting, and found that SAM did \textit{not} seem to work better overall as an optimizer, given a ticket for a different task (see Figure~\ref{fig:optimizer-transfer-heatmaps} in Appendix). Note that this does not directly contradict our results from \S\ref{sec:analysis}; SAM optimization does typically benefit same-task performance (see diagonals in these figures).

\subsubsection{Structured Pruning (Q0)}
\label{sec:cofi_structured_expts}
We use full-size fine-tuned BERT$_{base}$ models from \S\ref{sec:imp_expts} as teacher models for the layerwise distillation objective used throughout the structured pruning procedure. The pruning procedure itself is the same as \citet{xia2022structuredCoFi}'s, regardless of whether the teacher model was originally trained using SAM or vanilla Adam. Appendix~\ref{subsec:cofi_appendix} contains further implementation details.

\begin{table}[h]
\small
    \centering
    \begin{tabular}{l l | c | c}
    \hline
         Dataset & Optim. & Teacher Acc. & Pruned Acc. \\
         \hline
         SST-2 & Adam & $92.7_{0.1}$ & $90.2_{0.5}$ \\ 
         (67k) %
          & SAM & $93.1_{0.6}$ & $\mathbf{91.3_{0.3}}$ \\ %
    \hline
         QNLI & Adam & $91.5_{0.1}$ & $85.9_{0.4}$ \\
         (105k)
          & SAM & $91.3_{0.6}$ & $\mathbf{86.9_{0.4}}$ \\
    \hline
         QQP & Adam & $91.0_{0.1}$ & $90.0_{0.1}$ \\
         (364k) 
          & SAM & $91.1_{0.1}$ & $90.1_{0.1}$ \\
    \hline
         MNLI & Adam & $84.7_{0.4}$ & $80.2_{0.4}$ \\
         (393k) 
          & SAM & $85.3_{0.2}$ & $80.6_{0.1}$ \\
    \hline
    \end{tabular}
    \caption{Comparison between pruned models obtained using teacher models fine-tuned with Adam and SAM optimizers. Numbers reported are means$_{stddev}$ ($n=3$) for evaluation metrics on the development set. Compressed models are trained to reach $95\%$ sparsity using optimal values for $\lambda$ and finetuning learning rate from \citet{xia2022structuredCoFi}'s structured pruning setting.}
    \label{tab:CoFiTable}
\end{table}

In Table~\ref{tab:CoFiTable}, we show that SAM-optimized teacher models improve compressed student model performance in this structured pruning setting. 
SAM outperforming a vanilla Adam optimizer in this setting is a particularly desirable goal from a practical standpoint. First, unlike in a full IMP setting, we only use SAM for a single fine-tuning in the pruning pipeline, and so we only incur the computational overhead associated with SAM for a fraction of the overall pruning process. Second, training time for CoFi's pruning process itself has a tenfold speedup compared to the TinyBERT baseline \cite{jiao-etal-2020-tinybert}. Finally, the pruned models obtained in this setting perform inference as quickly as TinyBERT, which amounts to a tenfold speedup compared to the full BERT$_{base}$ model.

\begin{braindump}
Block Pruning (just pruning, we can expect higher accuracy at higher sparsities but less inference speedup overall)

CoFi Pruning (uses (layerwise) distillation objective for structured pruning. Bigger inference speedups)

\end{braindump}

\subsubsection{Post-Training Quantization (Q0)}
\label{sec:quantization}

\begin{figure}[ht!]
    \centering
    \includegraphics[scale=0.55]{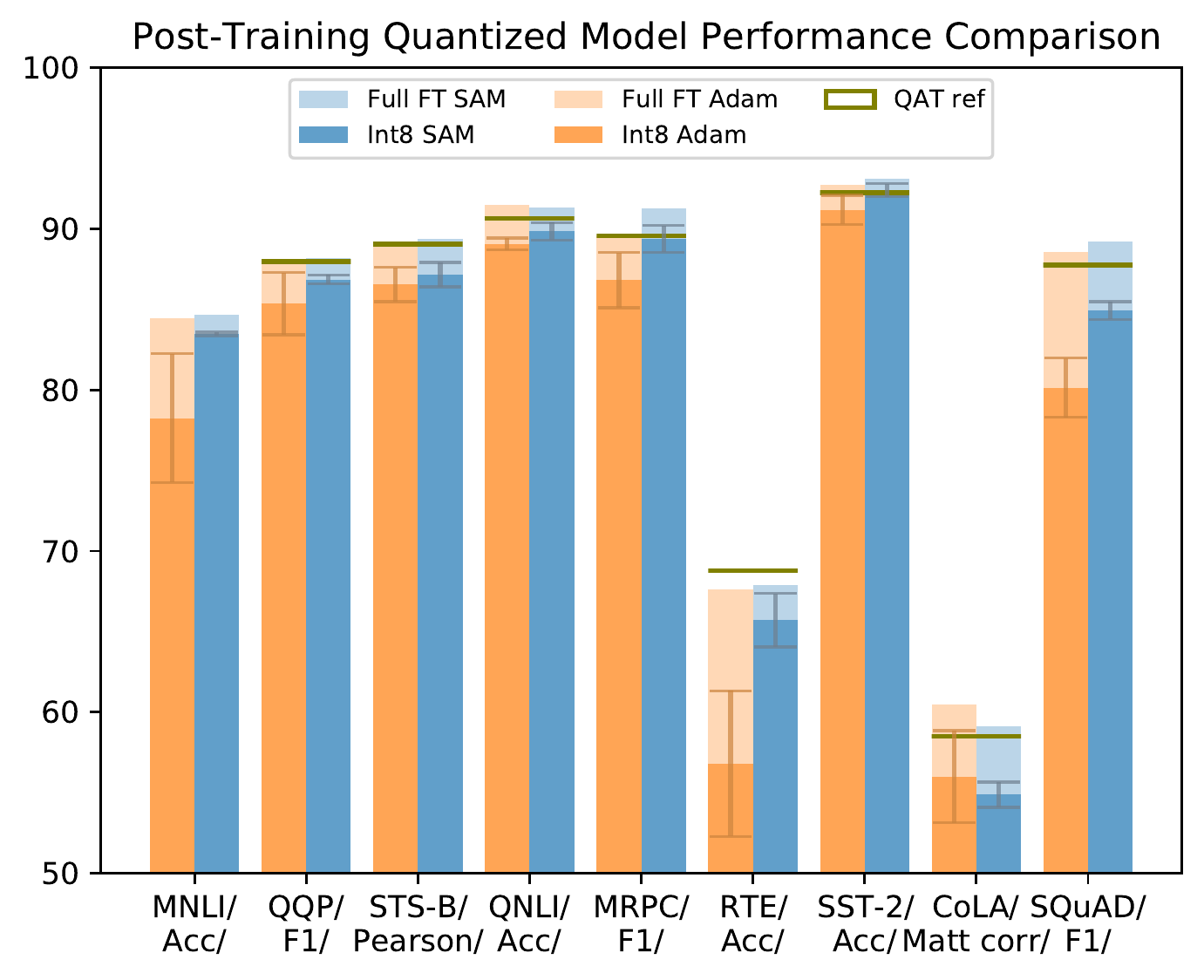}
    \caption{We compare full fine-tuned and quantized BERT-base models optimized with SAM and Adam. Error bars show standard deviations for $n=3$. Additionally, we show that applying a simpler post-training dynamic quantization technique on a SAM-optimized model can approach the reported performance of a model quantized through quantization-aware training (QAT) \cite{Zafrir2019Q8BERTQ8}. }
    \label{fig:ptdq_quant}
\end{figure}
From Figure~\ref{fig:ptdq_quant} (see \ref{subsec:more_quant} for actual numbers), we can make a few key observations: 1) SAM-trained models retain higher task performance after quantization compared to Adam-trained models. 2) SAM-trained models' higher performance is also more \textit{stable} across random seeds. Finally, 3) for some tasks, our SAM-trained models compressed with simple post-training dynamic quantization meet or approach the performance of \citet{Zafrir2019Q8BERTQ8}'s models quantized through quantization-aware training (QAT), which tends to yield better compressed models but is more complex to implement and use. The benefits of quantization vary under different settings and hardware, but our BERT$_{base}$ models quantized to Int8 precision have a 2.5x reduction in storage requirements and 1.5-2x faster inference.

\subsubsection{Evaluating Other Models (Q1)}
\label{sec:large_and_roberta}
In order to explore the applicability of our findings across model sizes and other BERT variants, 
we experiment with SAM and Adam optimizers on BERT$_{large}$ and RoBERTa$_{base}$ models in the IMP setting of \S\ref{sec:imp_expts}. We find that, indeed, SAM-optimized models fare better than Adam-optimized models in these other BERT variants. \ref{subsec:other_models} includes details relevant for reproducibility, as well as task-specific results.

More generally, our proposal to "train flat" with SAM is compatible with \citet{li2020train}'s recommendation to "train large, then compress", and with starting with a better-performing model before compression. Starting with BERT$_{large}$ or RoBERTa$_{base}$ can lead to clearly higher compressed accuracy at similar target sparsity levels and rarely leads to significantly \textit{worse} performance. However, the higher performance often does not simply follow a parallel pattern as in BERT$_{base}$ models throughout pruning; the initial performance gaps between BERT$_{large}$ and BERT$_{base}$ models tend to be preserved slightly more reliably at higher sparsity levels than the gaps between RoBERTa$_{base}$ and BERT$_{base}$ models. This prompts further investigation into the properties of the subnetworks found in these BERT variants and appeals to the potential compressibility of even larger models when flatness-optimized. 

\subsubsection{Stochastic Weight Averaging (Q4)}
\label{sec:swa_expts}
We conduct experiments matching the IMP setting with rewind from \S\ref{sec:imp_expts} using stochastic weight averaging (SWA). In Table \ref{tab:recreateBERTLT}, we report results on GLUE test set for SWA with IMP. Similar to SAM, we observe that SWA is superior to Adam optimized models at reference sparsity levels.

\section{Related Work}
\label{sec:relatedwork}

In this section, we draw connections to key related work. For a more general description of related work, please refer to \ref{sec:append_relatedwork} in Appendix.

First, we briefly recount previous work that we consider for our experimentation. We use Adam \cite{kingma2014adam} as a base optimizer, with comparisons between vanilla Adam and the addition of Sharpness-Aware-Minimization \cite{foret2021sharpnessaware}. We make further comparisons in a subset of our experiments with Stochastic Weight Averaging \cite{izmailov2018averaging} as an alternative method of inducing flatness. Using \citet{keskar2016large}'s $\epsilon$-sharpness metric, and based on \citet{mehta2021empirical}'s implementation, we verify that SAM and SWA induce flatness. We fine-tune BERT models \cite{devlin-etal-2019-bert, Liu2019RoBERTaAR} for GLUE \cite{wang2018glue} and SQuAD \cite{rajpurkar2016squad} language tasks. The compression methods we couple with ``training flat'' include: 1) unstructured IMP to find winning lottery tickets \cite{frankle2018lotteryticket}, referring to \citet{chen2020BertLT}'s implementation and reported results for making direct comparisons; 2) \citet{xia2022structuredCoFi}'s structured pruning method with a distillation objective; and 3) off-the-shelf post-training quantization, which we compare with reported results from \citet{Zafrir2019Q8BERTQ8}'s quantization-aware training method. %

\paragraph{Flatness and generalization}
Prior work has investigated the connection between flat minima and generalization (i.e., the gap between training accuracy and holdout set accuracy), starting with \citet{hochreiter1997flat}.

Subsequent work has continued on this front, exploring notions of sharpness and their predictiveness of generalization under different conditions \cite{pmlr-v151-bisla22a-sharpness-measures}, as well as empirical evaluations of flatness-inducing methods and their effects on generalization.
\citet{Jiang*2020Fantastic} find that sharpness is empirically predictive of generalization, including in particular a perturbation magnitude-aware metric very similar to the $\epsilon$-sharpness metric introduced in \cite{keskar2016large} and used in our paper. In this work, however, separate from generalization, %
we primarily focus on investigating the underexplored relationship between flatness and \textit{compression}.

\paragraph{Flatness and compression}
Previous work has mentioned flatness in the context of pruning. In fact, \citet{hochreiter1997flat}'s original ``Flat Minimum Search'' algorithm is explicitly designed to prune units, weights, and input lines as a mechanism for finding flat minima. \citet{lecun1989optimal} also propose pruning unimportant weights as a way to obtain improved generalization, although without any notions of flatness.

Since these earlier works, the deep learning landscape has changed such that effective model compression itself is now often a goal; model efficiency in terms of size and latency is a common priority. %

To the best of our knowledge, we are the first to directly relate loss landscape flatness with model compress\textit{ibility}. We view our results as complementary to the concurrent work by \citet{paul2022unmasking}, who study properties of winning tickets found during iterative magnitude pruning and find that IMP itself preferentially prunes parameters lying along flatter directions in the loss landscape; they also theorize that flatter landscapes allow for more aggressive pruning. We note that the optimizer, data, and model architectures they use are different from ours. While \citet{paul2022unmasking} use Stochastic Gradient Descent for image classification tasks on ResNet architectures, we use Adam as a base optimizer for text classification and question answering tasks with BERT architectures. Nonetheless, to the extent that findings from both papers can generalize to other settings, \citet{paul2022unmasking} lay theoretical groundwork which supports our explicit suggestion to ``train flat'' as a strategy for \textit{inducing} greater compressibility in neural models.

\section{Discussion}
\label{sec:discussion}

We show that in general, SAM helps models retain higher accuracy on a variety of language tasks at higher sparsity levels. This holds true in multiple unstructured iterative magnitude pruning settings, as well as in a structured pruning setting with a distillation objective. Moreover, our additional experiments and analyses point to SAM-learned \textit{structures} playing an important role in compressibility, as well as transferring well across tasks.

\subsection{Future work}

\paragraph{Beyond SAM and SWA (Q4)}
In this paper, we explore \cite{foret2021sharpnessaware}'s Sharpness-Aware-Minimization procedure specifically as a method for directly reaching flat minima and conduct additional comparisons with stochastic weight averaging \cite{izmailov2018averaging}. However, other methods such as entropy SGD \citep{Chaudhari2017EntropySGDBG} and label noise SGD \citep{damian2021label} have also been shown to encourage convergence to flat minima. Further exploration would provide clarity on the role of flatness in general in model compressibility versus properties specific to SAM and SWA. 

\paragraph{The role of pre-training (Q4)}
The BERT models we fine-tune in this work are pre-trained on various self-supervised auxiliary objectives with the goal of learning useful general representations of the English language. Recent work \cite{mehta2021empirical} has found that, empirically, pre-training is associated with convergence to flatter minima than obtained by training on the same task to the same accuracy from a random initialization. Subsequent work could compare compressibility of pre-trained models versus models trained from random initialization, as well as investigate the potential for further improving compressibility through flat \textit{pre-}training. Inducing flatness during pre-training would facilitate further experimentation with end-task-agnostic knowledge distillation.

\paragraph{Other evaluations (Q3)}
In general, we evaluate model performance in terms of task-specific metrics (on development/test splits) throughout this work. However, since compression is associated with negative consequences for model behavior not captured by task-specific accuracy \cite{hooker2020characterising, liebenwein2021lost}, there is a particular need for work to understand and influence these qualities. \citet{ribeiro-etal-2020-beyond} propose behavioral testing of NLP models to evaluate specific capabilities such as robustness to typos and simple coference resolution. \citet{xu-etal-2021-beyond} propose measures of probability and label \textit{loyalty} and robustness to input perturbations for evaluating compressed models beyond preserved accuracy. We briefly discuss preliminary observations of model behaviors with respect to \citet{ribeiro-etal-2020-beyond}'s Checklist items in \S\ref{sec:checklist_expts} in Appendix, but we emphasize that more work is needed to understand behaviors and behavior shifts of vanilla Adam and SAM models before and throughout pruning.

\section*{Limitations}
\label{sec:limits}

Many of the limitations of our work have to do with its computational requirements. First, the standard implementation of Sharpness-Aware Minimization that we use incurs significant computational overhead, so further investigation into adaptive \cite{pmlr-v139-kwon21b} and more efficient  \cite{sat4free, du2022efficientSAM} variations of SAM is warranted before considering adoption of our methods in practice. Meanwhile, we control for the number of training steps and sparsity level of our models without regard to wall-clock time in our experiments, but fine-tuning BERT$_{base}$ with standard SAM typically results in 1.5x-2x slower optimization steps. In general, many of our approaches are not strategies that can simply be applied off the shelf for practical benefits. In particular, current hardware and frameworks do not typically support reliable and proportionate efficiency gains from quantization to arbitrary precision and unstructured magnitude pruning. Moreover, the computation and storage requirements for our iterative magnitude pruning scheme are tenfold compared to the typical single fine-tuning performed on a pre-trained language model, due to the ten iterations performed and checkpoints saved. We benefited from access to a large compute cluster with dozens of GPUs, including A6000, a100, v100, RTX3090, RTX8000, and 2080Ti GPUs. We estimate having used at least 1000 GPU hours across experiments for this work, which inherently limits the full reproducibility of our results in limited compute scenarios.

Additionally, although we focus on optimizing for flatness directly in our experimentation with SAM \citep{foret2021sharpnessaware}, other methods, such as weight averaging \citep{izmailov2018averaging} (which we explore in less detail), entropy SGD \citep{Chaudhari2017EntropySGDBG}, and label noise SGD \citep{damian2021label}, have also been shown to encourage convergence to flat minima. Without explicitly exploring compressibility in models that have reached flat minima in alternative ways, we refrain from making strong claims about flat minima in general in this work.

Furthermore, the tasks we train our models on are limited to sentence classification and question answering tasks, all in only the English language. We evaluate our methods using only BERT variants: pre-trained language models trained with a token masking objective. 

Finally, although our measures of end-task accuracy and degree of compression are standard and useful for evaluating and comparing compressed models, they do not provide a complete picture of model behaviors and capabilities. Other desirable characteristics in compressed models (and models in general) include but are not limited to robustness to distribution shift, stability against catastrophic forgetting, and fairness in performance across demographic groups..

\section*{Ethics Statement}
\label{sec:ethics}
We reiterate that we are not proposing that our strategies or models be adopted off the shelf as is. This is especially true because our work does not include rigorous analysis of our compressed models' properties and behaviors outside of task accuracy, resilience to compression, and transferability to other tasks. Detailed study of properties such as long-tail performance, robustness to data distribution shifts, and fairness in performance across demographic groups, for example, which have important real-world implications, is outside of the scope of our current work. However, preliminary evaluations on certain model behaviors and properties suggests that many of our models which achieve high end-task performance are vulnerable to simple perturbations in data and lack basic desirable linguistic capabilities (although not necessarily more so than is ``typical'' of language models \cite{ribeiro-etal-2020-beyond, xu-etal-2021-beyond}.  %

\section*{Acknowledgements}
We are grateful to our anonymous reviewers, both of whom provided thoughtful and helpful feedback on extremely short notice. We would like to thank COMEDY (\textbf{CO}horts of \textbf{M}aarten Sap, \textbf{E}mma Strubell, \textbf{D}aniel Fried, and \textbf{Y}onatan Bisk) lab members for sharing insights and intuitions during initial discussions; Nupoor Gandhi, Jared Fernandez, Jeremiah Milbauer, Zhisong Zhang, and Josh Zhanson also gave constructive feedback on drafts and figures. We would like to acknowledge CMU Workhorse and TIR groups for providing compute resources for this work. Outside of CMU, we are appreciative of Mengzhou Xia for helping us reproduce structured pruning experiments using CoFi. This project is funded in part by DSO National Laboratories.

\bibliography{anthology,custom}

\begin{thebibliography}{77}
\expandafter\ifx\csname natexlab\endcsname\relax\def\natexlab#1{#1}\fi

\bibitem[{Ahia et~al.(2021)Ahia, Kreutzer, and
  Hooker}]{ahia-etal-2021-low-resource}
Orevaoghene Ahia, Julia Kreutzer, and Sara Hooker. 2021.
\newblock \href {https://doi.org/10.18653/v1/2021.findings-emnlp.282} {The
  low-resource double bind: An empirical study of pruning for low-resource
  machine translation}.
\newblock In \emph{Findings of the Association for Computational Linguistics:
  EMNLP 2021}, pages 3316--3333, Punta Cana, Dominican Republic. Association
  for Computational Linguistics.

\bibitem[{Ahmed and Wahed(2020)}]{nur2020dedemocratization}
Nur Ahmed and Muntasir Wahed. 2020.
\newblock \href {https://doi.org/10.48550/ARXIV.2010.15581} {The
  de-democratization of ai: Deep learning and the compute divide in artificial
  intelligence research}.

\bibitem[{Bahri et~al.(2022)Bahri, Mobahi, and Tay}]{bahri22acl}
Dara Bahri, Hossein Mobahi, and Yi~Tay. 2022.
\newblock \href {https://arxiv.org/abs/2110.08529} {Sharpness-aware
  minimization improves language model generalization}.
\newblock In \emph{Annual Conference of the Association for Computational
  Linguistics (ACL)}, Dublin, Ireland.

\bibitem[{Bai et~al.(2021)Bai, Zhang, Hou, Shang, Jin, Jiang, Liu, Lyu, and
  King}]{bai-etal-2021-binarybert}
Haoli Bai, Wei Zhang, Lu~Hou, Lifeng Shang, Jin Jin, Xin Jiang, Qun Liu,
  Michael Lyu, and Irwin King. 2021.
\newblock \href {https://doi.org/10.18653/v1/2021.acl-long.334}
  {{B}inary{BERT}: Pushing the limit of {BERT} quantization}.
\newblock In \emph{Proceedings of the 59th Annual Meeting of the Association
  for Computational Linguistics and the 11th International Joint Conference on
  Natural Language Processing (Volume 1: Long Papers)}, pages 4334--4348,
  Online. Association for Computational Linguistics.

\bibitem[{Bartoldson et~al.(2020)Bartoldson, Morcos, Barbu, and
  Erlebacher}]{bartoldson2020generalizationCompression}
Brian Bartoldson, Ari Morcos, Adrian Barbu, and Gordon Erlebacher. 2020.
\newblock The generalization-stability tradeoff in neural network pruning.
\newblock \emph{Advances in Neural Information Processing Systems},
  33:20852--20864.

\bibitem[{Bhandare et~al.(2019)Bhandare, Sripathi, Karkada, Menon, Choi, Datta,
  and Saletore}]{bhandare2019efficient}
Aishwarya Bhandare, Vamsi Sripathi, Deepthi Karkada, Vivek Menon, Sun Choi,
  Kushal Datta, and Vikram Saletore. 2019.
\newblock Efficient 8-bit quantization of transformer neural machine language
  translation model.
\newblock \emph{arXiv preprint arXiv:1906.00532}.

\bibitem[{Bisla et~al.(2022)Bisla, Wang, and
  Choromanska}]{pmlr-v151-bisla22a-sharpness-measures}
Devansh Bisla, Jing Wang, and Anna Choromanska. 2022.
\newblock \href {https://proceedings.mlr.press/v151/bisla22a.html} {Low-pass
  filtering sgd for recovering flat optima in the deep learning optimization
  landscape}.
\newblock In \emph{Proceedings of The 25th International Conference on
  Artificial Intelligence and Statistics}, volume 151 of \emph{Proceedings of
  Machine Learning Research}, pages 8299--8339. PMLR.

\bibitem[{Blalock et~al.(2020)Blalock, Ortiz, Frankle, and
  Guttag}]{blalock2020state}
Davis Blalock, Jose Javier~Gonzalez Ortiz, Jonathan Frankle, and John Guttag.
  2020.
\newblock What is the state of neural network pruning?
\newblock In \emph{Proceedings of Machine Learning and Systems (MLSys}.

\bibitem[{Brown et~al.(2020)Brown, Mann, Ryder, Subbiah, Kaplan, Dhariwal,
  Neelakantan, Shyam, Sastry, Askell, Agarwal, Herbert-Voss, Krueger, Henighan,
  Child, Ramesh, Ziegler, Wu, Winter, Hesse, Chen, Sigler, Litwin, Gray, Chess,
  Clark, Berner, McCandlish, Radford, Sutskever, and Amodei}]{brown2020gpt3}
Tom Brown, Benjamin Mann, Nick Ryder, Melanie Subbiah, Jared~D Kaplan, Prafulla
  Dhariwal, Arvind Neelakantan, Pranav Shyam, Girish Sastry, Amanda Askell,
  Sandhini Agarwal, Ariel Herbert-Voss, Gretchen Krueger, Tom Henighan, Rewon
  Child, Aditya Ramesh, Daniel Ziegler, Jeffrey Wu, Clemens Winter, Chris
  Hesse, Mark Chen, Eric Sigler, Mateusz Litwin, Scott Gray, Benjamin Chess,
  Jack Clark, Christopher Berner, Sam McCandlish, Alec Radford, Ilya Sutskever,
  and Dario Amodei. 2020.
\newblock \href
  {https://proceedings.neurips.cc/paper/2020/file/1457c0d6bfcb4967418bfb8ac142f64a-Paper.pdf}
  {Language models are few-shot learners}.
\newblock In \emph{Advances in Neural Information Processing Systems},
  volume~33, pages 1877--1901. Curran Associates, Inc.

\bibitem[{Buciluǎ et~al.(2006)Buciluǎ, Caruana, and
  Niculescu-Mizil}]{bucilua2006compression}
Cristian Buciluǎ, Rich Caruana, and Alexandru Niculescu-Mizil. 2006.
\newblock \href {https://doi.org/10.1145/1150402.1150464} {Model compression}.
\newblock In \emph{Proceedings of the 12th ACM SIGKDD International Conference
  on Knowledge Discovery and Data Mining}, KDD '06, page 535–541, New York,
  NY, USA. Association for Computing Machinery.

\bibitem[{Cer et~al.(2017)Cer, Diab, Agirre, Lopez-Gazpio, and
  Specia}]{cer2017semeval}
Daniel Cer, Mona Diab, Eneko Agirre, Inigo Lopez-Gazpio, and Lucia Specia.
  2017.
\newblock Semeval-2017 task 1: Semantic textual similarity-multilingual and
  cross-lingual focused evaluation.
\newblock \emph{arXiv preprint arXiv:1708.00055}.

\bibitem[{Chaudhari et~al.(2017)Chaudhari, Choromańska, Soatto, LeCun,
  Baldassi, Borgs, Chayes, Sagun, and Zecchina}]{Chaudhari2017EntropySGDBG}
Pratik Chaudhari, Anna Choromańska, Stefano Soatto, Yann LeCun, Carlo
  Baldassi, Christian Borgs, Jennifer~T. Chayes, Levent Sagun, and Riccardo
  Zecchina. 2017.
\newblock Entropy-sgd: Biasing gradient descent into wide valleys.
\newblock \emph{ArXiv}, abs/1611.01838.

\bibitem[{Chen et~al.(2020)Chen, Frankle, Chang, Liu, Zhang, Wang, and
  Carbin}]{chen2020BertLT}
Tianlong Chen, Jonathan Frankle, Shiyu Chang, Sijia Liu, Yang Zhang, Zhangyang
  Wang, and Michael Carbin. 2020.
\newblock \href
  {https://proceedings.neurips.cc/paper/2020/file/b6af2c9703f203a2794be03d443af2e3-Paper.pdf}
  {The lottery ticket hypothesis for pre-trained bert networks}.
\newblock In \emph{Advances in Neural Information Processing Systems},
  volume~33, pages 15834--15846. Curran Associates, Inc.

\bibitem[{Chen et~al.(2021)Chen, Cheng, Wang, Gan, Wang, and
  Liu}]{chen-etal-2021-earlybert}
Xiaohan Chen, Yu~Cheng, Shuohang Wang, Zhe Gan, Zhangyang Wang, and Jingjing
  Liu. 2021.
\newblock \href {https://doi.org/10.18653/v1/2021.acl-long.171} {{E}arly{BERT}:
  Efficient {BERT} training via early-bird lottery tickets}.
\newblock In \emph{Proceedings of the 59th Annual Meeting of the Association
  for Computational Linguistics and the 11th International Joint Conference on
  Natural Language Processing (Volume 1: Long Papers)}, pages 2195--2207,
  Online. Association for Computational Linguistics.

\bibitem[{Damian et~al.(2021)Damian, Ma, and Lee}]{damian2021label}
Alex Damian, Tengyu Ma, and Jason~D. Lee. 2021.
\newblock \href {https://openreview.net/forum?id=x2TMPhseWAW} {Label noise
  {SGD} provably prefers flat global minimizers}.
\newblock In \emph{Advances in Neural Information Processing Systems}.

\bibitem[{Devlin et~al.(2019)Devlin, Chang, Lee, and
  Toutanova}]{devlin-etal-2019-bert}
Jacob Devlin, Ming-Wei Chang, Kenton Lee, and Kristina Toutanova. 2019.
\newblock \href {https://doi.org/10.18653/v1/N19-1423} {{BERT}: Pre-training of
  deep bidirectional transformers for language understanding}.
\newblock In \emph{Proceedings of the 2019 Conference of the North {A}merican
  Chapter of the Association for Computational Linguistics: Human Language
  Technologies, Volume 1 (Long and Short Papers)}, pages 4171--4186,
  Minneapolis, Minnesota. Association for Computational Linguistics.

\bibitem[{Diffenderfer and Kailkhura(2021)}]{diffenderfer2021multiprize}
James Diffenderfer and Bhavya Kailkhura. 2021.
\newblock \href {https://openreview.net/forum?id=U_mat0b9iv} {Multi-prize
  lottery ticket hypothesis: Finding accurate binary neural networks by pruning
  a randomly weighted network}.
\newblock In \emph{International Conference on Learning Representations}.

\bibitem[{Dinh et~al.(2017)Dinh, Pascanu, Bengio, and
  Bengio}]{pmlr-v70-dinh17b}
Laurent Dinh, Razvan Pascanu, Samy Bengio, and Yoshua Bengio. 2017.
\newblock \href {https://proceedings.mlr.press/v70/dinh17b.html} {Sharp minima
  can generalize for deep nets}.
\newblock In \emph{Proceedings of the 34th International Conference on Machine
  Learning}, volume~70 of \emph{Proceedings of Machine Learning Research},
  pages 1019--1028. PMLR.

\bibitem[{Dodge et~al.(2022)Dodge, Prewitt, des Combes, Odmark, Schwartz,
  Strubell, Luccioni, Smith, DeCario, and Buchanan}]{dodge2022measuring}
Jesse Dodge, Taylor Prewitt, Remi~Tachet des Combes, Erika Odmark, Roy
  Schwartz, Emma Strubell, Alexandra~Sasha Luccioni, Noah~A. Smith, Nicole
  DeCario, and Will Buchanan. 2022.
\newblock Measuring the carbon intensity of ai in cloud instances.
\newblock In \emph{ACM Conference on Fairness, Accountability, and Transparency
  (ACM FAccT)}.

\bibitem[{Dolan and Brockett(2005)}]{dolan2005automatically}
Bill Dolan and Chris Brockett. 2005.
\newblock Automatically constructing a corpus of sentential paraphrases.
\newblock In \emph{Third International Workshop on Paraphrasing (IWP2005)}.

\bibitem[{Du et~al.(2022{\natexlab{a}})Du, Yan, Feng, Zhou, Zhen, Goh, and
  Tan}]{du2022efficientSAM}
Jiawei Du, Hanshu Yan, Jiashi Feng, Joey~Tianyi Zhou, Liangli Zhen, Rick
  Siow~Mong Goh, and Vincent Tan. 2022{\natexlab{a}}.
\newblock \href {https://openreview.net/forum?id=n0OeTdNRG0Q} {Efficient
  sharpness-aware minimization for improved training of neural networks}.
\newblock In \emph{International Conference on Learning Representations}.

\bibitem[{Du et~al.(2022{\natexlab{b}})Du, Zhou, Feng, Tan, and
  Zhou}]{sat4free}
Jiawei Du, Daquan Zhou, Jiashi Feng, Vincent Y.~F. Tan, and Joey~Tianyi Zhou.
  2022{\natexlab{b}}.
\newblock \href {https://doi.org/10.48550/ARXIV.2205.14083} {Sharpness-aware
  training for free}.

\bibitem[{Foret et~al.(2021)Foret, Kleiner, Mobahi, and
  Neyshabur}]{foret2021sharpnessaware}
Pierre Foret, Ariel Kleiner, Hossein Mobahi, and Behnam Neyshabur. 2021.
\newblock \href {https://openreview.net/forum?id=6Tm1mposlrM} {Sharpness-aware
  minimization for efficiently improving generalization}.
\newblock In \emph{International Conference on Learning Representations}.

\bibitem[{Frankle and Carbin(2019)}]{frankle2018lotteryticket}
Jonathan Frankle and Michael Carbin. 2019.
\newblock \href {https://openreview.net/forum?id=rJl-b3RcF7} {The lottery
  ticket hypothesis: Finding sparse, trainable neural networks}.
\newblock In \emph{International Conference on Learning Representations}.

\bibitem[{Frankle et~al.(2020)Frankle, Dziugaite, Roy, and
  Carbin}]{frankle2020linear}
Jonathan Frankle, Gintare~Karolina Dziugaite, Daniel~M. Roy, and Michael
  Carbin. 2020.
\newblock Linear mode connectivity and the lottery ticket hypothesis.
\newblock In \emph{Proceedings of the 37th International Conference on Machine
  Learning}, ICML'20. JMLR.org.

\bibitem[{Gholami et~al.(2021)Gholami, Kim, Dong, Yao, Mahoney, and
  Keutzer}]{gholami2021survey}
Amir Gholami, Sehoon Kim, Zhen Dong, Zhewei Yao, Michael~W. Mahoney, and Kurt
  Keutzer. 2021.
\newblock \href {http://arxiv.org/abs/2103.13630} {A survey of quantization
  methods for efficient neural network inference}.

\bibitem[{Gray and Neuhoff(1998)}]{gray1998quantization}
R.M. Gray and D.L. Neuhoff. 1998.
\newblock \href {https://doi.org/10.1109/18.720541} {Quantization}.
\newblock \emph{IEEE Transactions on Information Theory}, 44(6):2325--2383.

\bibitem[{Guo et~al.(2021)Guo, Rush, and Kim}]{guo-etal-2021-parameter}
Demi Guo, Alexander Rush, and Yoon Kim. 2021.
\newblock \href {https://doi.org/10.18653/v1/2021.acl-long.378}
  {Parameter-efficient transfer learning with diff pruning}.
\newblock In \emph{Proceedings of the 59th Annual Meeting of the Association
  for Computational Linguistics and the 11th International Joint Conference on
  Natural Language Processing (Volume 1: Long Papers)}, pages 4884--4896,
  Online. Association for Computational Linguistics.

\bibitem[{Han et~al.(2015)Han, Mao, and Dally}]{han2015StdPruning1}
Song Han, Huizi Mao, and William~J. Dally. 2015.
\newblock \href {https://doi.org/10.48550/ARXIV.1510.00149} {Deep compression:
  Compressing deep neural networks with pruning, trained quantization and
  huffman coding}.

\bibitem[{Hao et~al.(2019)Hao, Dong, Wei, and Xu}]{hao2019visualizing}
Yaru Hao, Li~Dong, Furu Wei, and Ke~Xu. 2019.
\newblock Visualizing and understanding the effectiveness of bert.
\newblock In \emph{Proceedings of the 2019 Conference on Empirical Methods in
  Natural Language Processing and the 9th International Joint Conference on
  Natural Language Processing (EMNLP-IJCNLP)}, pages 4134--4143.

\bibitem[{Hinton et~al.(2014)Hinton, Vinyals, and Dean}]{hinton2014distilling}
Geoffrey Hinton, Oriol Vinyals, and Jeff Dean. 2014.
\newblock \href {https://arxiv.org/abs/1503.02531} {Distilling the knowledge in
  a neural network}.
\newblock In \emph{NIPS 2014 Deep Learning Workshop}.

\bibitem[{Hochreiter and Schmidhuber(1997)}]{hochreiter1997flat}
Sepp Hochreiter and J{\"u}rgen Schmidhuber. 1997.
\newblock Flat minima.
\newblock \emph{Neural computation}, 9(1):1--42.

\bibitem[{Hooker et~al.(2020)Hooker, Moorosi, Clark, Bengio, and
  Denton}]{hooker2020characterising}
Sara Hooker, Nyalleng Moorosi, Gregory Clark, Samy Bengio, and Emily Denton.
  2020.
\newblock Characterising bias in compressed models.
\newblock In \emph{Fifth Workshop on Human Interpretability in Machine Learning
  (WHI)}.

\bibitem[{Izmailov et~al.(2018)Izmailov, Podoprikhin, Garipov, Vetrov, and
  Wilson}]{izmailov2018averaging}
Pavel Izmailov, Dmitrii Podoprikhin, Timur Garipov, Dmitry Vetrov, and
  Andrew~Gordon Wilson. 2018.
\newblock Averaging weights leads to wider optima and better generalization.
\newblock \emph{arXiv preprint arXiv:1803.05407}.

\bibitem[{Jacob et~al.(2018)Jacob, Kligys, Chen, Zhu, Tang, Howard, Adam, and
  Kalenichenko}]{Jacob2018QuantizationAT}
Benoit Jacob, Skirmantas Kligys, Bo~Chen, Menglong Zhu, Matthew Tang, Andrew~G.
  Howard, Hartwig Adam, and Dmitry Kalenichenko. 2018.
\newblock Quantization and training of neural networks for efficient
  integer-arithmetic-only inference.
\newblock \emph{2018 IEEE/CVF Conference on Computer Vision and Pattern
  Recognition}, pages 2704--2713.

\bibitem[{Jiang* et~al.(2020)Jiang*, Neyshabur*, Mobahi, Krishnan, and
  Bengio}]{Jiang*2020Fantastic}
Yiding Jiang*, Behnam Neyshabur*, Hossein Mobahi, Dilip Krishnan, and Samy
  Bengio. 2020.
\newblock \href {https://openreview.net/forum?id=SJgIPJBFvH} {Fantastic
  generalization measures and where to find them}.
\newblock In \emph{International Conference on Learning Representations}.

\bibitem[{Jiao et~al.(2020)Jiao, Yin, Shang, Jiang, Chen, Li, Wang, and
  Liu}]{jiao-etal-2020-tinybert}
Xiaoqi Jiao, Yichun Yin, Lifeng Shang, Xin Jiang, Xiao Chen, Linlin Li, Fang
  Wang, and Qun Liu. 2020.
\newblock \href {https://doi.org/10.18653/v1/2020.findings-emnlp.372}
  {{T}iny{BERT}: Distilling {BERT} for natural language understanding}.
\newblock In \emph{Findings of the Association for Computational Linguistics:
  EMNLP 2020}, pages 4163--4174, Online. Association for Computational
  Linguistics.

\bibitem[{Keskar et~al.(2017)Keskar, Mudigere, Nocedal, Smelyanskiy, and
  Tang}]{keskar2016large}
Nitish~Shirish Keskar, Dheevatsa Mudigere, Jorge Nocedal, Mikhail Smelyanskiy,
  and Ping Tak~Peter Tang. 2017.
\newblock \href {https://openreview.net/forum?id=H1oyRlYgg} {On large-batch
  training for deep learning: Generalization gap and sharp minima}.
\newblock In \emph{5th International Conference on Learning Representations,
  {ICLR} 2017, Toulon, France, April 24-26, 2017, Conference Track
  Proceedings}. OpenReview.net.

\bibitem[{Kim et~al.(2021)Kim, Gholami, Yao, Mahoney, and
  Keutzer}]{pmlr-v139-kim21d}
Sehoon Kim, Amir Gholami, Zhewei Yao, Michael~W. Mahoney, and Kurt Keutzer.
  2021.
\newblock \href {https://proceedings.mlr.press/v139/kim21d.html} {I-bert:
  Integer-only bert quantization}.
\newblock In \emph{Proceedings of the 38th International Conference on Machine
  Learning}, volume 139 of \emph{Proceedings of Machine Learning Research},
  pages 5506--5518. PMLR.

\bibitem[{Kingma and Ba(2014)}]{kingma2014adam}
Diederik~P Kingma and Jimmy Ba. 2014.
\newblock Adam: A method for stochastic optimization.
\newblock \emph{arXiv preprint arXiv:1412.6980}.

\bibitem[{Kuhn et~al.(2021)Kuhn, Lyle, Gomez, Rothfuss, and
  Gal}]{kuhn2021robustnessGeneralizationPruning}
Lorenz Kuhn, Clare Lyle, Aidan~N. Gomez, Jonas Rothfuss, and Yarin Gal. 2021.
\newblock \href {https://doi.org/10.48550/ARXIV.2103.06002} {Robustness to
  pruning predicts generalization in deep neural networks}.

\bibitem[{Kwon et~al.(2021)Kwon, Kim, Park, and Choi}]{pmlr-v139-kwon21b}
Jungmin Kwon, Jeongseop Kim, Hyunseo Park, and In~Kwon Choi. 2021.
\newblock \href {https://proceedings.mlr.press/v139/kwon21b.html} {Asam:
  Adaptive sharpness-aware minimization for scale-invariant learning of deep
  neural networks}.
\newblock In \emph{Proceedings of the 38th International Conference on Machine
  Learning}, volume 139 of \emph{Proceedings of Machine Learning Research},
  pages 5905--5914. PMLR.

\bibitem[{Lagunas et~al.(2021)Lagunas, Charlaix, Sanh, and
  Rush}]{lagunas-etal-2021-block}
Fran{\c{c}}ois Lagunas, Ella Charlaix, Victor Sanh, and Alexander Rush. 2021.
\newblock \href {https://doi.org/10.18653/v1/2021.emnlp-main.829} {Block
  pruning for faster transformers}.
\newblock In \emph{Proceedings of the 2021 Conference on Empirical Methods in
  Natural Language Processing}, pages 10619--10629, Online and Punta Cana,
  Dominican Republic. Association for Computational Linguistics.

\bibitem[{LeCun et~al.(1989)LeCun, Denker, and Solla}]{lecun1989optimal}
Yann LeCun, John Denker, and Sara Solla. 1989.
\newblock \href
  {https://proceedings.neurips.cc/paper/1989/file/6c9882bbac1c7093bd25041881277658-Paper.pdf}
  {Optimal brain damage}.
\newblock In \emph{Advances in Neural Information Processing Systems},
  volume~2. Morgan-Kaufmann.

\bibitem[{Li et~al.(2020)Li, Wallace, Shen, Lin, Keutzer, Klein, and
  Gonzalez}]{li2020train}
Zhuohan Li, Eric Wallace, Sheng Shen, Kevin Lin, Kurt Keutzer, Dan Klein, and
  Joey Gonzalez. 2020.
\newblock \href {http://proceedings.mlr.press/v119/li20m.html} {Train large,
  then compress: Rethinking model size for efficient training and inference of
  transformers}.
\newblock In \emph{ICML}, pages 5958--5968.

\bibitem[{Liang et~al.(2021)Liang, Zuo, Chen, Jiang, Liu, He, Zhao, and
  Chen}]{liang-etal-2021-super}
Chen Liang, Simiao Zuo, Minshuo Chen, Haoming Jiang, Xiaodong Liu, Pengcheng
  He, Tuo Zhao, and Weizhu Chen. 2021.
\newblock \href {https://doi.org/10.18653/v1/2021.acl-long.510} {Super tickets
  in pre-trained language models: From model compression to improving
  generalization}.
\newblock In \emph{Proceedings of the 59th Annual Meeting of the Association
  for Computational Linguistics and the 11th International Joint Conference on
  Natural Language Processing (Volume 1: Long Papers)}, pages 6524--6538,
  Online. Association for Computational Linguistics.

\bibitem[{Liebenwein et~al.(2021)Liebenwein, Baykal, Carter, Gifford, and
  Rus}]{liebenwein2021lost}
Lucas Liebenwein, Cenk Baykal, Brandon Carter, David Gifford, and Daniela Rus.
  2021.
\newblock \href
  {https://proceedings.mlsys.org/paper/2021/file/2a79ea27c279e471f4d180b08d62b00a-Paper.pdf}
  {Lost in pruning: The effects of pruning neural networks beyond test
  accuracy}.
\newblock In \emph{Proceedings of Machine Learning and Systems}, volume~3,
  pages 93--138.

\bibitem[{Liu et~al.(2019)Liu, Ott, Goyal, Du, Joshi, Chen, Levy, Lewis,
  Zettlemoyer, and Stoyanov}]{Liu2019RoBERTaAR}
Yinhan Liu, Myle Ott, Naman Goyal, Jingfei Du, Mandar Joshi, Danqi Chen, Omer
  Levy, Mike Lewis, Luke Zettlemoyer, and Veselin Stoyanov. 2019.
\newblock Roberta: A robustly optimized bert pretraining approach.
\newblock \emph{ArXiv}, abs/1907.11692.

\bibitem[{Louizos et~al.(2018)Louizos, Welling, and
  Kingma}]{louizos2018learning}
Christos Louizos, Max Welling, and Diederik~P. Kingma. 2018.
\newblock \href {https://openreview.net/forum?id=H1Y8hhg0b} {Learning sparse
  neural networks through l\_0 regularization}.
\newblock In \emph{International Conference on Learning Representations}.

\bibitem[{Mehta et~al.(2021)Mehta, Patil, Chandar, and
  Strubell}]{mehta2021empirical}
Sanket~Vaibhav Mehta, Darshan Patil, Sarath Chandar, and Emma Strubell. 2021.
\newblock \href {https://arxiv.org/abs/2112.09153} {An empirical investigation
  of the role of pre-training in lifelong learning}.
\newblock \emph{arXiv preprint arXiv:2112.09153}.

\bibitem[{Michel et~al.(2019)Michel, Levy, and Neubig}]{michel2019sixteen}
Paul Michel, Omer Levy, and Graham Neubig. 2019.
\newblock \href
  {https://proceedings.neurips.cc/paper/2019/file/2c601ad9d2ff9bc8b282670cdd54f69f-Paper.pdf}
  {Are sixteen heads really better than one?}
\newblock In \emph{Advances in Neural Information Processing Systems},
  volume~32. Curran Associates, Inc.

\bibitem[{Neyshabur et~al.(2020)Neyshabur, Sedghi, and
  Zhang}]{neyshabur2020what}
Behnam Neyshabur, Hanie Sedghi, and Chiyuan Zhang. 2020.
\newblock \href
  {https://proceedings.neurips.cc/paper/2020/file/0607f4c705595b911a4f3e7a127b44e0-Paper.pdf}
  {What is being transferred in transfer learning?}
\newblock In \emph{Advances in Neural Information Processing Systems},
  volume~33, pages 512--523. Curran Associates, Inc.

\bibitem[{Paul et~al.(2022)Paul, Chen, Larsen, Frankle, Ganguli, and
  Dziugaite}]{paul2022unmasking}
Mansheej Paul, Feng Chen, Brett~W. Larsen, Jonathan Frankle, Surya Ganguli, and
  Gintare~Karolina Dziugaite. 2022.
\newblock \href {https://doi.org/10.48550/ARXIV.2210.03044} {Unmasking the
  lottery ticket hypothesis: What's encoded in a winning ticket's mask?}

\bibitem[{Raffel et~al.(2020)Raffel, Shazeer, Roberts, Lee, Narang, Matena,
  Zhou, Li, and Liu}]{JMLR:t5raffel}
Colin Raffel, Noam Shazeer, Adam Roberts, Katherine Lee, Sharan Narang, Michael
  Matena, Yanqi Zhou, Wei Li, and Peter~J. Liu. 2020.
\newblock \href {http://jmlr.org/papers/v21/20-074.html} {Exploring the limits
  of transfer learning with a unified text-to-text transformer}.
\newblock \emph{Journal of Machine Learning Research}, 21(140):1--67.

\bibitem[{Rajpurkar et~al.(2016)Rajpurkar, Zhang, Lopyrev, and
  Liang}]{rajpurkar2016squad}
Pranav Rajpurkar, Jian Zhang, Konstantin Lopyrev, and Percy Liang. 2016.
\newblock Squad: 100,000+ questions for machine comprehension of text.
\newblock In \emph{Proceedings of the 2016 Conference on Empirical Methods in
  Natural Language Processing}, pages 2383--2392.

\bibitem[{Renda et~al.(2020{\natexlab{a}})Renda, Frankle, and
  Carbin}]{Renda2020ComparingStdPruning2rewinding}
Alex Renda, Jonathan Frankle, and Michael Carbin. 2020{\natexlab{a}}.
\newblock \href {https://openreview.net/forum?id=S1gSj0NKvB} {Comparing
  rewinding and fine-tuning in neural network pruning}.
\newblock In \emph{International Conference on Learning Representations}.

\bibitem[{Renda et~al.(2020{\natexlab{b}})Renda, Frankle, and
  Carbin}]{Renda2020Comparing}
Alex Renda, Jonathan Frankle, and Michael Carbin. 2020{\natexlab{b}}.
\newblock \href {https://openreview.net/forum?id=S1gSj0NKvB} {Comparing
  rewinding and fine-tuning in neural network pruning}.
\newblock In \emph{International Conference on Learning Representations}.

\bibitem[{Ribeiro et~al.(2020)Ribeiro, Wu, Guestrin, and
  Singh}]{ribeiro-etal-2020-beyond}
Marco~Tulio Ribeiro, Tongshuang Wu, Carlos Guestrin, and Sameer Singh. 2020.
\newblock \href {https://doi.org/10.18653/v1/2020.acl-main.442} {Beyond
  accuracy: Behavioral testing of {NLP} models with {C}heck{L}ist}.
\newblock In \emph{Proceedings of the 58th Annual Meeting of the Association
  for Computational Linguistics}, pages 4902--4912, Online. Association for
  Computational Linguistics.

\bibitem[{Sanh et~al.(2019)Sanh, Debut, Chaumond, and
  Wolf}]{sanh2019distilBERT}
Victor Sanh, Lysandre Debut, Julien Chaumond, and Thomas Wolf. 2019.
\newblock \href {https://doi.org/10.48550/ARXIV.1910.01108} {Distilbert, a
  distilled version of bert: smaller, faster, cheaper and lighter}.

\bibitem[{Sanh et~al.(2020)Sanh, Wolf, and Rush}]{sanh2020movement}
Victor Sanh, Thomas Wolf, and Alexander Rush. 2020.
\newblock \href
  {https://proceedings.neurips.cc/paper/2020/file/eae15aabaa768ae4a5993a8a4f4fa6e4-Paper.pdf}
  {Movement pruning: Adaptive sparsity by fine-tuning}.
\newblock In \emph{Advances in Neural Information Processing Systems},
  volume~33, pages 20378--20389. Curran Associates, Inc.

\bibitem[{Shen et~al.(2020)Shen, Dong, Ye, Ma, Yao, Gholami, Mahoney, and
  Keutzer}]{shen2020qbert}
Sheng Shen, Zhen Dong, Jiayu Ye, Linjian Ma, Zhewei Yao, Amir Gholami,
  Michael~W. Mahoney, and Kurt Keutzer. 2020.
\newblock \href {https://doi.org/10.1609/aaai.v34i05.6409} {Q-bert: Hessian
  based ultra low precision quantization of bert}.
\newblock \emph{Proceedings of the AAAI Conference on Artificial Intelligence},
  34(05):8815--8821.

\bibitem[{Socher et~al.(2013)Socher, Perelygin, Wu, Chuang, Manning, Ng, and
  Potts}]{socher2013recursive}
Richard Socher, Alex Perelygin, Jean Wu, Jason Chuang, Christopher~D Manning,
  Andrew~Y Ng, and Christopher Potts. 2013.
\newblock Recursive deep models for semantic compositionality over a sentiment
  treebank.
\newblock In \emph{Proceedings of the 2013 conference on empirical methods in
  natural language processing}, pages 1631--1642.

\bibitem[{Strubell et~al.(2019)Strubell, Ganesh, and
  McCallum}]{strubell-etal-2019-energy}
Emma Strubell, Ananya Ganesh, and Andrew McCallum. 2019.
\newblock \href {https://doi.org/10.18653/v1/P19-1355} {Energy and policy
  considerations for deep learning in {NLP}}.
\newblock In \emph{Proceedings of the 57th Annual Meeting of the Association
  for Computational Linguistics}, pages 3645--3650, Florence, Italy.
  Association for Computational Linguistics.

\bibitem[{Sun et~al.(2019)Sun, Cheng, Gan, and Liu}]{sun-etal-2019-patient}
Siqi Sun, Yu~Cheng, Zhe Gan, and Jingjing Liu. 2019.
\newblock \href {https://doi.org/10.18653/v1/D19-1441} {Patient knowledge
  distillation for {BERT} model compression}.
\newblock In \emph{Proceedings of the 2019 Conference on Empirical Methods in
  Natural Language Processing and the 9th International Joint Conference on
  Natural Language Processing (EMNLP-IJCNLP)}, pages 4323--4332, Hong Kong,
  China. Association for Computational Linguistics.

\bibitem[{Sun et~al.(2020)Sun, Yu, Song, Liu, Yang, and
  Zhou}]{sun-etal-2020-mobilebert}
Zhiqing Sun, Hongkun Yu, Xiaodan Song, Renjie Liu, Yiming Yang, and Denny Zhou.
  2020.
\newblock \href {https://doi.org/10.18653/v1/2020.acl-main.195}
  {{M}obile{BERT}: a compact task-agnostic {BERT} for resource-limited
  devices}.
\newblock In \emph{Proceedings of the 58th Annual Meeting of the Association
  for Computational Linguistics}, pages 2158--2170, Online. Association for
  Computational Linguistics.

\bibitem[{Vanhoucke et~al.(2011)Vanhoucke, Senior, and
  Mao}]{vanhoucke2011improving}
Vincent Vanhoucke, Andrew Senior, and Mark~Z. Mao. 2011.
\newblock Improving the speed of neural networks on cpus.
\newblock In \emph{Deep Learning and Unsupervised Feature Learning Workshop,
  NIPS 2011}.

\bibitem[{Wang et~al.(2018)Wang, Singh, Michael, Hill, Levy, and
  Bowman}]{wang2018glue}
Alex Wang, Amanpreet Singh, Julian Michael, Felix Hill, Omer Levy, and Samuel~R
  Bowman. 2018.
\newblock \href {https://arxiv.org/abs/1804.07461} {Glue: A multi-task
  benchmark and analysis platform for natural language understanding}.
\newblock \emph{arXiv preprint arXiv:1804.07461}.

\bibitem[{Warstadt et~al.(2019)Warstadt, Singh, and
  Bowman}]{warstadt2019neural}
Alex Warstadt, Amanpreet Singh, and Samuel~R Bowman. 2019.
\newblock Neural network acceptability judgments.
\newblock \emph{Transactions of the Association for Computational Linguistics},
  7:625--641.

\bibitem[{Williams et~al.(2018)Williams, Nangia, and
  Bowman}]{williams2018broad}
Adina Williams, Nikita Nangia, and Samuel Bowman. 2018.
\newblock A broad-coverage challenge corpus for sentence understanding through
  inference.
\newblock In \emph{Proceedings of the 2018 Conference of the North American
  Chapter of the Association for Computational Linguistics: Human Language
  Technologies, Volume 1 (Long Papers)}, pages 1112--1122.

\bibitem[{Wolf et~al.(2019)Wolf, Debut, Sanh, Chaumond, Delangue, Moi, Cistac,
  Rault, Louf, Funtowicz et~al.}]{wolf2019huggingface}
Thomas Wolf, Lysandre Debut, Victor Sanh, Julien Chaumond, Clement Delangue,
  Anthony Moi, Pierric Cistac, Tim Rault, R{\'e}mi Louf, Morgan Funtowicz,
  et~al. 2019.
\newblock Huggingface's transformers: State-of-the-art natural language
  processing.
\newblock \emph{arXiv preprint arXiv:1910.03771}.

\bibitem[{Wu et~al.(2017)Wu, Zhu, and E}]{wu2017understanding}
Lei Wu, Zhanxing Zhu, and Weinan E. 2017.
\newblock Towards understanding generalization of deep learning: Perspective of
  loss landscapes.
\newblock In \emph{ICML Workshop on Principled Approaches to Deep Learning
  (PADL)}.

\bibitem[{Xia et~al.(2022)Xia, Zhong, and Chen}]{xia2022structuredCoFi}
Mengzhou Xia, Zexuan Zhong, and Danqi Chen. 2022.
\newblock Structured pruning learns compact and accurate models.
\newblock In \emph{Association for Computational Linguistics (ACL)}.

\bibitem[{Xu et~al.(2021)Xu, Zhou, Ge, Xu, McAuley, and
  Wei}]{xu-etal-2021-beyond}
Canwen Xu, Wangchunshu Zhou, Tao Ge, Ke~Xu, Julian McAuley, and Furu Wei. 2021.
\newblock \href {https://doi.org/10.18653/v1/2021.emnlp-main.832} {Beyond
  preserved accuracy: Evaluating loyalty and robustness of {BERT} compression}.
\newblock In \emph{Proceedings of the 2021 Conference on Empirical Methods in
  Natural Language Processing}, pages 10653--10659, Online and Punta Cana,
  Dominican Republic. Association for Computational Linguistics.

\bibitem[{Yu(2020)}]{yu2020algorithmic}
Peter~K. Yu. 2020.
\newblock The algorithmic divide and equality in the age of artificial
  intelligence.
\newblock \emph{Florida Law Review}, 72:331--89.

\bibitem[{Zafrir et~al.(2019)Zafrir, Boudoukh, Izsak, and
  Wasserblat}]{Zafrir2019Q8BERTQ8}
Ofir Zafrir, Guy Boudoukh, Peter Izsak, and Moshe Wasserblat. 2019.
\newblock Q8bert: Quantized 8bit bert.
\newblock \emph{2019 Fifth Workshop on Energy Efficient Machine Learning and
  Cognitive Computing - NeurIPS Edition (EMC2-NIPS)}, pages 36--39.

\bibitem[{Zhou et~al.(2020)Zhou, Gu, and Neubig}]{Zhou2020Understanding}
Chunting Zhou, Jiatao Gu, and Graham Neubig. 2020.
\newblock \href {https://openreview.net/forum?id=BygFVAEKDH} {Understanding
  knowledge distillation in non-autoregressive machine translation}.
\newblock In \emph{International Conference on Learning Representations}.

\bibitem[{Zhou et~al.(2019)Zhou, Veitch, Austern, Adams, and
  Orbanz}]{zhou2018nonvacuous}
Wenda Zhou, Victor Veitch, Morgane Austern, Ryan~P. Adams, and Peter Orbanz.
  2019.
\newblock \href {https://openreview.net/forum?id=BJgqqsAct7} {Non-vacuous
  generalization bounds at the imagenet scale: a {PAC}-bayesian compression
  approach}.
\newblock In \emph{International Conference on Learning Representations}.

\end{thebibliography}
\bibliographystyle{acl_natbib}

\clearpage\newpage
\appendix

\section{Appendix}
\label{sec:appendix}

\subsection{Extended Related Work}
\label{sec:append_relatedwork}

\paragraph{Model compression in NLP}
Approaches for model compression aim to replicate the end-task performance of a large, accurate model while requiring fewer parameters and floating-point operations, and many approaches for model compression have been successfully applied to tasks in NLP. Specific model compression techniques include \textit{pruning}, where individual model parameters (unstructured pruning) or entire weight matrices (structured pruning) are removed entirely \citep{lecun1989optimal, blalock2020state, sanh2020movement, lagunas-etal-2021-block}, \textit{quantization}, where model weights, activations, gradients, and/or add-multiply accumulators are reduced in precision from 32-bit floating point representations to floating or fixed-point representations as low as one or two bits \citep{gray1998quantization, vanhoucke2011improving, gholami2021survey}, and \textit{knowledge distillation} where a smaller model is trained to replicate the predictions, and often intermediate embedded representations, of a larger model \citep{bucilua2006compression, hinton2014distilling, sanh2019distilBERT, jiao-etal-2020-tinybert}.

Unstructured pruning can achieve some of the highest sparsity levels using various criteria and schedules for determining which parameters to prune %
\cite{frankle2018lotteryticket, chen2020BertLT, sanh2020movement,guo-etal-2021-parameter}, though sparsity patterns resulting from unstructured pruning often do not result in latency reduction on modern accelerator hardware.
Work in structured pruning has explored removing entire parameter matrices such as self-attention heads, hidden units, and entire layers \citep{michel2019sixteen, lagunas-etal-2021-block, xia2022structuredCoFi}, with basic underlying hardware constraints in mind. Pruning is often combined with a distillation objective, which provides complementary gains, likely by reducing complexity of the dataset \citep{Zhou2020Understanding}.

Distillation is a prominent, practical method for compression that is widely used in NLP. Work on distillation in NLP has focused largely on the task-agnostic setting of compressing general-purpose pre-trained models such as BERT \citep{sanh2019distilBERT, sun-etal-2019-patient, sun-etal-2020-mobilebert} but task-specific distillation has also been reported to work well \citep{jiao-etal-2020-tinybert}. %

Approaches for model quantization can be categorized into post-training quantization, where general-purpose models are quantized at test-time \cite{Jacob2018QuantizationAT, bhandare2019efficient,  pmlr-v139-kim21d}, and quantization-aware training, where models incorporate simulated quantization error during training in order to learn more quantizable parameters \citep{Zafrir2019Q8BERTQ8,bai-etal-2021-binarybert}. Quantization-aware training tends to lead to higher accuracy quantized inference, but post-training quantization can be applied on-the-fly to any model at inference time.

In this work we experiment with a variety of compression methods including structured and unstructured pruning \citep{xia2022structuredCoFi, chen2020BertLT}, and out-of-the-box \textsc{Int8} post-training dynamic quantization,\footnote{\protect{\url{https://pytorch.org/tutorials/recipes/recipes/dynamic_quantization.html}}} highlighting both practical (structured pruning, quantization) and theoretical (unstructured pruning) findings. While we do not experiment with distillation directly, our chosen structured pruning method also incorporates a distillation objective.

\paragraph{Learning compressible models}
Most closely related to our work are methods for learning compressible models, and the study of what makes models more compressible. Quantization-aware training is an example of such training for compressibility, and training for sparsity using $\ell_0$ regularization \citep{louizos2018learning} is a parallel method for pruning.

Learning sparse models from scratch has proven difficult, despite the fact that deep neural networks are vastly over-parameterized. \citet{frankle2018lotteryticket} formalized the \textit{Lottery Ticket Hypothesis}, which posits that large, overparameterized neural network models contain sparse subnetworks, or \textit{winning tickets}, that can be trained from scratch (or close to it; see  \citet{frankle2020linear}) to match the end-task performance of the full model. This influential work has spurred much research into better understanding neural network models, including pre-trained language models, from the perspective of winning tickets \citep{chen2020BertLT, Renda2020Comparing, diffenderfer2021multiprize} and how to leverage winning tickets to perform better model compression \citep{chen-etal-2021-earlybert,liang-etal-2021-super}. \citet{li2020train} showed that larger pre-trained language models are more compressible than smaller ones, which they hypothesize is related to larger models being more likely to contain winning tickets.

\paragraph{Flat minima in neural networks.}
\citet{hochreiter1997flat} were among the first to discuss the relationship between flat basins in the loss landscape and generalization in neural networks, defining a flat minimum as ``a large connected region in weight space where the error remains approximately constant.'' %
We use the $\epsilon$-sharpness definition of \citet{keskar2016large} which defines sharpness as maximum loss within a neighborhood bounded by $\epsilon$. Others have used Hessian-based measures to identify high minima with high curvature \citep{Chaudhari2017EntropySGDBG}. Note that care is needed when making inferences based on current measurements of sharpness, which is an active area of research; it has been shown that flat minima defined in this way can be rescaled to sharp minima (and still generalize) \citep{pmlr-v70-dinh17b}.

Most previous results related to flat minima have focused on generalizability \citep{hao2019visualizing, neyshabur2020what}. A common explanation for the good generalization of models converged to flat minima is that flatter models are less complex \citep{wu2017understanding}. 

Flat minima may be particularly of interest in NLP, where a pretrain-then-finetune paradigm is often employed to leverage general representations learned from an extensive pre-training process during a much shorter fine-tuning process on a more specific end task. Indeed, pre-training provides a flat prior that can provide benefits in the contexts of lifelong learning \citep{mehta2021empirical} and generalization \citep{hao2019visualizing, bahri22acl}.

\begin{table*}[h]
\small
    \centering
    \begin{tabular}{lc|r r r r r r r r}%
    \toprule
         epsilon ($\epsilon$) & Dataset & MNLI & QQP & STS-B & QNLI & MRPC & RTE & SST-2 & CoLA\\
         \midrule
         $5\times 10^{-3}$ & Adam & $28.3_{3.6}$ & $34.7_{5.0}$ & $160.9_{34.3}$ & $30.1_{6.0}$ & $50.8_{24.8}$ & $49.0_{5.6}$ & $29.9_{10.3}$ & $38.3_{3.7}$ \\
         & SAM & $14.2_{0.9}$ & $9.3_{1.7}$ & $45.8_{1.3}$ & $17.8_{3.4}$ & $40.4_{5.3}$ & $28.4_{8.7}$ & $13.7_{2.3}$ & $29.4_{9.8}$  \\
         \midrule
        $1\times 10^{-3}$ & Adam & $5.0_{0.6}$ & $6.5_{0.9}$ & $11.8_{3.1}$ & $6.6_{2.5}$ & $6.1_{1.0}$ & $11.9_{3.0}$ & $4.5_{0.6}$ & $9.5_{2.2}$  \\
         & SAM & $2.6_{0.2}$ & $1.9_{0.3}$ & $4.5_{0.4}$ & $3.5_{1.3}$ & $7.0_{0.7}$ & $4.3_{2.4}$ & $2.2_{0.1}$ & $6.8_{2.8}$  \\
         \midrule
         $5\times 10^{-4}$ & Adam & $2.3_{0.2}$ & $3.4_{0.2}$ & $4.3_{1.2}$ & $3.2_{1.5}$ & $2.8_{0.3}$ & $6.5_{2.1}$ & $2.3_{0.4}$ & $5.6_{2.0}$ \\
         & SAM & $1.3_{0.1}$ & $0.9_{0.1}$ & $1.9_{0.3}$ & $1.5_{0.3}$ & $3.2_{0.6}$ & $2.1_{1.1}$ & $1.0_{0.1}$ & $3.4_{1.4}$ \\
    \bottomrule
    \end{tabular}
    \caption{Evaluating sharpness metric for vanilla Adam and SAM optimized models at full size (i.e., no compression). We observe that SAM optimized models have significantly lower sharpness values (lower corresponds to flatter minima) compared to vanilla Adam. These results provide quantitative evidence that SAM indeed leads to flatter loss basins.}
    \label{tab:sharpnessMetricSparsity0}
\end{table*}

\begin{table*}[h]
\small
    \centering
    \begin{tabular}{lc|r r r r r r r r}%
    \toprule
         epsilon ($\epsilon$) & Dataset & MNLI & QQP & STS-B & QNLI & MRPC & RTE & SST-2 & CoLA \\
         & Sparsity & $70\%$ & $90\%$ & $50\%$ & $70\%$ & $50\%$ & $60\%$  & $60\%$ & $50\%$ \\
         \midrule
         $5\times 10^{-3}$ & Adam & $56.4_{1.0}$ & $137.9_{40.4}$ & $232.5_{30.0}$ & $23.8_{3.9}$ & $42.3_{26.9}$ & $65.7_{8.8}$ & $85.8_{33.8}$ & $33.8_{2.7}$ \\
         & SAM & $42.4_{7.9}$ & $65.3_{1.9}$ & $184.0_{7.3}$ & $17.2_{6.0}$ & $47.3_{18.0}$ & $50.8_{6.1}$ & $26.9_{7.7}$ & $23.9_{4.5}$ \\
         & SWA & $37.3_{9.9}$ & $34.7_{1.2}$ & $81.3_{22.3}$ & $33.5_{7.5}$ & $31.6_{6.6}$ & $42.1_{12.3}$ & $23.7_{10.3}$ & $26.5_{9.5}$ \\
         \midrule
        $1\times 10^{-3}$ & Adam & $6.6_{0.9}$ & $5.8_{1.0}$ & $20.2_{6.1}$ & $5.9_{2.2}$ & $8.6_{2.2}$ & $16.4_{4.0}$ & $5.8_{0.7}$ & $6.2_{1.4}$ \\
         & SAM & $3.1_{0.4}$ & $5.4_{0.6}$ & $13.8_{1.2}$ & $3.2_{1.1}$ & $8.3_{1.6}$ & $12.0_{3.8}$ & $4.1_{0.8}$ & $4.2_{0.3}$ \\
         & SWA & $10.7_{1.6}$ & $5.2_{0.6}$ & $5.4_{1.1}$ & $8.9_{0.2}$ & $12.4_{0.8}$ & $11.8_{4.1}$ & $5.2_{0.7}$ & $7.9_{2.2}$ \\
         \midrule
         $5\times 10^{-4}$ & Adam & $3.3_{0.2}$ & $2.3_{0.3}$ & $7.1_{2.5}$ & $2.1_{0.6}$ & $5.8_{0.5}$ & $7.9_{1.1}$ & $3.6_{0.9}$ & $3.9_{0.6}$ \\
         & SAM & $1.1_{0.3}$ & $2.0_{0.6}$ & $5.1_{0.6}$ & $1.2_{0.0}$ & $3.6_{0.6}$ & $5.9_{2.0}$ & $1.9_{0.4}$ & $2.3_{0.4}$ \\
         & SWA & $4.4_{0.5}$ & $2.9_{0.0}$ & $2.2_{0.4}$ & $4.7_{0.1}$ & $5.9_{0.7}$ & $6.0_{2.0}$ & $2.5_{0.5}$ & $3.4_{0.2}$ \\
    \bottomrule
    \end{tabular}
    \caption{Evaluating sharpness metric for vanilla Adam, SAM and SWA optimized models at \citet{chen2020BertLT}'s reference sparsities. In general we observe that SAM and SWA optimized models have lower sharpness values (lower corresponds to flatter minima) compared to vanilla Adam at various sparsity levels across different tasks (all results are averaged over $3$ runs).}
    \label{tab:sharpnessMetricSparsityRef}
\end{table*}

\subsection{Flat Minima: Additional details}
\label{sec:sharpnessmetric}

\paragraph{Stochastic Weight Averaging (SWA).}
We consider last $50\%$ of the model checkpoints for equal averaging. Specifically, for RTE, MRPC, STS-B, and CoLA we fine-tune for $10$ epochs and average $5$ checkpoints from epochs $6$ to $10$. For SST-2, QNLI, QQP, and MNLI we fine-tune for $3$ epochs and retain checkpoints after every $0.5$ epochs and equal average checkpoints after $2, 2.5, 3$ epochs. \citet{izmailov2018averaging} suggests modified learning rate scheduler like cyclical or constant so that towards the later stage of training, the underlying optimizer explores diverse solutions and average over them would lead to flatter solution. To simulate this behavior, for all our SWA experiments, we set our initial learning rate to a high value of $8e-5$ and linearly decay it to $0$ with no warmup steps.
\paragraph{Evaluating sharpness.} \citet{keskar2016large} propose a computationally feasible metric for measuring the sharpness of a minimizer over an $\epsilon$-neighborhood in the loss landscape.
We report sharpness metrics (lower value means flatter low loss region) in Tables~\ref{tab:sharpnessMetricSparsity0} and \ref{tab:sharpnessMetricSparsityRef}. Overall we see that SAM and SWA optimized models have significantly smaller sharpness values (or flatter low loss regions) as compared to vanilla Adam optimizer, thus, providing convincing evidence that these methods indeed find flatter solutions.
\begin{figure*}[h!]
    \centering
    \begin{tabular}{ll}
    \includegraphics[width=0.4\linewidth]{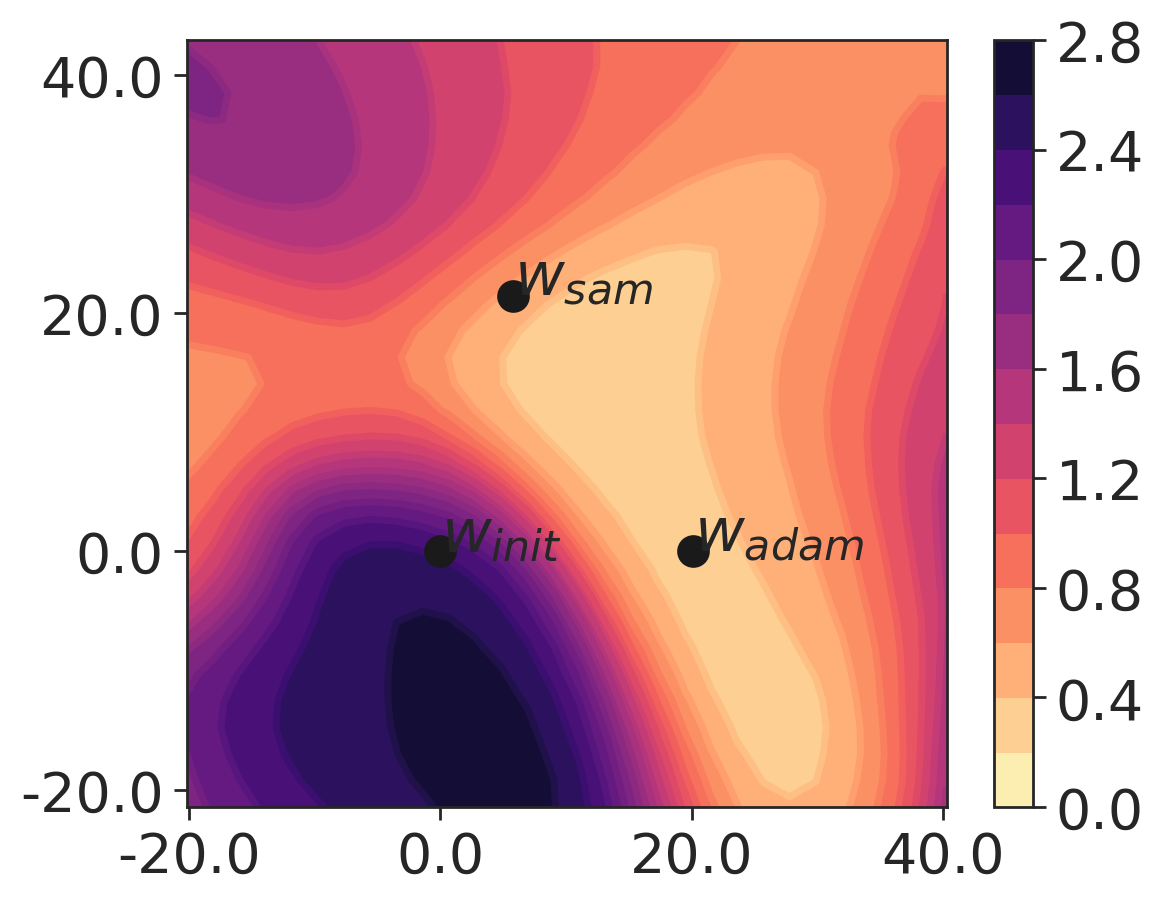} & 
    \includegraphics[width=0.4\linewidth]{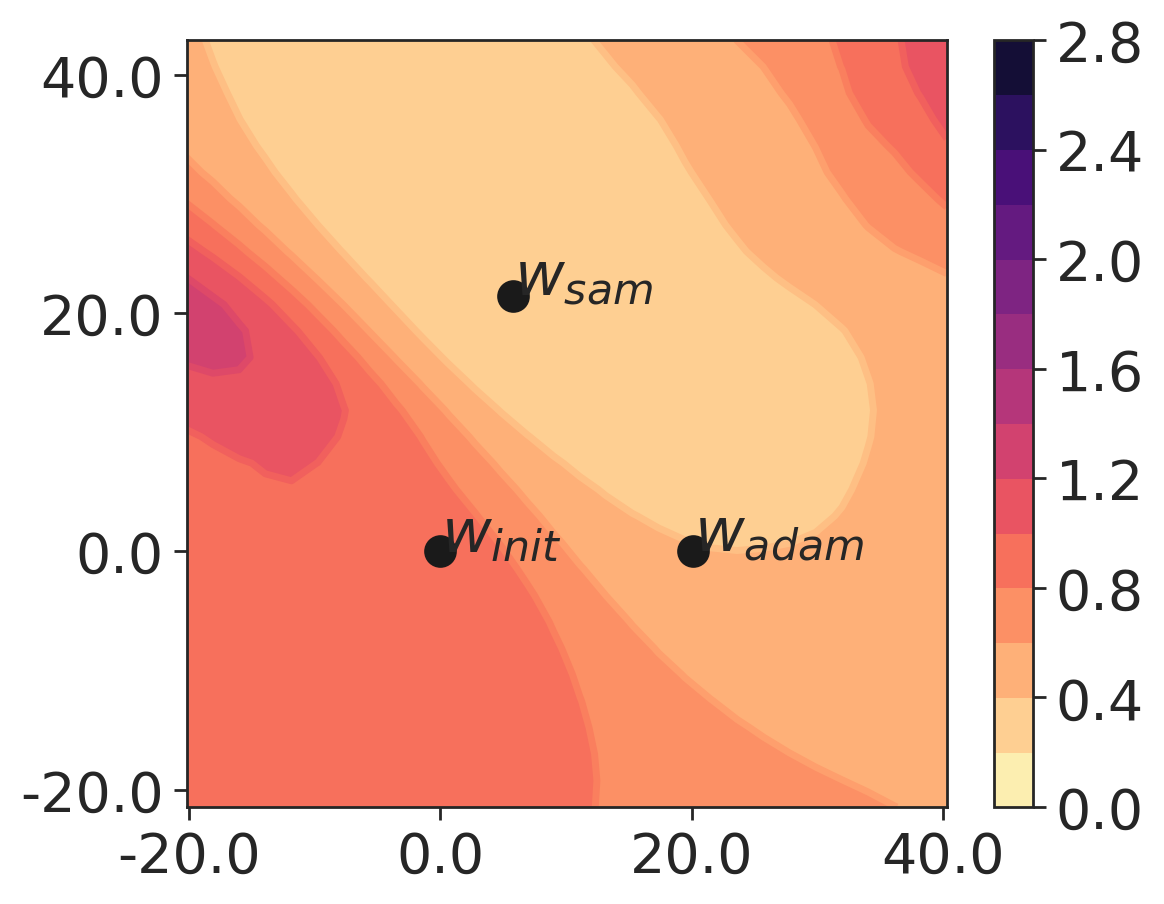}
    \end{tabular}
    
    \caption{Visualization of loss contours for QQP on BERT$_{base}$ models finetuned on the task using Adam and SAM optimizers ($w_{adam}$ and $w_{sam}$), as well as the pre-trained BERT base initialization ($w_{init}$). On the left, all models are fitted with the linear classifier head originally trained with $w_{adam}$, while on the right side models are fitted with the linear classifier head originally trained with $w_{sam}$. The SAM-optimized model sits in a noticeably flatter, wider basin than the Adam-optimized model when both are fitted with their respective classifier heads. These results provide qualitative evidence that SAM indeed leads to flatter loss basins.}
    \label{fig:sam_gives_flatter}
\end{figure*}

\paragraph{Loss contours.}
In Figure~\ref{fig:sam_gives_flatter}, we visualize contour plots of the loss landscape for the QQP task to qualitatively compare the sharpness of the solutions that Adam- and SAM-optimized BERT$_{base}$ models find. We observe that the SAM-optimized model sits in a noticeably flatter, wider basin than the Adam-optimized model when both are fitted with their respective classifier heads. These analyses verify that SAM indeed leads to flatter minima in comparison to Adam.  %

\subsection{Non-Iterative Unstructured Magnitude Pruning}
\label{subsec:noniter}

\begin{figure}[ht!]
    \centering
    \includegraphics[scale=0.55]{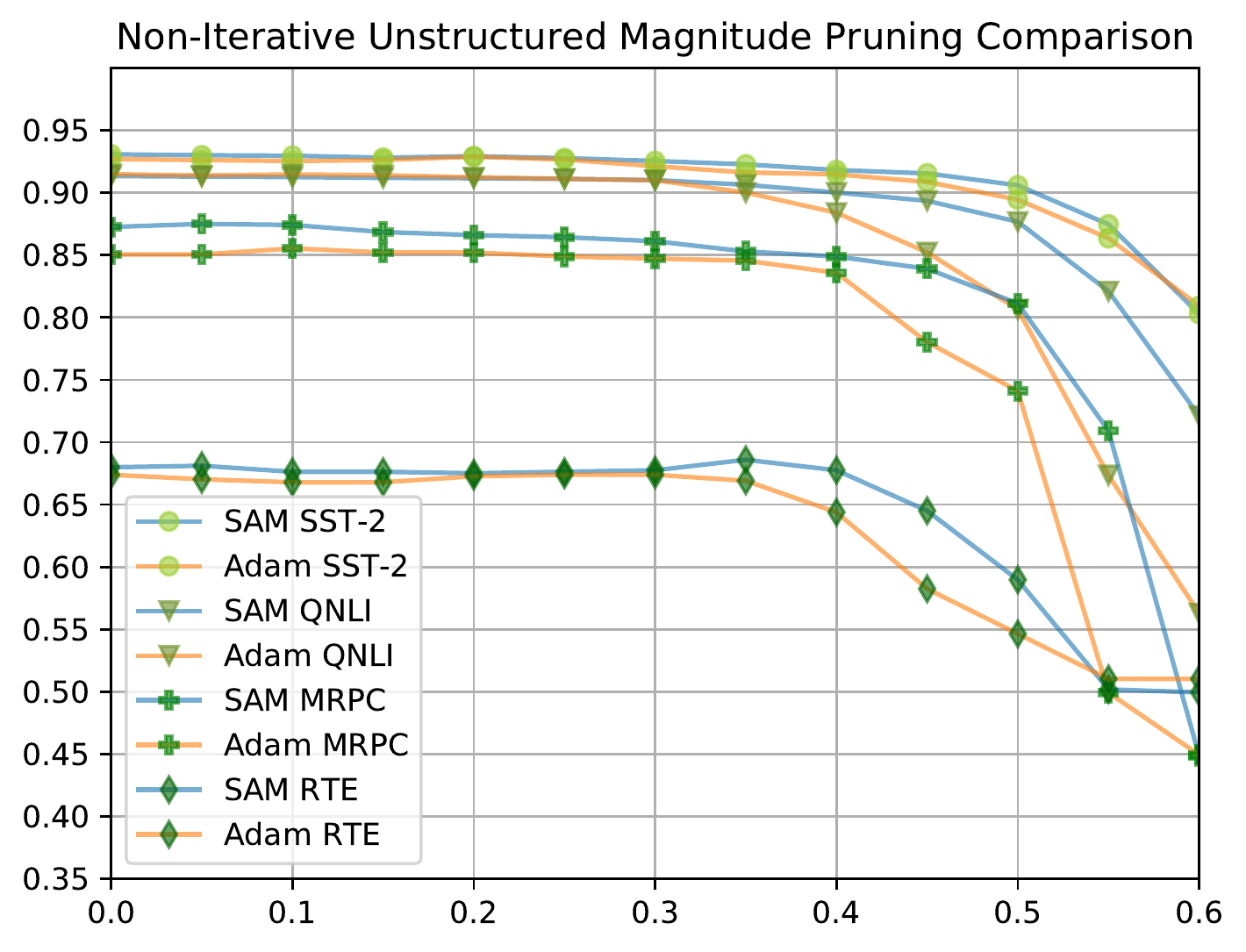}
    \caption{SAM- and Adam-optimized models evaluated directly after pruning a proportion of parameters from the full-sized model. Models fine-tuned with SAM hold up better to non-iterative magnitude pruning as well.}
    \label{fig:noniter_plot}
\end{figure}

In a preliminary analysis, we subject full-size fine-tuned BERT$_{base}$ models to one-shot unstructured magnitude pruning and evaluate on the same task \textit{without any subsequent training}. Figure~\ref{fig:noniter_plot} displays development set accuracies at sparsity levels of increments of $5\%$, up to $60\%$ of prunable parameters masked to $0$. Accuracy values are plotted at averages over $n=3$ seeds for each of the sparsity levels and GLUE tasks displayed (SST-2, QNLI, MRPC, RTE). Interestingly, even under this non-iterative pruning setting, performance does not drop off noticeably in either SAM or Adam-optimized models until at least around $30\%$ sparsity for the GLUE tasks displayed. Similarly to all iterative pruning settings we explore, models optimized with SAM retain full model size accuracies at higher sparsity levels than their Adam-optimized counterparts. The SAM-optimized RTE models at $35-40\%$ sparsity have \textit{higher} accuracy than the full-sized uncompressed model, reminiscent of the pattern we observe in Figure~\ref{fig:rte_plot}, albeit at lower sparsity levels. 

\subsection{Iterative Magnitude Pruning Reproducibility and Hyperparameters}
\label{subsec:imp_reprod}
Following \citet{chen2020BertLT}, we use a maximum sequence length of $128$, batch size of $32$, learning rate to $2e-5$, and linear decay of learning rate from initial value to zero with no warmup period. For tasks with smaller datasets (RTE, MRPC, CoLA, STS-B), we fine-tune models for $10$ epochs, evaluate them after every epoch and retain the checkpoint yielding best task-specific performance on the hold-out validation set (whereas \citet{chen2020BertLT} finetune for only $3$ for all tasks). For tasks with comparatively larger datasets (MNLI, QQP, QNLI, SST-2), we fine-tune models for $3$ epochs. %
We set Adam's weight decay to $\epsilon=0$ in order to remove the potential confound of regularization on models' amenability to magnitude pruning. This differs from \citet{chen2020BertLT}'s $\epsilon=1\times 10^{-8}$, but we observed that our differences do not systematically affect the trends originally reported other than to improve full-size model performance and allow for a fairer comparison between SAM and Adam optimizers. In particular, training for only $3$ epochs on the smaller tasks with our SAM optimizer does not allow models to converge.

\subsection{Comparison with BERT Lottery Ticket Hypothesis Numbers}
\label{subsec:imp_devset}
We present development set numbers on GLUE and SQuAD tasks in Table \ref{tab:recreateBERTLT_DEV}. For comparison, we include the reference (\textit{Ref}) metrics reported by \citet{chen2020BertLT} at their reported winning ticket sparsity levels.

\begin{table*}[h]
\scriptsize
    \centering
    \begin{tabular}{m{0.01\textwidth} r|c c c c c c c c c}%
    \toprule
         & Dataset & MNLI & QQP & STS-B & QNLI & MRPC & RTE & SST-2 & CoLA & SQuAD \\
         & Sparsity & $70\%$ & $90\%$ & $50\%$ & $70\%$ & $50\%$ & $60\%$  & $60\%$ & $50\%$ & $40\%$ \\
         & Metric & Matched acc. & Acc. & Pearson Cor. & Acc. & Acc. & Acc. & Acc. & Matthew's Cor. & F1 \\
         \midrule
         Full & \textit{Ref} & $82.4_{0.5}$ & $90.2_{0.5}$ & $88.4_{0.3}$ & $89.1_{1.0}$ & $85.2_{0.1}$ & $66.2_{3.6}$ & $92.1_{0.1}$ & $54.5_{0.4}$ & $88.1_{0.6}$ \\
         FT & Adam & $84.7_{0.4}$ & $91.0_{0.1}$ & $89.0_{0.1}$ & $91.5_{0.1}$ & $85.5_{1.7}$ & $67.6_{1.5}$ & $92.7_{0.1}$ & $58.1_{0.4}$  & $88.5_{0.2}$ \\
         & SAM & $85.3_{0.2}$ & $91.1_{0.1}$ & $89.4_{0.2}$ & $91.3_{0.6}$ & $87.0_{0.7}$ & $67.0_{0.8}$ & $93.1_{0.6}$ & $59.5_{0.4}$ & $89.2_{0.1}$ \\
         \midrule
         \textbf{IMP} & \textit{Ref} & $82.6_{0.2}$ & $90.0_{0.2}$ & $88.2_{0.2}$ & $88.9_{0.4}$ & $84.9_{0.4}$ & $66.0_{2.4}$ & $91.9_{0.5}$ & $53.8_{0.9}$ & $87.7_{0.5}$ \\
         & Adam & $82.6_{0.3}$ & $85.0_{0.2}$ & $88.3_{0.2}$ & $88.6_{ 0.1}$ & $84.2_{0.1}$ & $63.7_{0.9}$ & $91.7_{0.4}$ & $56.4_{1.4}$ & $86.9_{0.3}$ \\
         & SAM & $\mathbf{83.5_{0.1}}$ & $84.9_{0.1}$ & $\mathbf{88.9_{ 0.2}}$ & $\mathbf{89.6_{0.1}}$ & $\mathbf{85.8_{1.0}}$ & $\mathbf{71.8_{1.7}}$ & $\mathbf{93.2_{0.4}}$ & $56.1_{0.9}$ & $\mathbf{87.8_{0.2}}$ \\
         \cmidrule(lr){2-11}
         $\pm 10\%$ & Adam & $82.1_{0.3}$ & $87.2_{0.3}$ & $88.3_{0.2}$ & $88.0_{0.2}$ & $83.7_{0.4}$ & $64.2_{0.2}$ & $91.5_{0.5}$ & $55.0_{0.4}$ & $86.8_{0.4}$ \\
         & SAM & $\mathbf{83.1_{0.1}}$ & $87.4_{0.4}$ & $\mathbf{88.8_{ 0.2}}$ & $\mathbf{89.1_{0.2}}$ & $\mathbf{85.5_{0.5}}$ & $\mathbf{68.9_{0.4}}$ & $\mathbf{92.9_{0.3}}$ & $56.1_{0.2}$ & $\mathbf{88.3_{0.5}}$ \\
         \midrule
         \textbf{Std} & \textit{Ref} & $82.1$ & $90.0$ & $88.5$ & $89.9$ & $85.8$ & $63.0$ & $90.0$ & $52.0$ & $87.1$ \\
         & Adam & $82.7$ & $88.1$ & $89.2$ & $89.5$ & $85.5$ & $65.3$ & $91.1$ & $54.2$ & $86.3_{0.2}$ \\
         & SAM & $\mathbf{83.3}$ & $\mathbf{89.6}$ & $\mathbf{89.6}$ & $\mathbf{90.0}$ & $85.3$ & $\mathbf{68.2}$ & $\mathbf{92.1}$ & $\mathbf{56.8}$ & $\mathbf{87.1_{0.2}}$ \\
    \bottomrule
    \end{tabular}
    \caption{We report task metrics on the development set at \citet{chen2020BertLT}'s reference sparsities for Adam and SAM-optimized BERT-base models in their (1) Iterative Magnitude Pruning (\textbf{IMP}) and (2) \textbf{Std} pruning settings. We include \citet{chen2020BertLT}'s reference (\textit{Ref}) metrics in addition to our reported metrics (Adam, SAM). When applicable, we report mean and standard deviation calculated over 3 random seeds.}
    \label{tab:recreateBERTLT_DEV}
\end{table*}

\subsection{Additional Individual Task Plots for BERT$_{base}$ IMP}
\begin{figure*}[h]
    \centering
    \begin{subfigure}{.33\textwidth}
      \centering
      \includegraphics[width=\textwidth]{"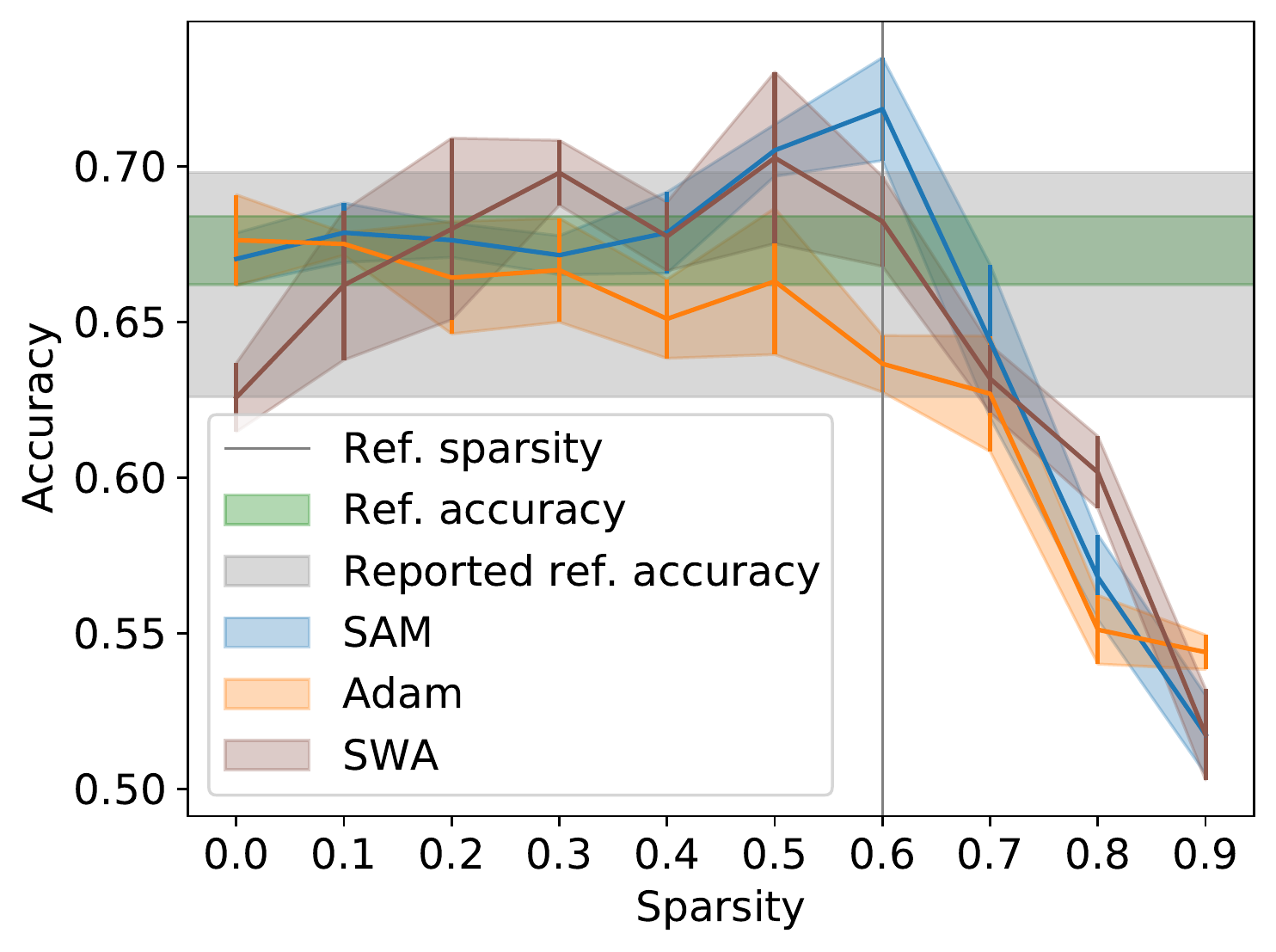"}
      \subcaption{RTE}
      \label{fig:rte_imp_swa}
    \end{subfigure}\hspace{\fill}%
    \begin{subfigure}{.33\textwidth}
      \centering
      \includegraphics[width=\textwidth]{"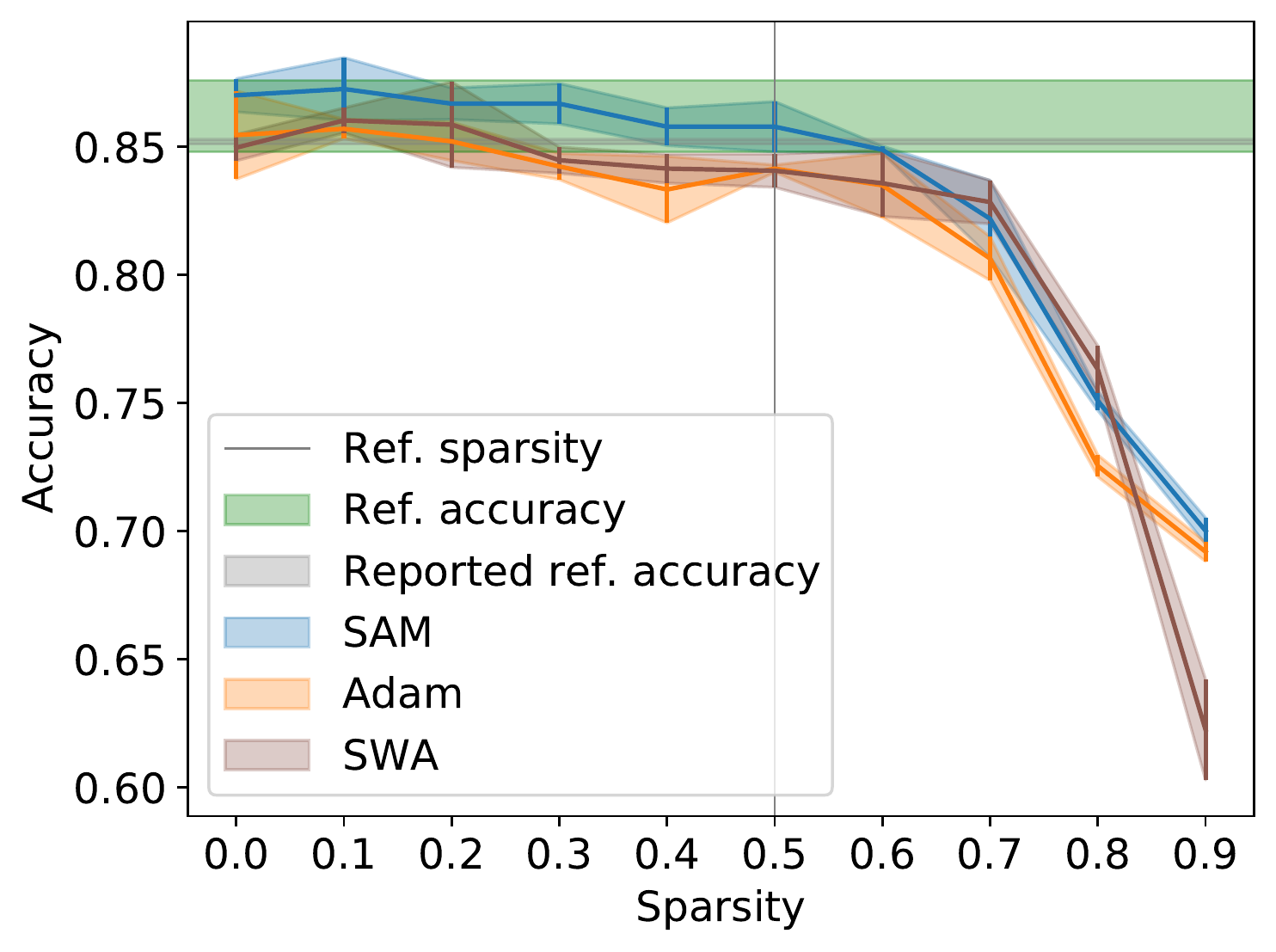"}
      \caption{MRPC}
      \label{fig:mrpc_imp_swa}
    \end{subfigure}\hspace{\fill}%
    \begin{subfigure}{.33\textwidth}
      \centering
      \includegraphics[width=\textwidth]{"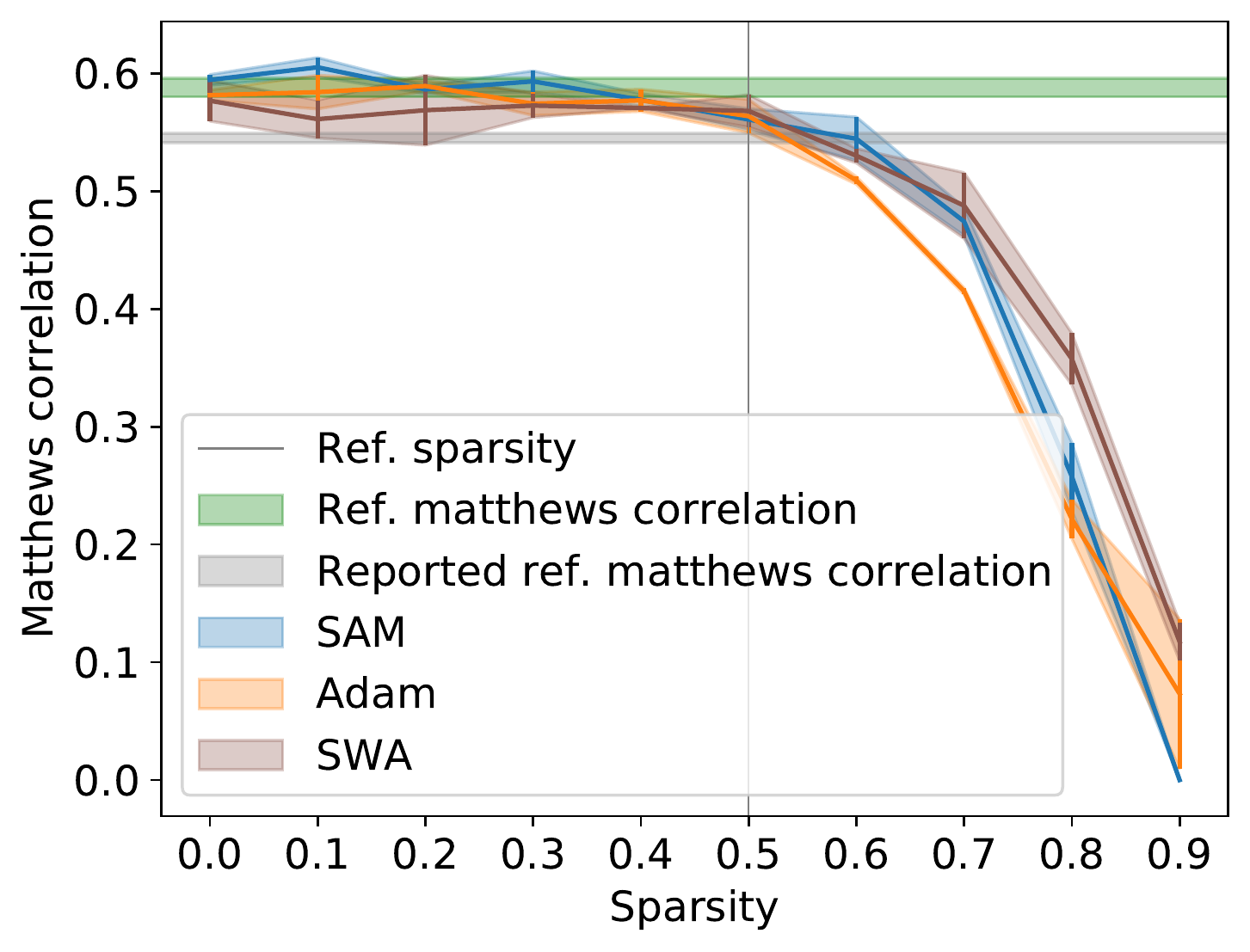"}
      \caption{CoLA}
      \label{fig:cola_imp_swa}
    \end{subfigure}\hspace{\fill}%
    \bigskip
    \begin{subfigure}{.33\textwidth}
      \centering
      \includegraphics[width=\textwidth]{"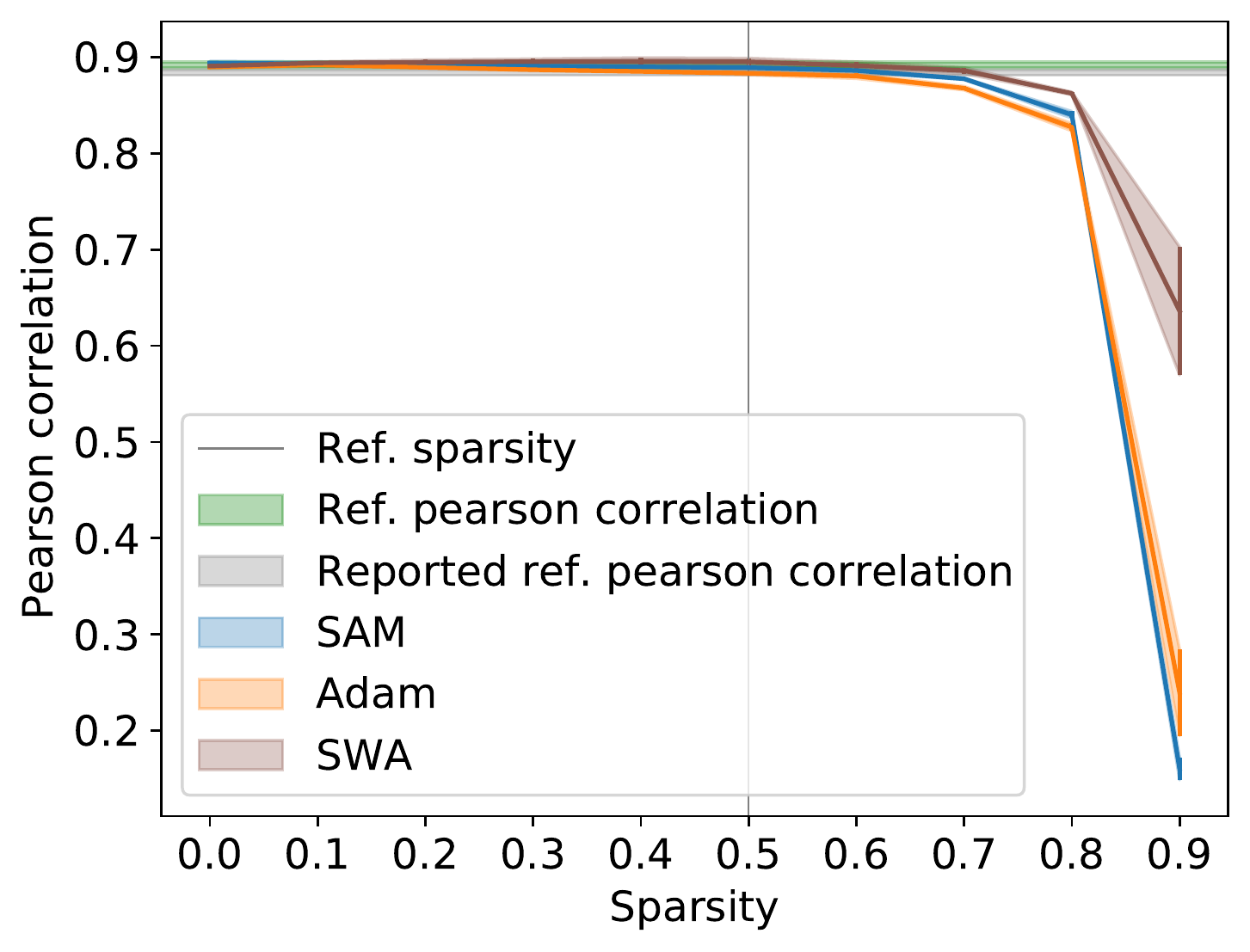"}
      \caption{STS-B}
      \label{fig:stsb_imp_swa}
    \end{subfigure}\hspace{\fill}%
    \begin{subfigure}{.33\textwidth}
      \centering
      \includegraphics[width=\textwidth]{"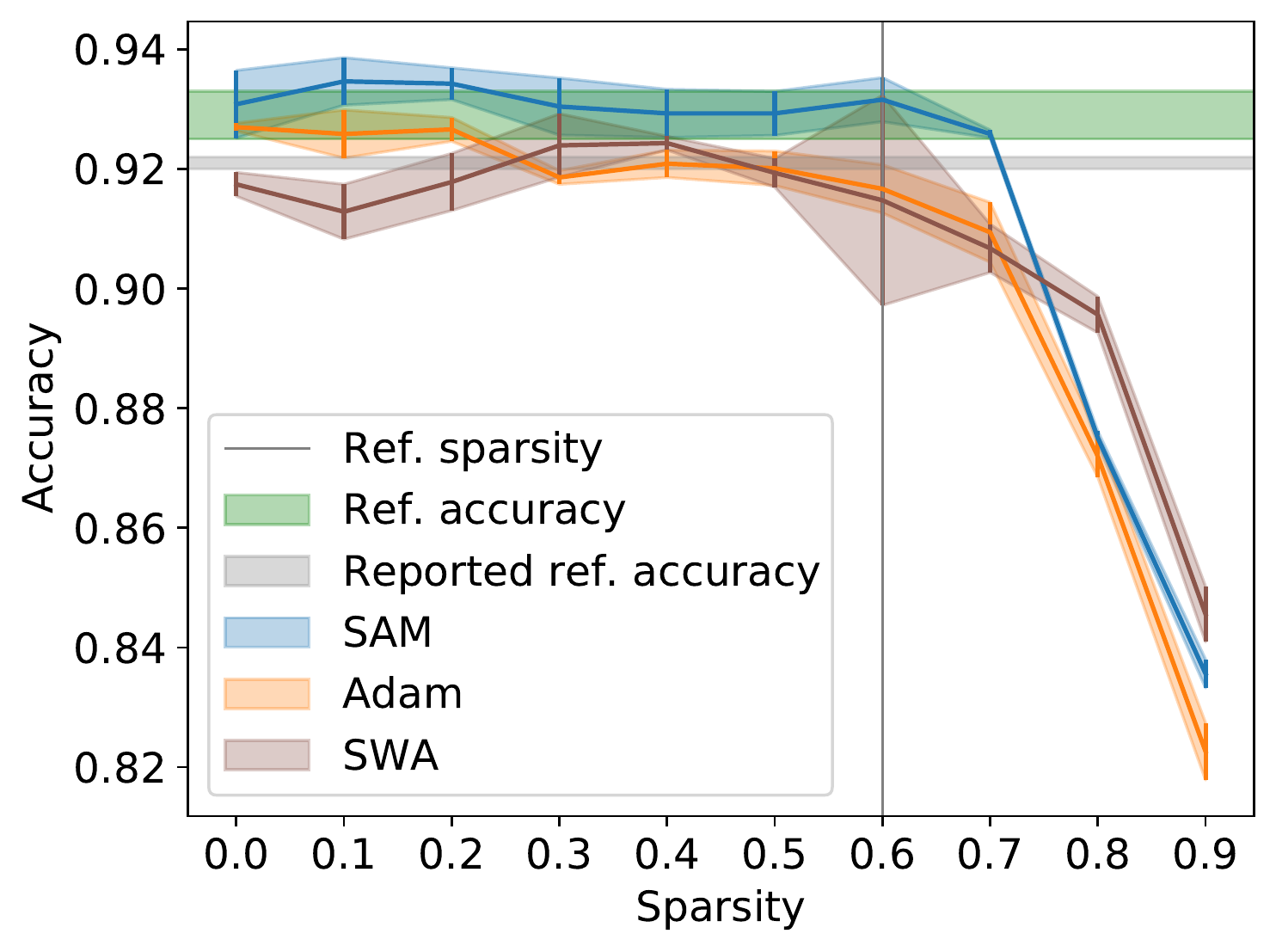"}
      \subcaption{SST-2}
      \label{fig:sst2_imp_swa}
    \end{subfigure}\hspace{\fill}%
    \begin{subfigure}{.33\textwidth}
      \centering
      \includegraphics[width=\textwidth]{"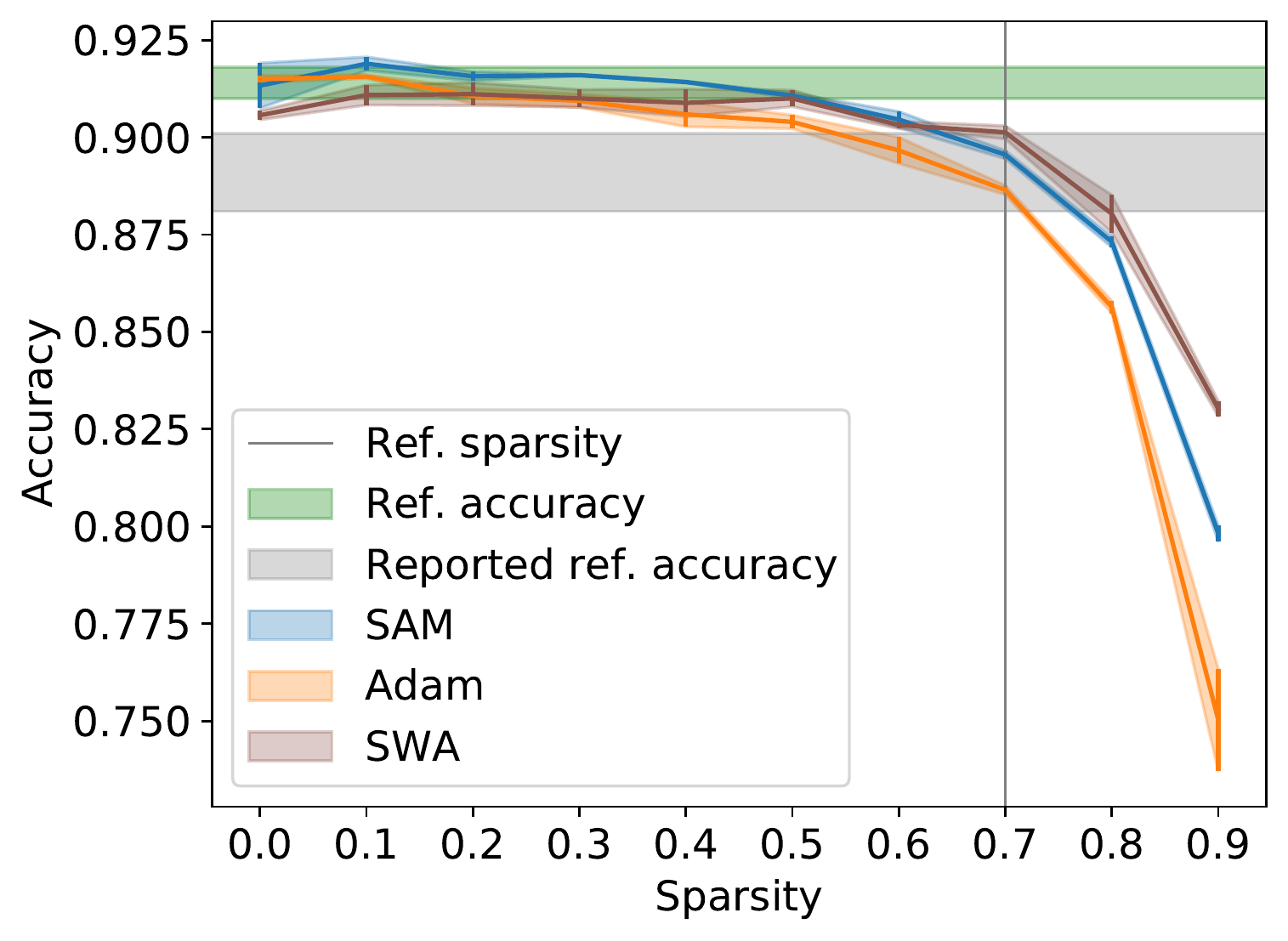"}
      \caption{QNLI}
      \label{fig:qnli_imp_swa}
    \end{subfigure}\hspace{\fill}%
    \bigskip
    \begin{subfigure}{.33\textwidth}
      \centering
      \includegraphics[width=\textwidth]{"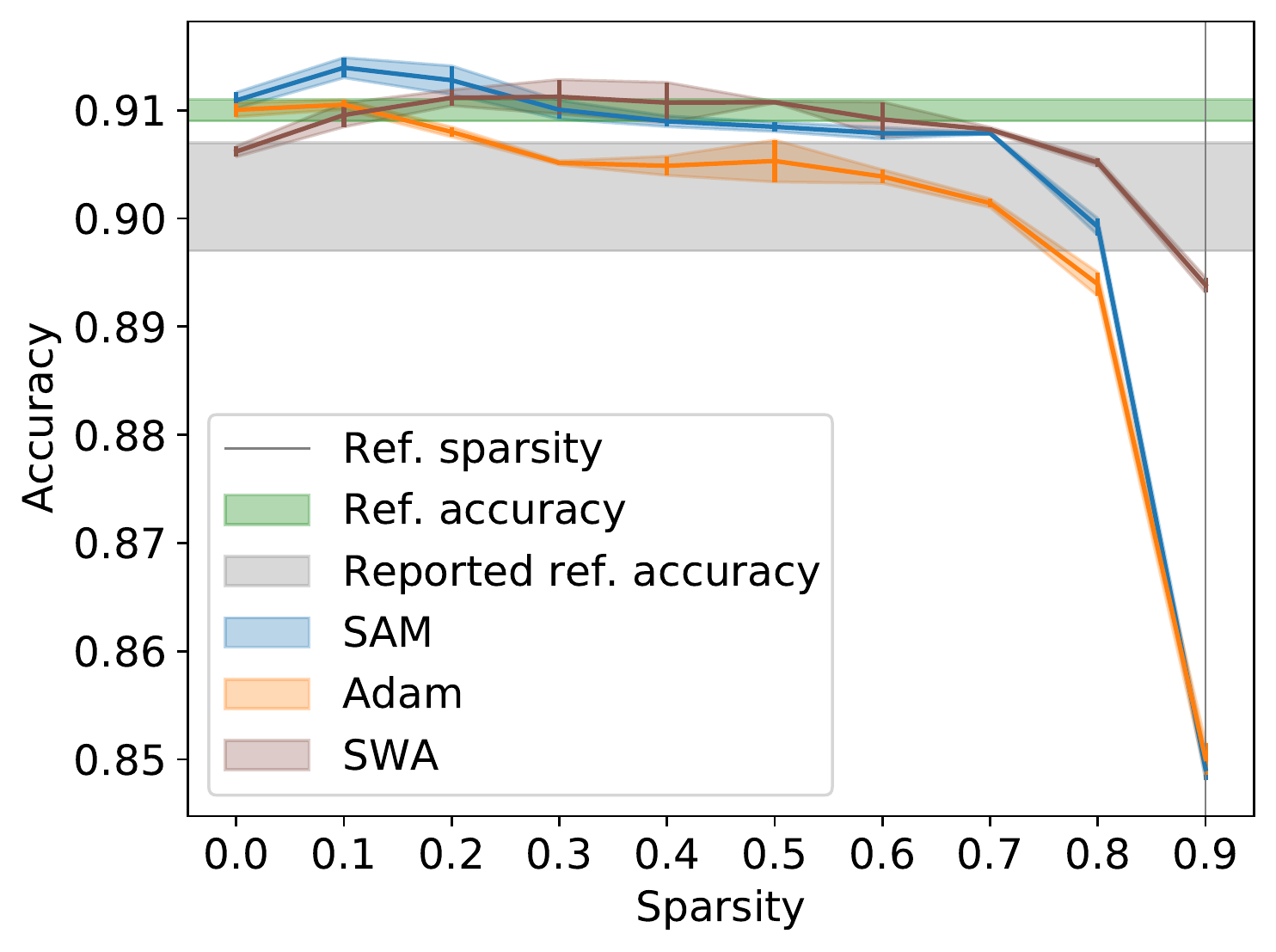"}
      \caption{QQP}
      \label{fig:qqp_imp_swa}
    \end{subfigure}\hspace{\fill}%
    \begin{subfigure}{.33\textwidth}
      \centering
      \includegraphics[width=\textwidth]{"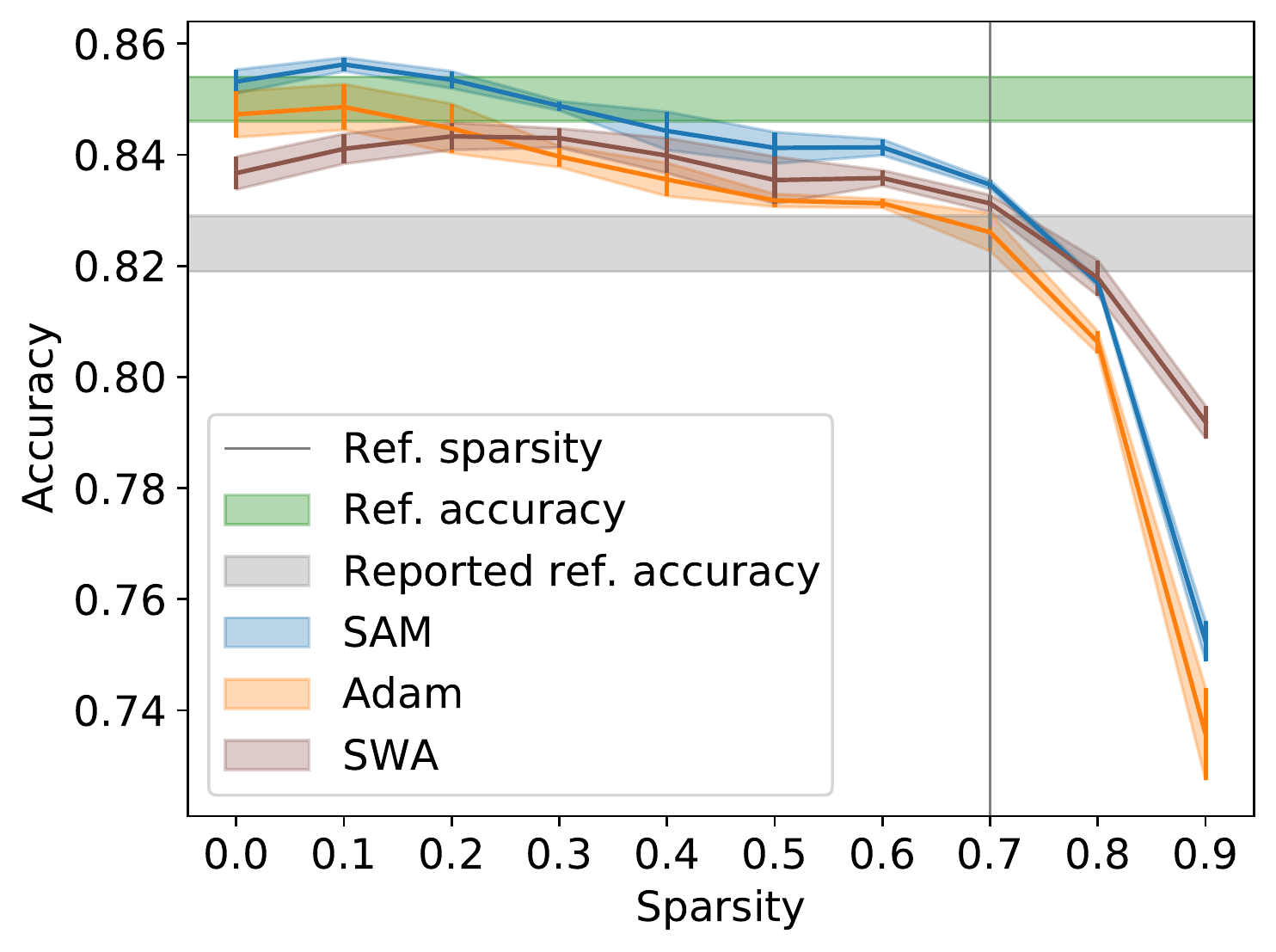"}
      \caption{MNLI}
      \label{fig:mnli_imp_swa}
    \end{subfigure}\hspace{\fill}
    \begin{subfigure}{.33\textwidth}
      \centering
      \includegraphics[width=\textwidth]{"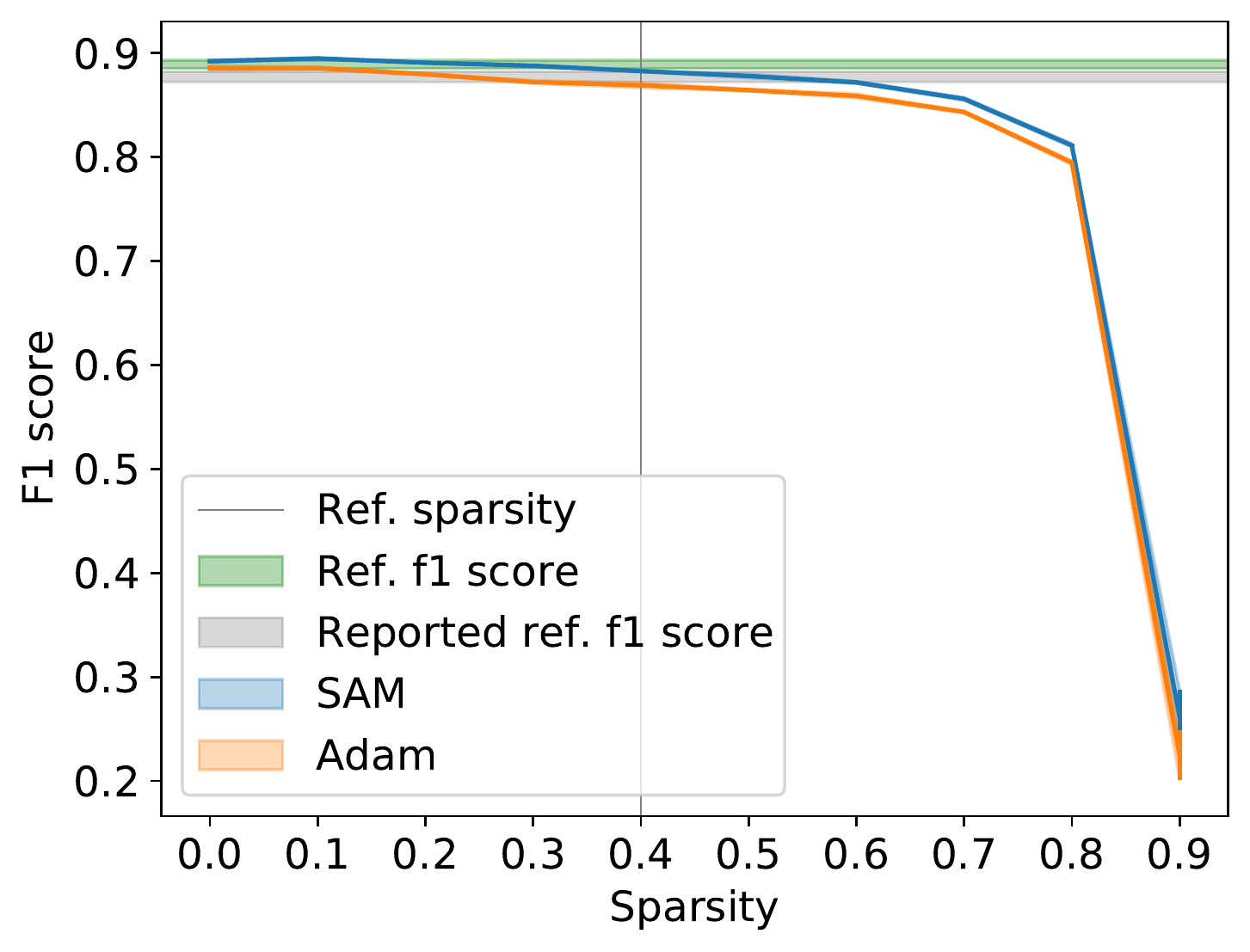"}
      \caption{SQuAD}
      \label{fig:squad_imp}
    \end{subfigure}\hspace{\fill}
    \caption{Individual plots showing sparsity vs. task metrics (validation set) for GLUE throughout IMP. The vertical lines and gray horizontal bands mark reference sparsity and "winning ticket" evaluation metric values that were obtained by \citet{chen2020BertLT}. The green horizontal bands mark the initial performance of our full fine-tuned (uncompressed) models.}
    \label{fig:task_IMP_plotsV2}
\end{figure*}

In Figure~\ref{fig:task_IMP_plotsV2}, we present the SQuAD plot comparing SAM- and Adam-optimized BERT$_{base}$ models in the unstructured IMP setting in Figure~\ref{fig:task_IMP_plots}, as well as versions of the GLUE plots in including SWA performance.

\begin{figure*}[h]
    \centering
    \begin{subfigure}{.325\textwidth}
      \centering
      \includegraphics[width=\textwidth]{"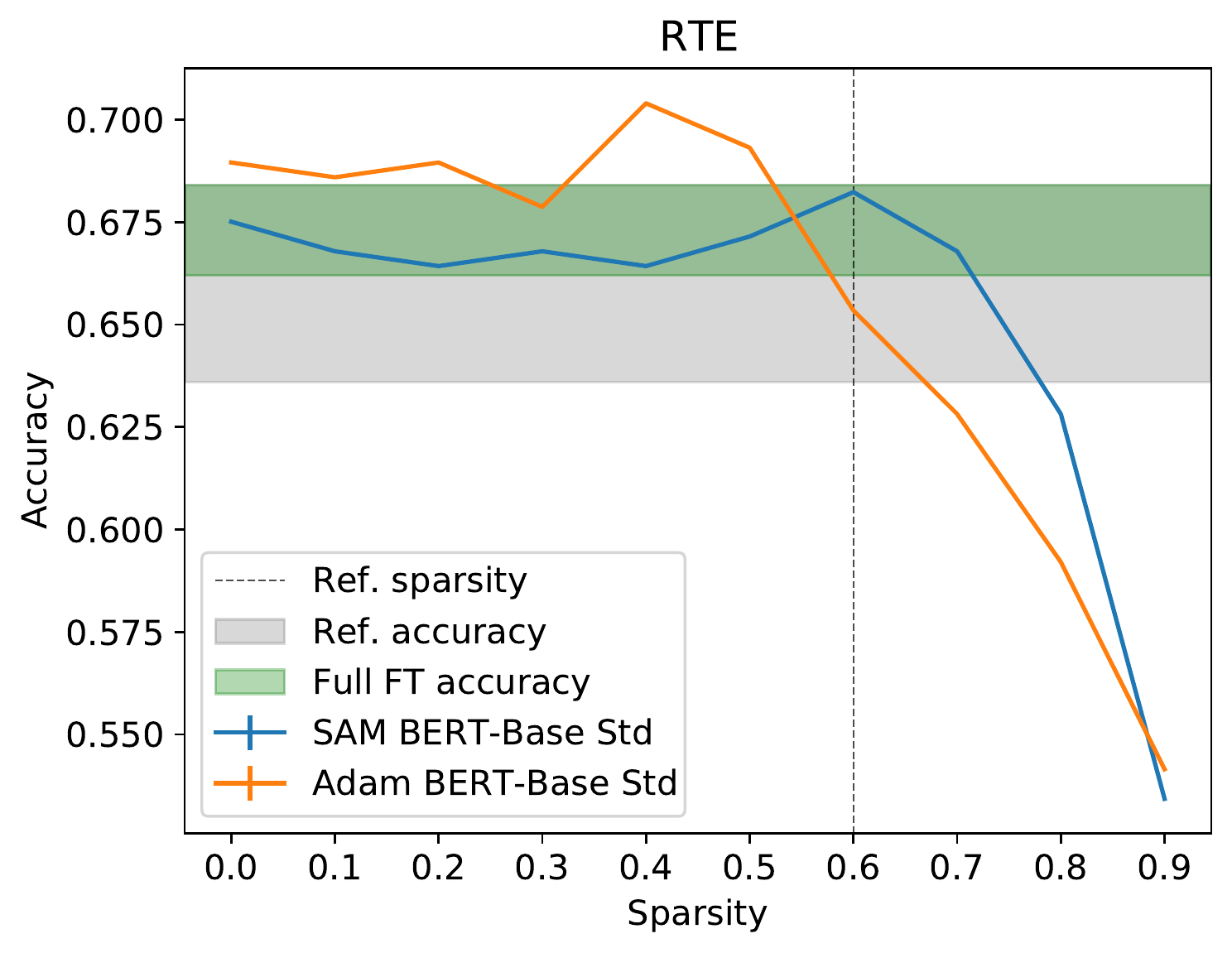"}
      \subcaption{RTE, Standard Pruning}%
      \label{fig:rte_std}
    \end{subfigure}\hspace{\fill}%
    \begin{subfigure}{.330\textwidth}
      \centering
      \includegraphics[width=\textwidth]{"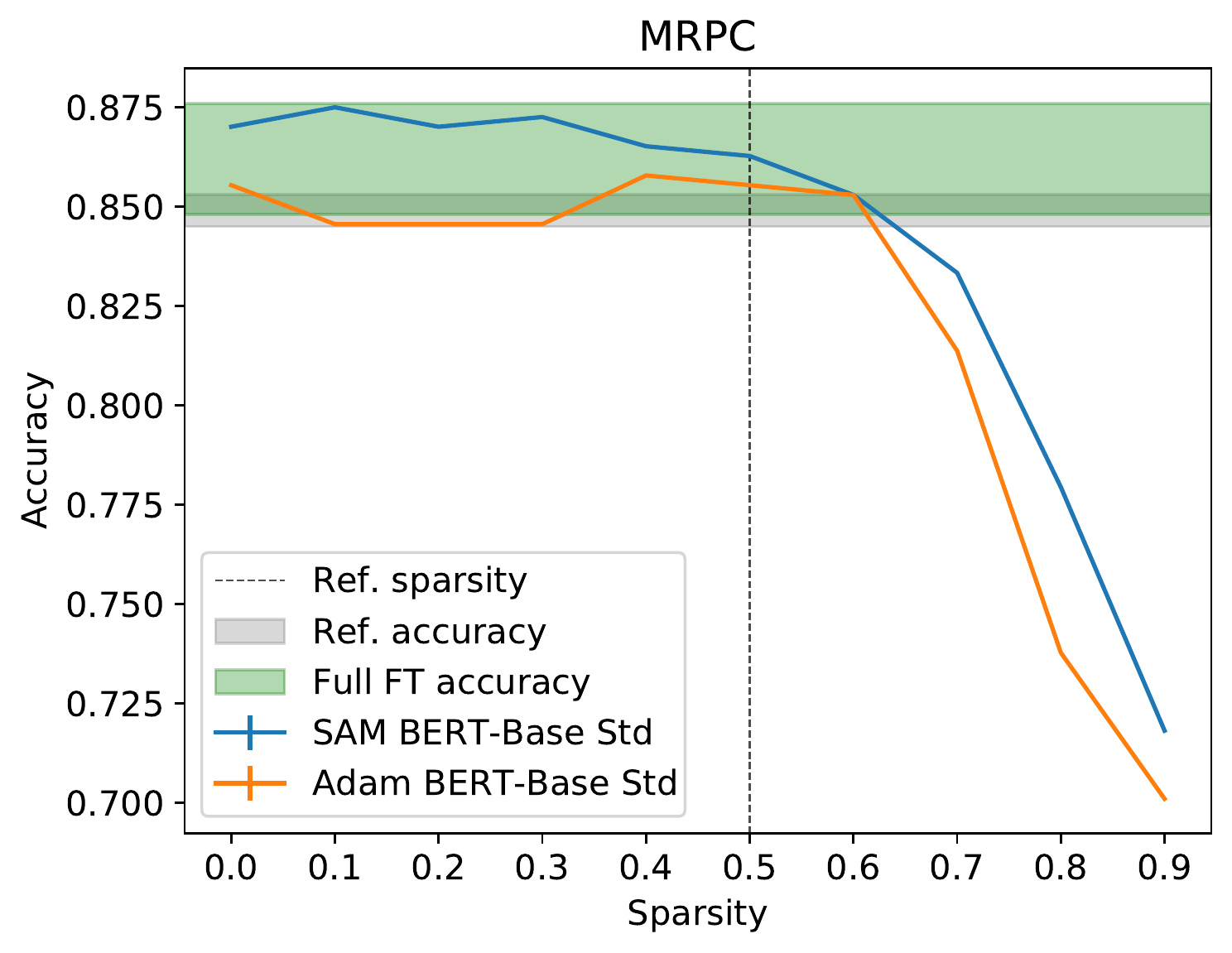"}
      \caption{MRPC, Standard Pruning}%
      \label{fig:mrpc_std}
    \end{subfigure}\hspace{\fill}%
    \begin{subfigure}{.330\textwidth}
      \centering
      \includegraphics[width=\textwidth]{"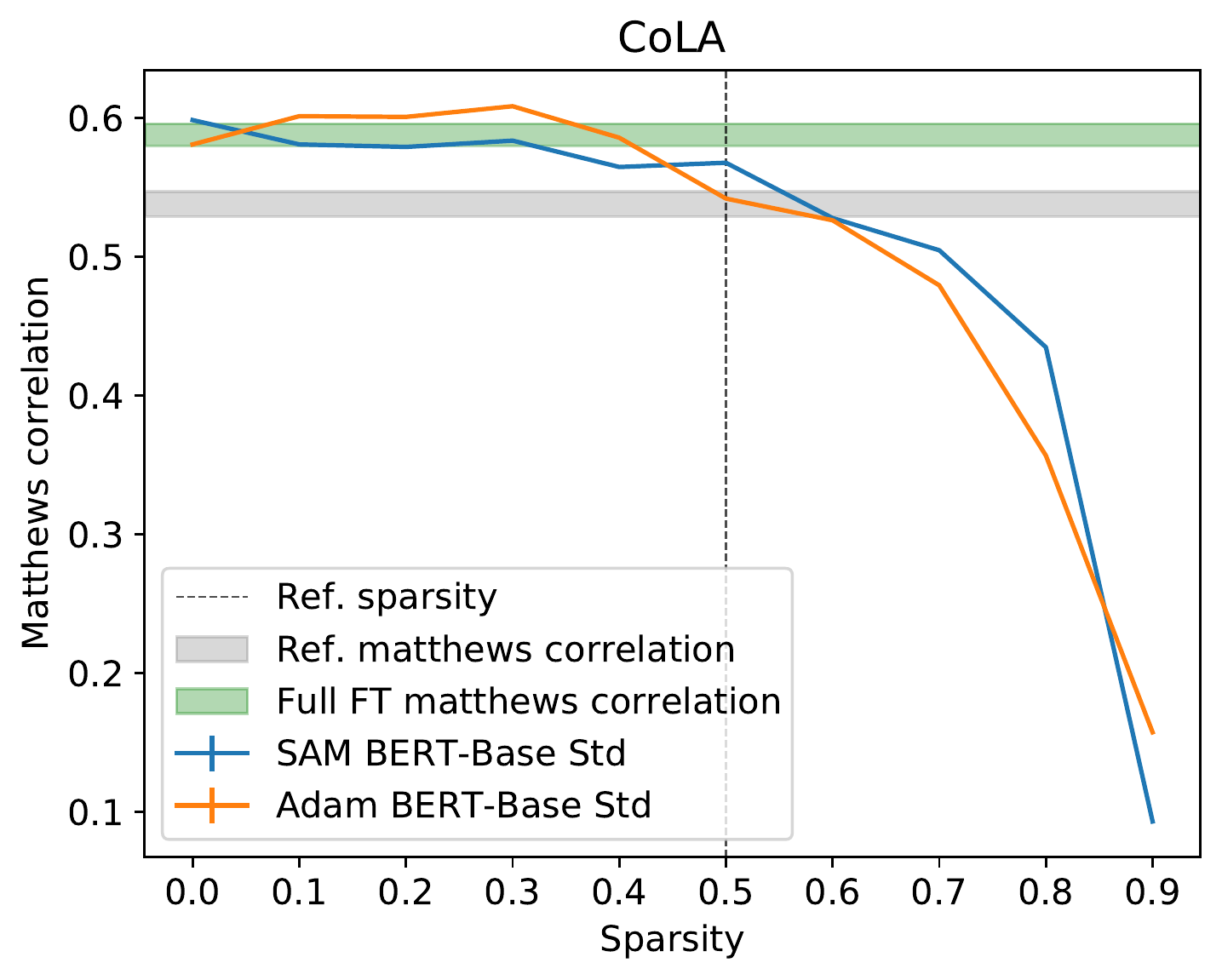"}
      \caption{CoLA, Standard Pruning}%
      \label{fig:cola_std}
    \end{subfigure}\hspace{\fill}%
    \begin{subfigure}{.330\textwidth}
      \centering
      \includegraphics[width=\textwidth]{"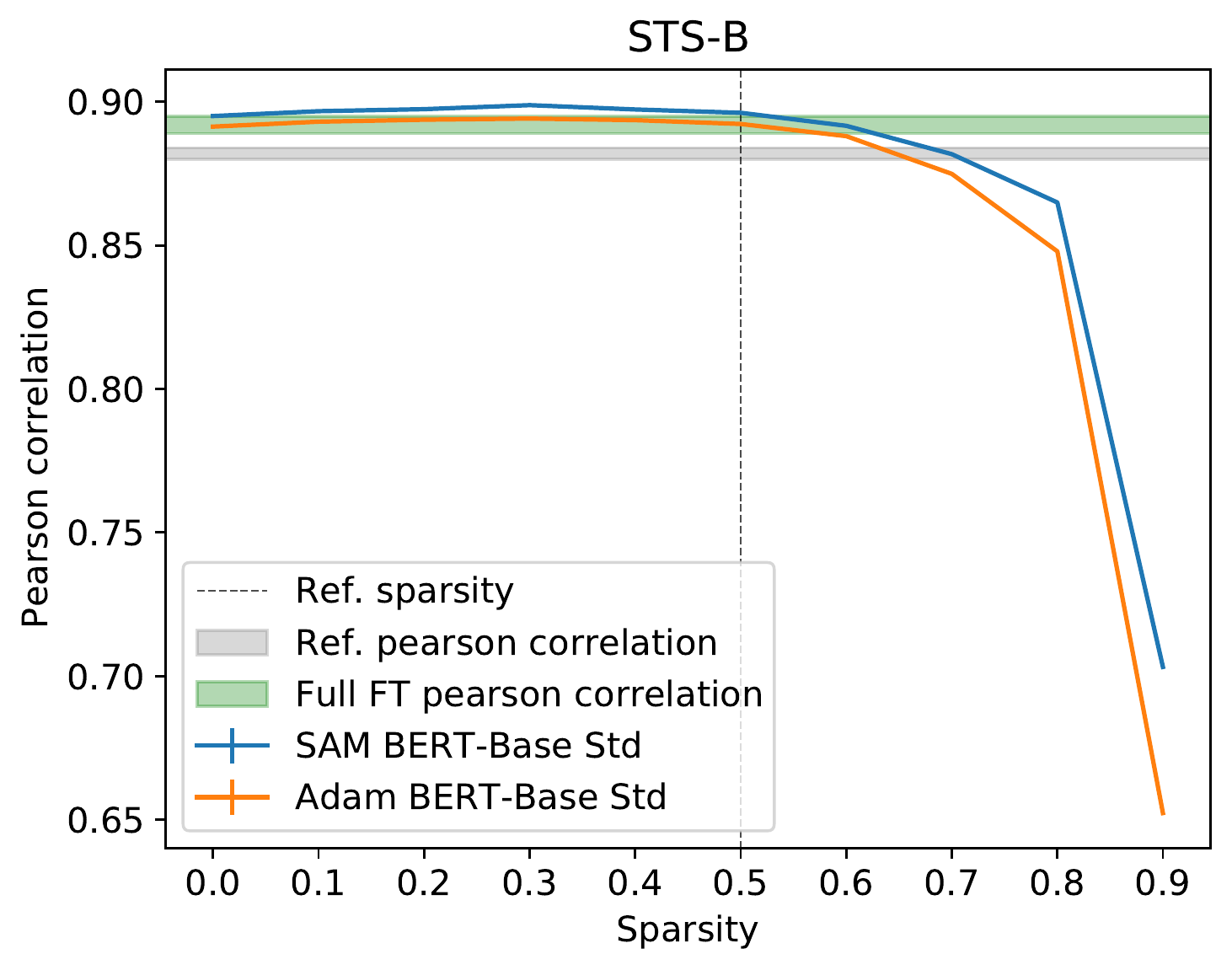"}
      \caption{STS-B, Standard Pruning}%
      \label{fig:stsb_std}
    \end{subfigure}\hspace{\fill}%
    \bigskip
    \begin{subfigure}{.325\textwidth}
      \centering
      \includegraphics[width=\textwidth]{"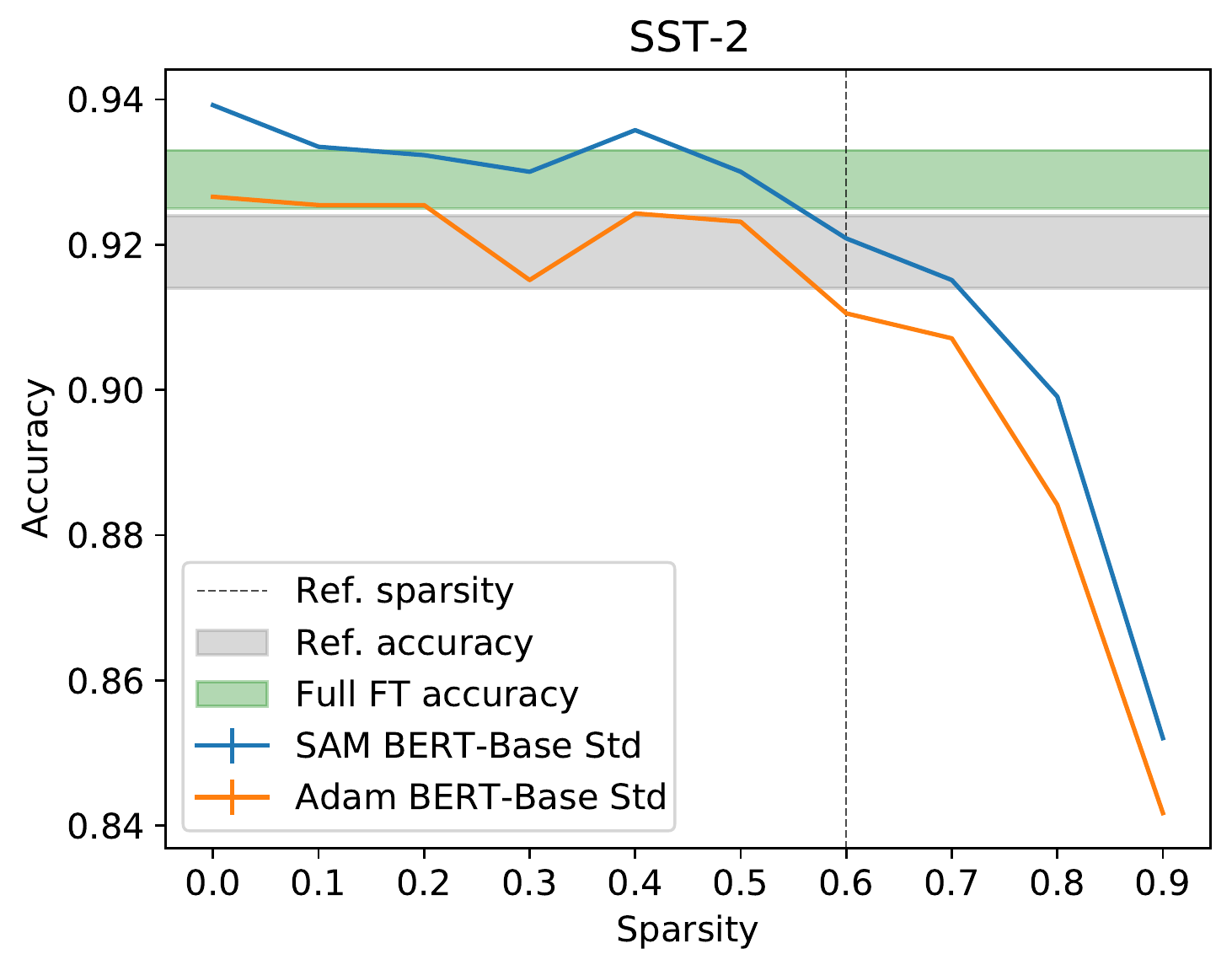"}
      \subcaption{SST-2, Standard Pruning}%
      \label{fig:sst2_std}
    \end{subfigure}\hspace{\fill}%
    \begin{subfigure}{.325\textwidth}
      \centering
      \includegraphics[width=\textwidth]{"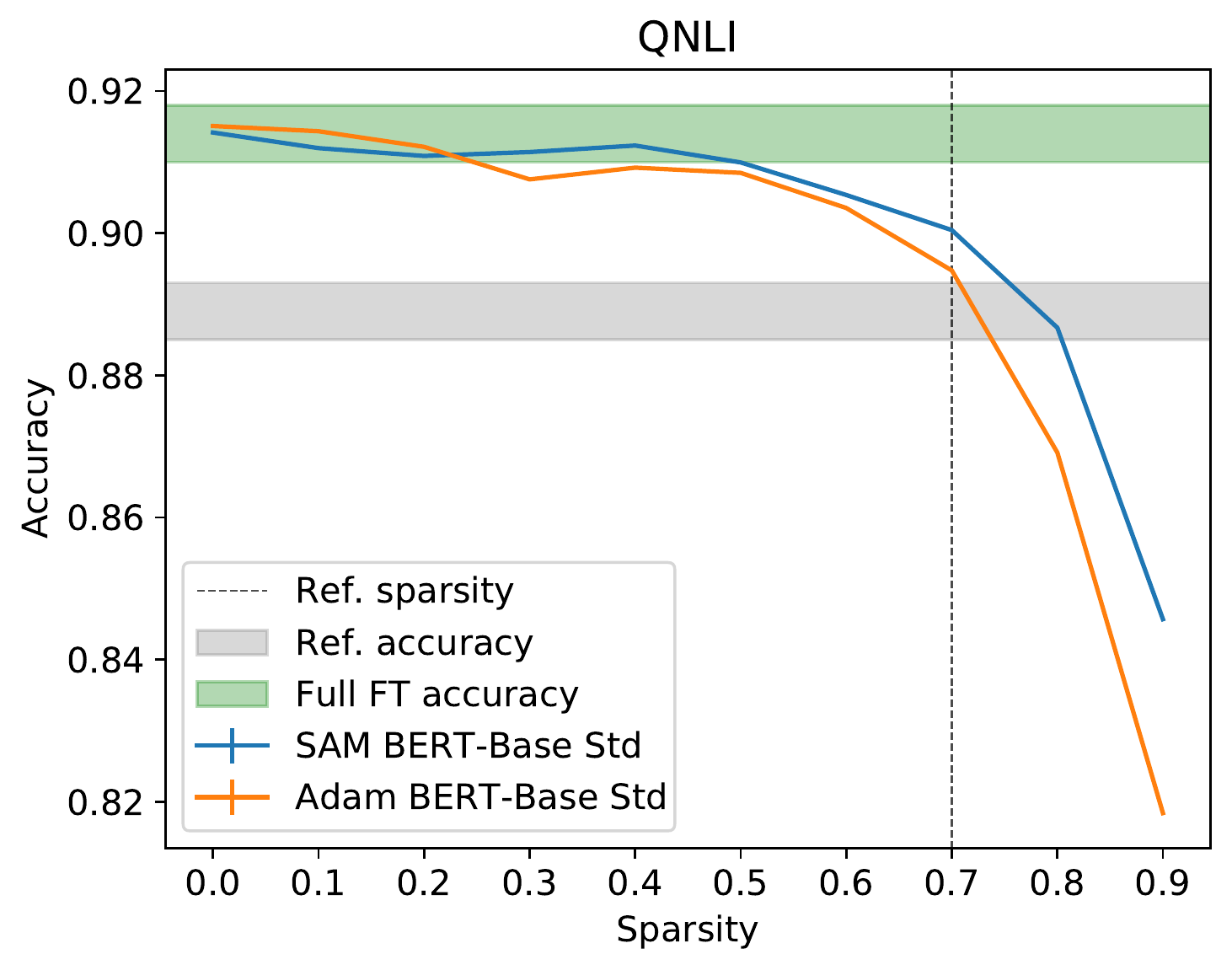"}
      \caption{QNLI, Standard Pruning}%
      \label{fig:qnli_std}
    \end{subfigure}\hspace{\fill}%
    \begin{subfigure}{.325\textwidth}
      \centering
      \includegraphics[width=\textwidth]{"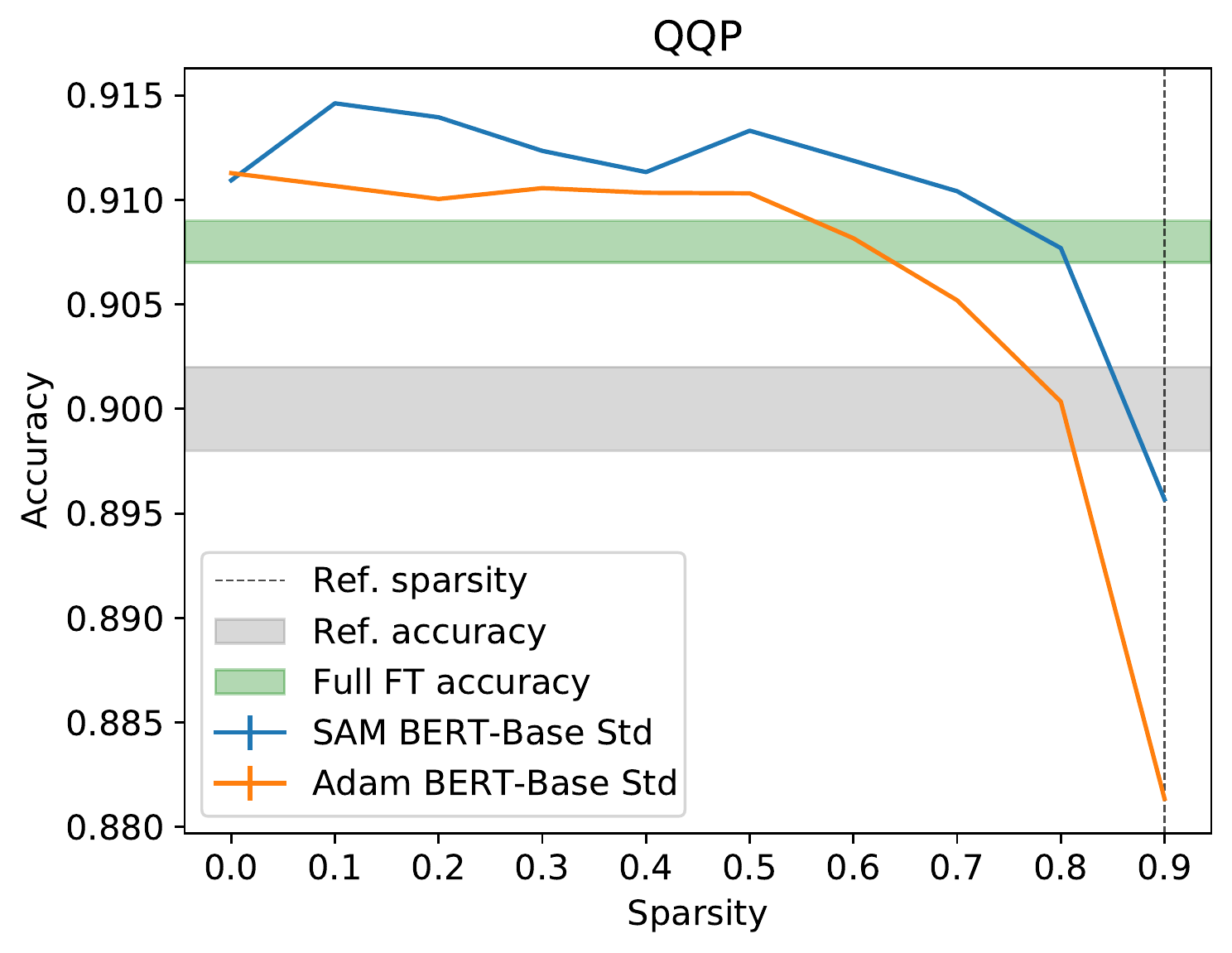"}
      \caption{QQP, Standard Pruning}%
      \label{fig:qqp_std}
    \end{subfigure}\hspace{\fill}%
    \begin{subfigure}{.325\textwidth}
      \centering
      \includegraphics[width=\textwidth]{"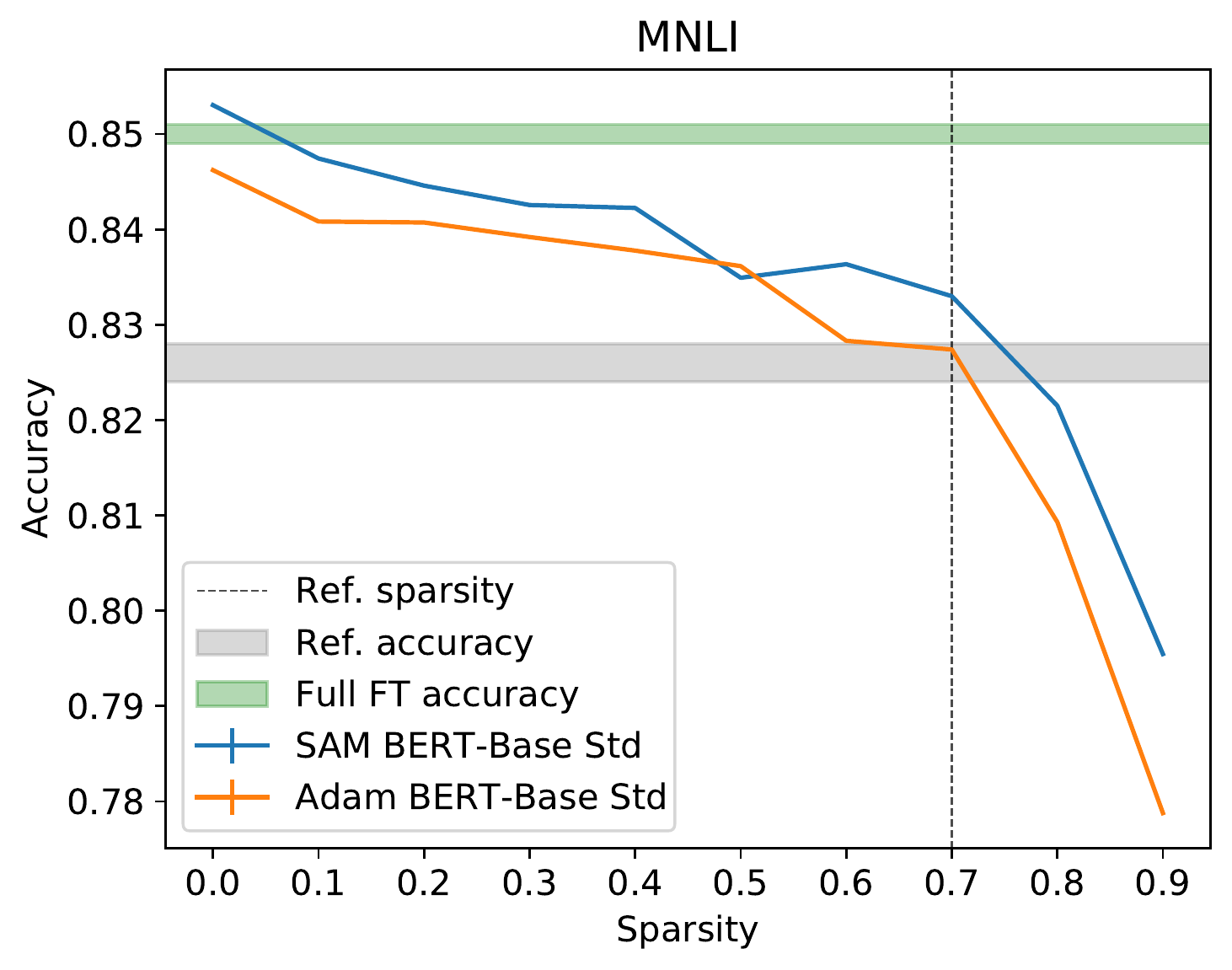"}
      \caption{MNLI, Standard Pruning}%
      \label{fig:mnli_std}
    \end{subfigure}\hspace{\fill}
    \begin{subfigure}{.325\textwidth}
      \centering
      \includegraphics[width=\textwidth]{"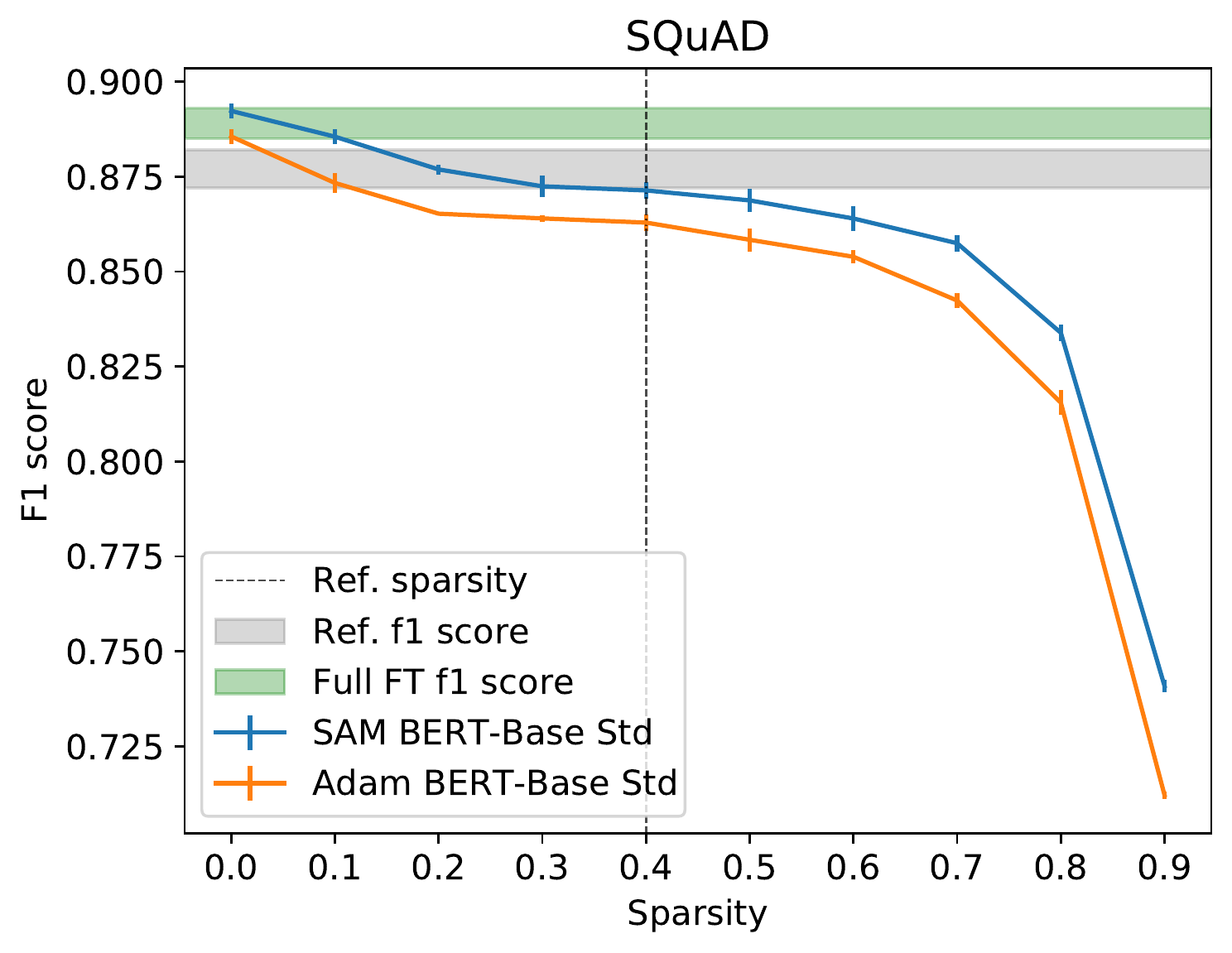"}
      \caption{SQuAD, Standard Pruning}%
      \label{fig:squad_plot}
    \end{subfigure}\hspace{\fill}
    \caption{Individual plots showing sparsity vs. accuracy for GLUE tasks and SQuAD in BERT$_{base}$ models compressed with standard pruning (IMP with no rewinding of weights). The vertical lines and gray horizontal bands mark reference sparsity and "winning ticket" evaluation metric values that were obtained by \citet{chen2020BertLT}. The green horizontal bands mark the initial performance of our full fine-tuned models.}
    \label{fig:task_std_plots}
\end{figure*}

Plots for BERT$_{base}$ trained on GLUE and SQuAD tasks with unstructured \textit{standard pruning} are shown in Figure~\ref{fig:task_std_plots}.

\subsection{Is SAM just implicitly doing $\ell_1$ regularization?}
\label{subsec:l1reg_appendix}
No. It is clear that $\ell_1$ regularization induces sparsity in a different way compared to SAM. Although in some cases $\ell_1$ regularization can help a model reach higher accuracies at certain sparsity levels, we observe that simply optimizing with Adam throughout an iterative pruning process does not allow the model to reach SAM-optimized models' compression performance. Moreover, $\ell_1$ regularization can actually hurt compression performance in some cases. 

Further investigation is needed to understand the specific mechanisms allowing SAM to induce greater compresseibility in models.

\begin{figure*}[h]
    \centering
    \begin{subfigure}{.495\textwidth}
      \centering
      \includegraphics[width=\textwidth]{"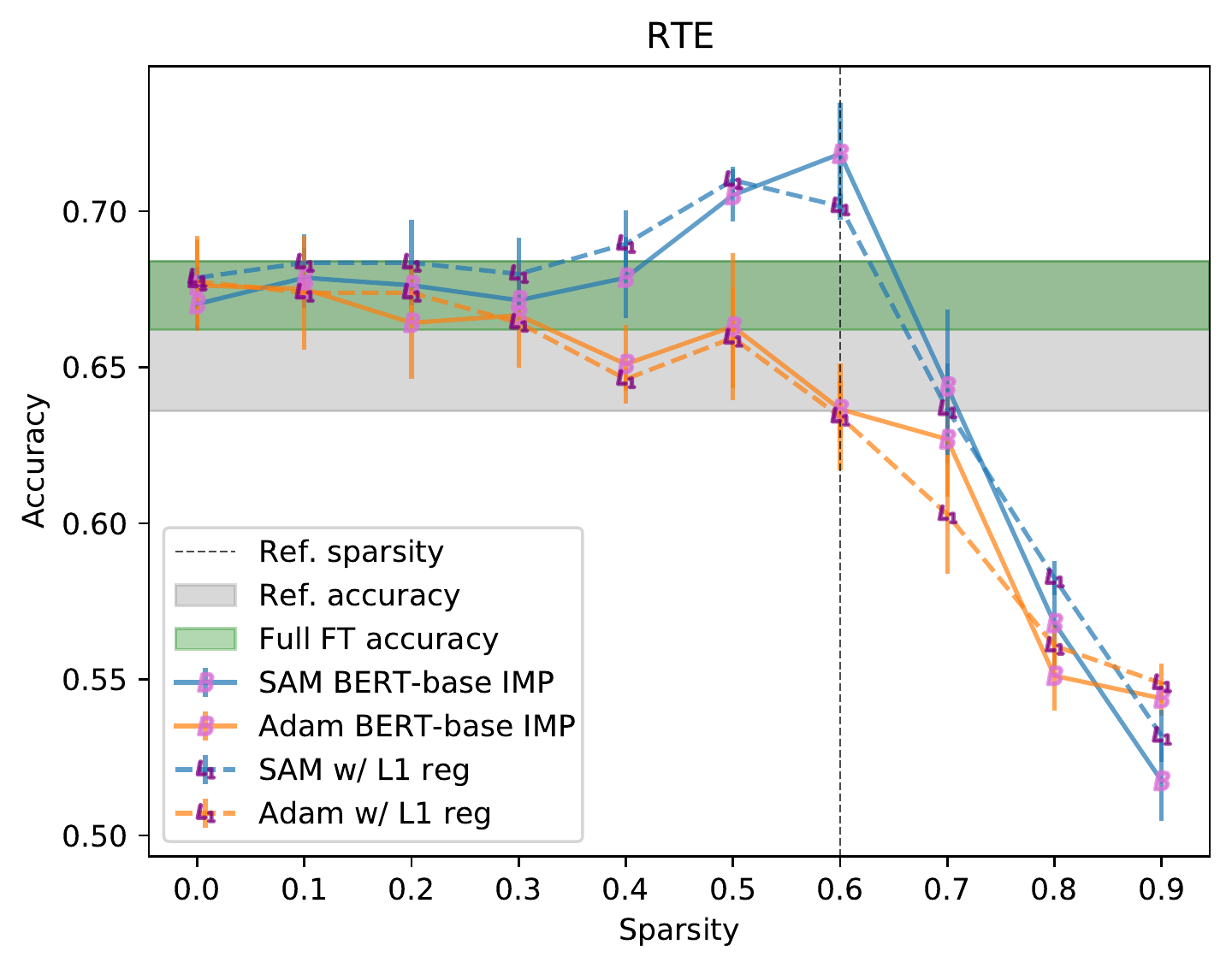"}
      \subcaption{RTE, IMP w/ $\ell_1$ Regularization}
      \label{fig:rte_l1_imp}
    \end{subfigure}\hspace{\fill}%
    \begin{subfigure}{.495\textwidth}
      \centering
      \includegraphics[width=\textwidth]{"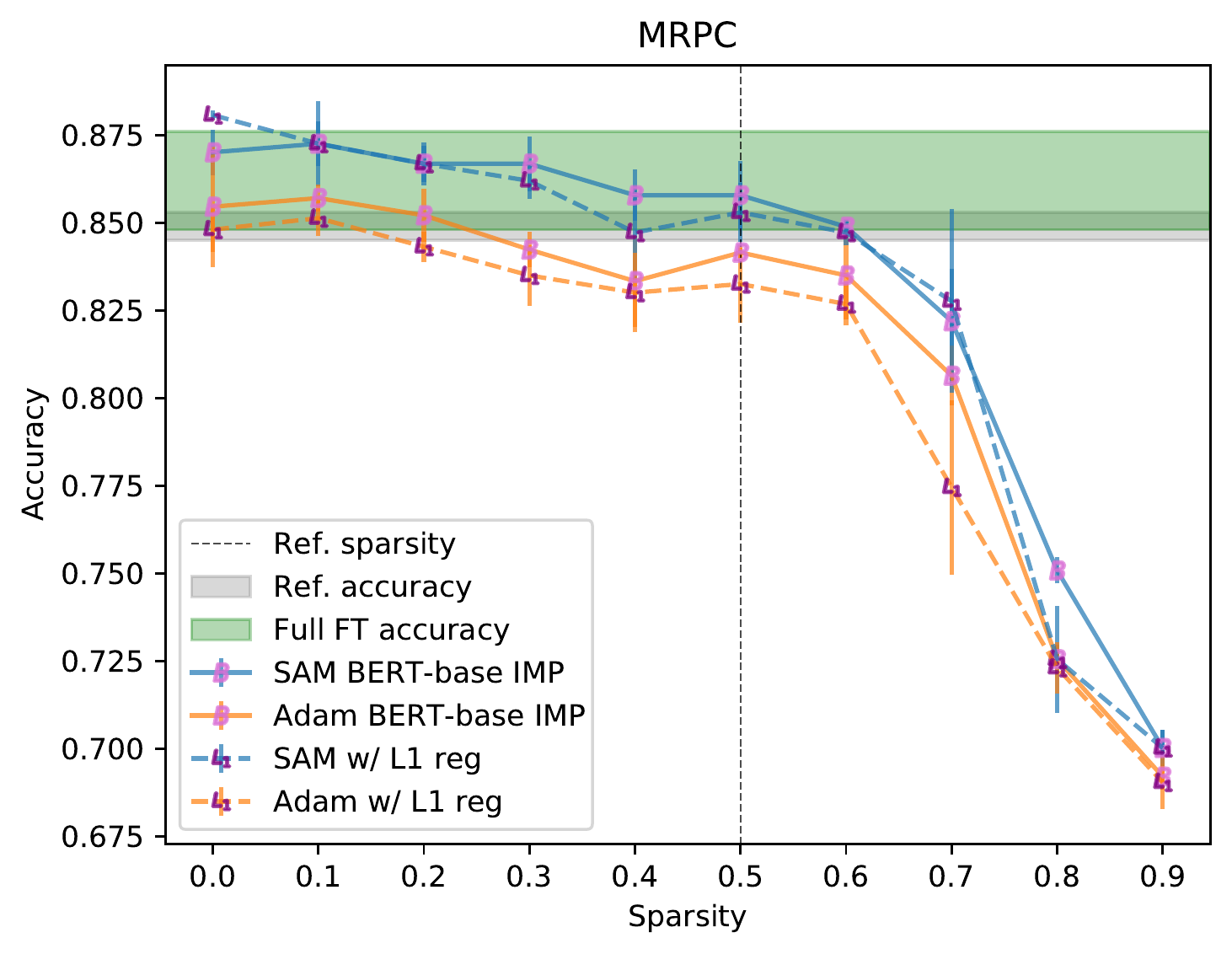"}
      \caption{MRPC, IMP w/ $\ell_1$ Regularization}
      \label{fig:mrpc_l1_imp}
    \end{subfigure}\hspace{\fill}%
    \begin{subfigure}{.495\textwidth}
      \centering
      \includegraphics[width=\textwidth]{"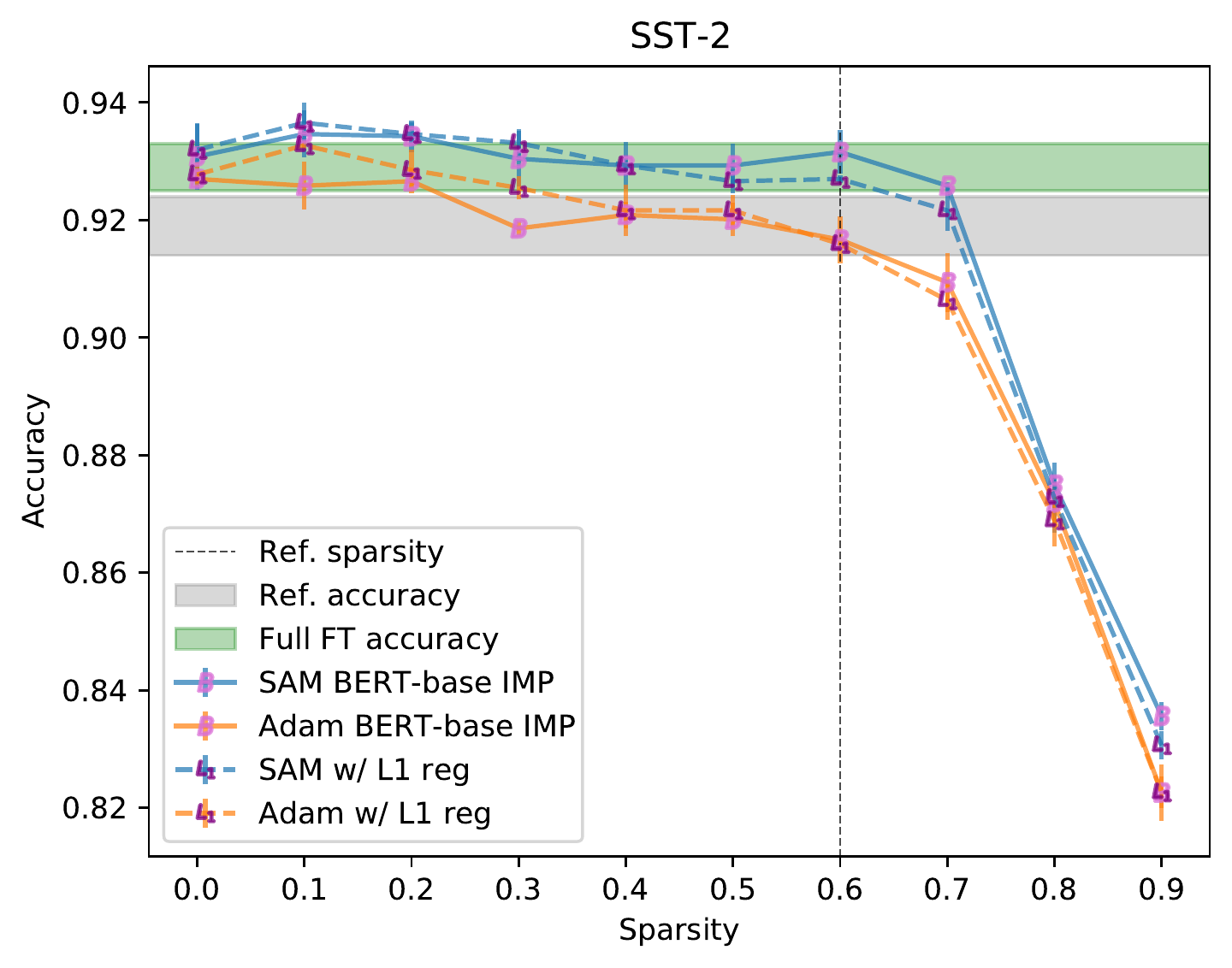"}
      \subcaption{SST-2, IMP w/ $\ell_1$ Regularization}
      \label{fig:sst2_l1_imp}
    \end{subfigure}\hspace{\fill}%
    \begin{subfigure}{.495\textwidth}
      \centering
      \includegraphics[width=\textwidth]{"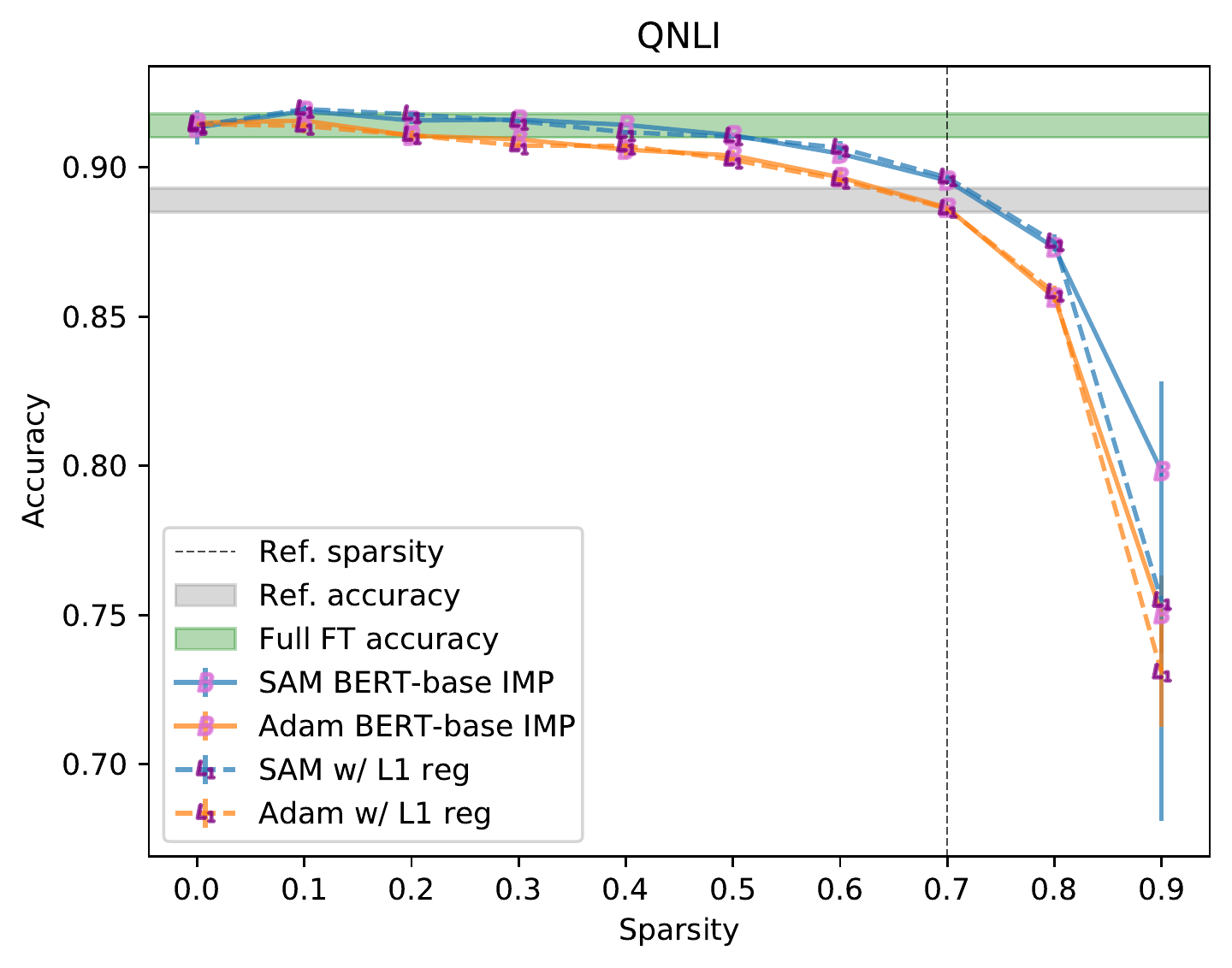"}
      \caption{QNLI, IMP w/ $\ell_1$ Regularization}
      \label{fig:qnli_l1_imp}
    \end{subfigure}\hspace{\fill}%

    \caption{Individual plots showing sparsity vs. accuracy for GLUE tasks in $\ell_1$-regularized BERT$_{base}$ models compressed with iterative magnitude pruning (IMP), with regular BERT$_{base}$ models for comparison. The vertical lines and gray horizontal bands mark reference sparsity and "winning ticket" evaluation metric values that were obtained by \citet{chen2020BertLT}. The green horizontal bands mark the initial performance of our full fine-tuned models.}
    \label{fig:task_l1_imp_plots}
\end{figure*}

\begin{figure*}[h]
    \centering
    \begin{subfigure}{.495\textwidth}
      \centering
      \includegraphics[width=\textwidth]{"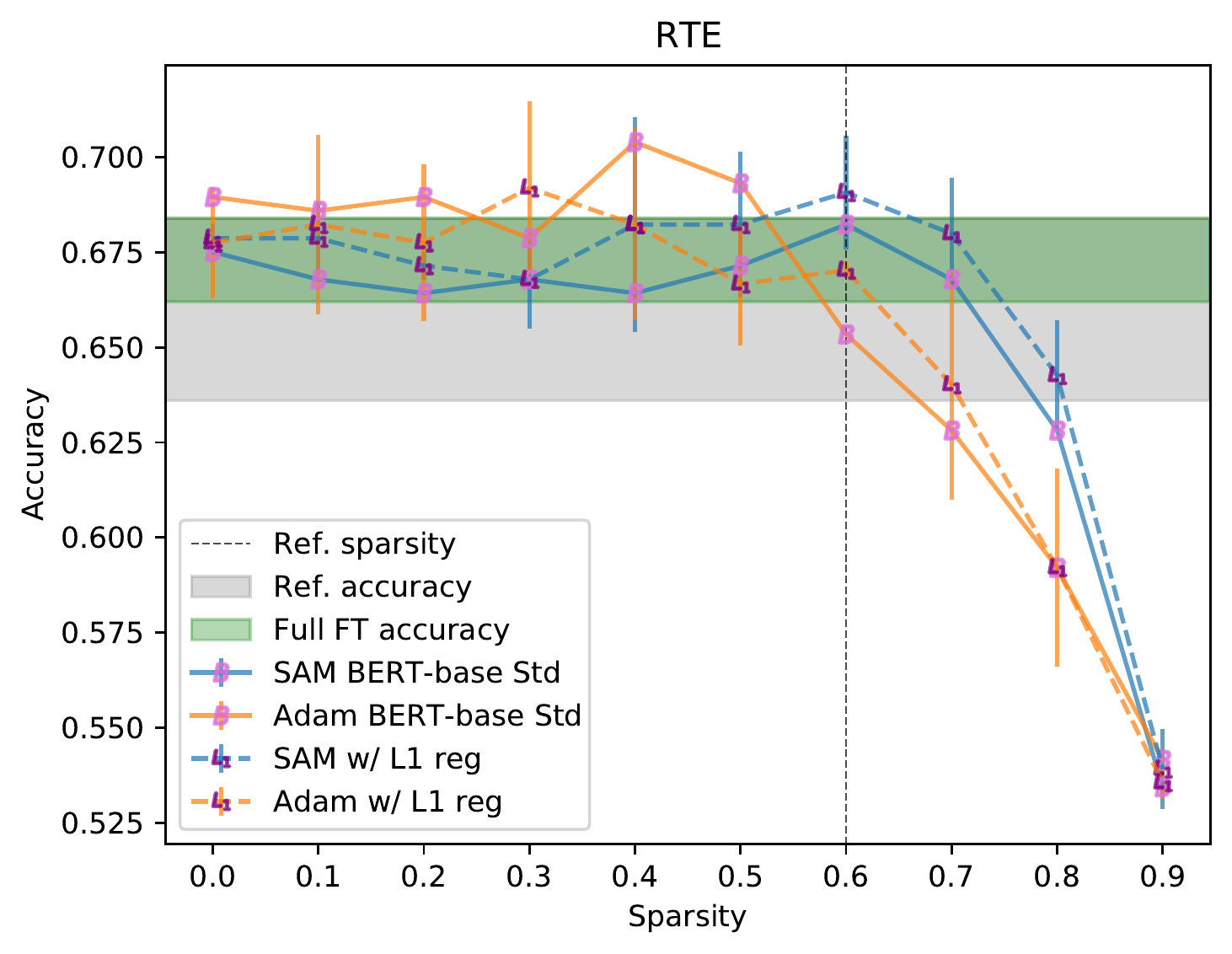"}
      \subcaption{RTE, Standard Pruning w/ $\ell_1$ Regularization}
      \label{fig:rte_l1_std}
    \end{subfigure}\hspace{\fill}%
    \begin{subfigure}{.495\textwidth}
      \centering
      \includegraphics[width=\textwidth]{"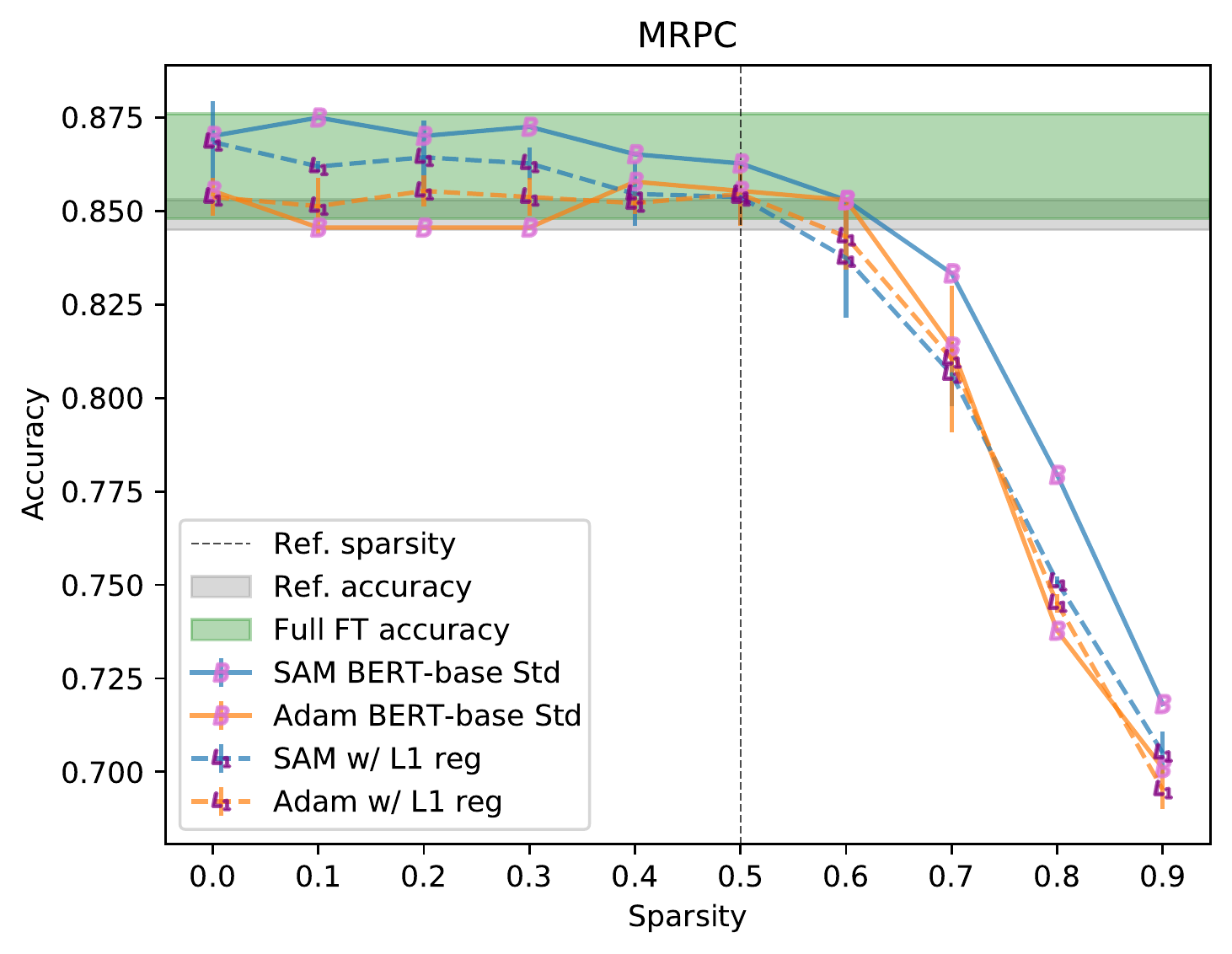"}
      \caption{MRPC, Standard Pruning w/ $\ell_1$ Regularization}
      \label{fig:mrpc_l1_std}
    \end{subfigure}\hspace{\fill}%
    \begin{subfigure}{.495\textwidth}
      \centering
      \includegraphics[width=\textwidth]{"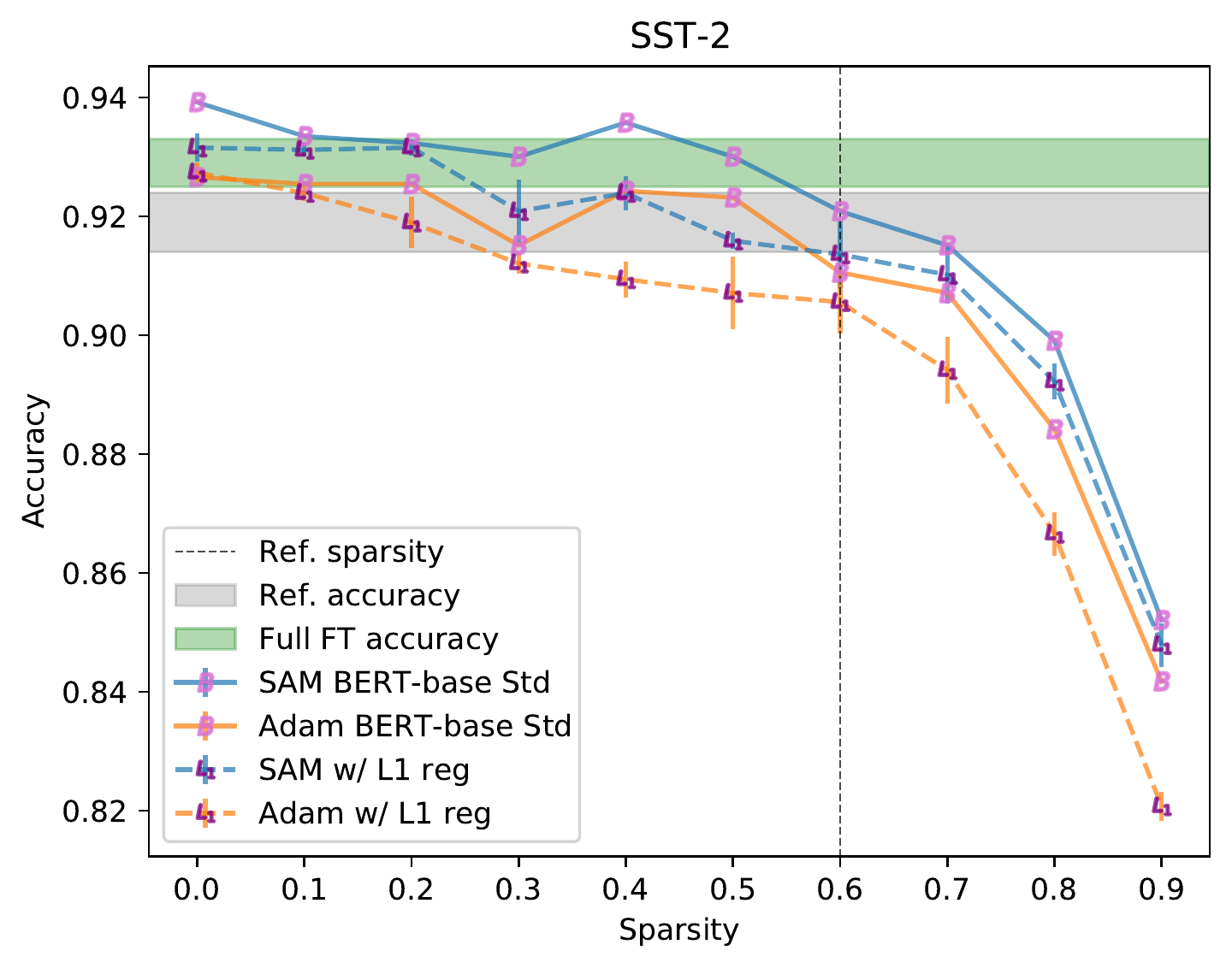"}
      \subcaption{SST-2, Standard Pruning w/ $\ell_1$ Regularization}
      \label{fig:sst2_l1_std}
    \end{subfigure}\hspace{\fill}%
    \begin{subfigure}{.495\textwidth}
      \centering
      \includegraphics[width=\textwidth]{"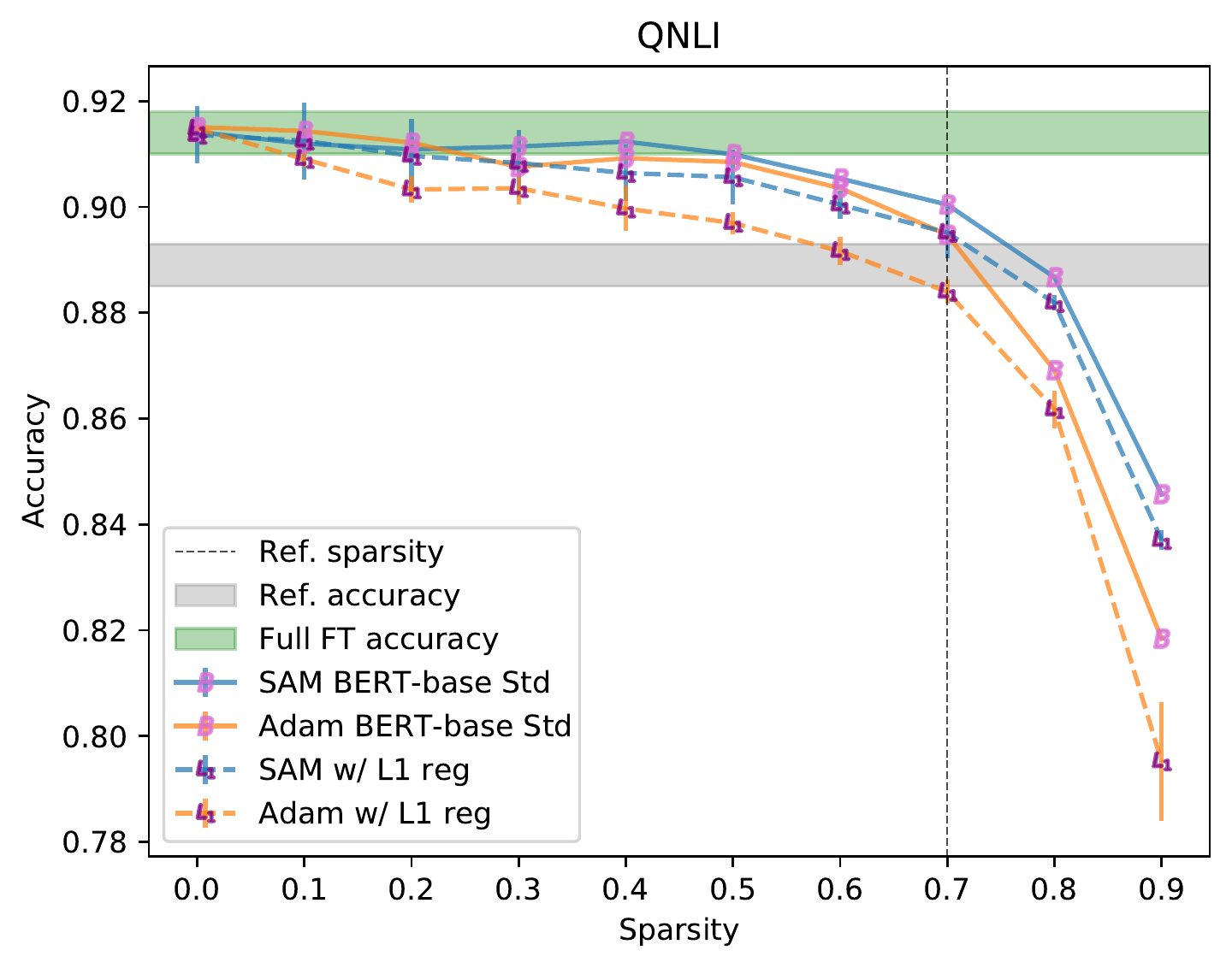"}
      \caption{QNLI, Standard Pruning w/ $\ell_1$ Regularization}
      \label{fig:qnli_l1_std}
    \end{subfigure}\hspace{\fill}%

    \caption{Individual plots showing sparsity vs. accuracy for GLUE tasks and SQuAD in BERT$_{base}$ models trained with $\ell_1$ regularization during iterative compression with standard pruning, with BERT$_{base}$ models trained without regularization during standard pruning for comparison. The vertical lines and gray horizontal bands mark reference sparsity and "winning ticket" evaluation metric values that were obtained by \citet{chen2020BertLT}. The green horizontal bands mark the initial performance of our full fine-tuned models.}
    \label{fig:task_l1_stdprune_plots}
\end{figure*}

\subsection{Detailed Structure vs Optimization Results}
Table \ref{tab:OptimizerSwapAnalysis} contains numbers presented in Figure \ref{fig:OptimizerSwap} from \S\ref{sec:analysis}.

\begin{figure*}[h]
    \centering
    \includegraphics[width=\textwidth]{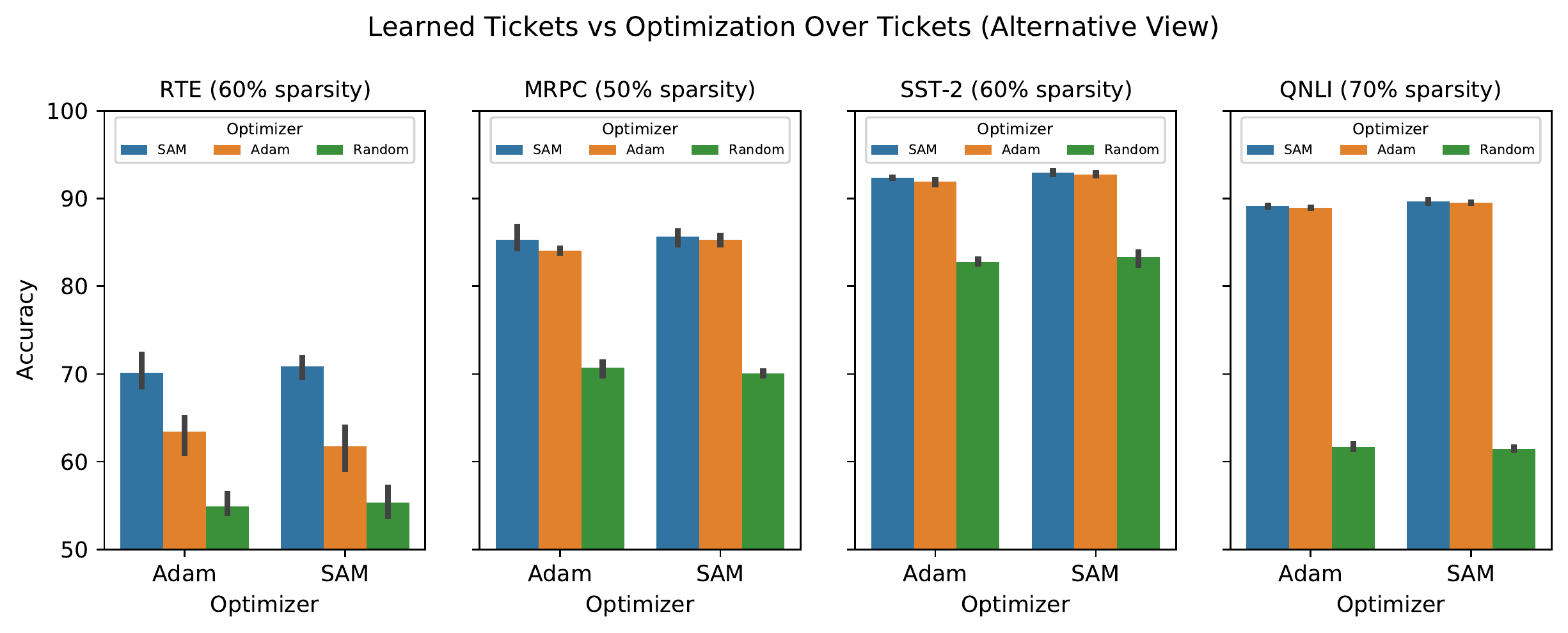}
    \caption{An alternative view of the same information contained in Figure~\ref{fig:OptimizerSwap}. Note that, for example, in RTE, \textcolor{RoyalBlue}{SAM} tickets clearly outperform vanilla \textcolor{YellowOrange}{Adam} tickets regardless of the optimizer used for the final fine-tuning.}
    \label{fig:OptimizerSwapAltView}
\end{figure*}

Figure \ref{fig:OptimizerSwapAltView} presents the same information as Figure \ref{fig:OptimizerSwap} in an alternative view, featuring colored bars representing ticket performance over different optimizers.
\begin{table}[h!]
\small
    \centering
    \begin{tabular}{l l l | r}
    \hline
         Dataset & Ticket & Optim. & Accuracy \\
         \hline
         RTE & \textcolor{Gray}{Random} & \textcolor{YellowOrange}{Adam} & $54.9_{1.3}$ \\
         ($60\%$) &  & \textcolor{RoyalBlue}{SAM} & $55.4_{1.6}$ \\
         & \textcolor{YellowOrange}{Adam} & \textcolor{YellowOrange}{Adam} & $63.7_{0.9}$ \\
         &  & \textcolor{RoyalBlue}{SAM} & $61.7_{2.4}$ \\
         & \textcolor{RoyalBlue}{SAM} & \textcolor{YellowOrange}{Adam} & $70.2_{1.9}$ \\
         &  & \textcolor{RoyalBlue}{SAM} & $71.8_{1.7}$ \\
         \hline
         MRPC & \textcolor{Gray}{Random} & \textcolor{YellowOrange}{Adam} & $70.8_{0.8}$ \\
         ($50\%$) &  & \textcolor{RoyalBlue}{SAM} & $70.1_{0.2}$ \\
         & \textcolor{YellowOrange}{Adam} & \textcolor{YellowOrange}{Adam} & $84.2_{0.1}$ \\
         &  & \textcolor{RoyalBlue}{SAM} & $85.3_{0.5}$ \\
         & \textcolor{RoyalBlue}{SAM} & \textcolor{YellowOrange}{Adam} & $85.3_{1.3}$ \\
         &  & \textcolor{RoyalBlue}{SAM} & $85.7_{0.8}$ \\
         \hline
         SST-2 & \textcolor{Gray}{Random} & \textcolor{YellowOrange}{Adam} & $82.8_{0.2}$ \\
         ($60\%$) &  & \textcolor{RoyalBlue}{SAM} & $83.3_{0.8}$ \\
         & \textcolor{YellowOrange}{Adam} & \textcolor{YellowOrange}{Adam} & $91.9_{0.3}$ \\
         &  & \textcolor{RoyalBlue}{SAM} & $92.7_{0.1}$ \\
         & \textcolor{RoyalBlue}{SAM} & \textcolor{YellowOrange}{Adam} & $92.4_{0.1}$ \\
         &  & \textcolor{RoyalBlue}{SAM} & $92.9_{0.2}$ \\
        \hline
         QNLI & \textcolor{Gray}{Random} & \textcolor{YellowOrange}{Adam} & $61.7_{0.3}$ \\
         ($70\%$) &  & \textcolor{RoyalBlue}{SAM} & $61.5_{0.1}$ \\
         & \textcolor{YellowOrange}{Adam} & \textcolor{YellowOrange}{Adam} & $89.0_{0.05}$ \\
         &  & \textcolor{RoyalBlue}{SAM} & $89.5_{0.04}$ \\
         & \textcolor{RoyalBlue}{SAM} & \textcolor{YellowOrange}{Adam} & $89.1_{0.1}$ \\
         &  & \textcolor{RoyalBlue}{SAM} & $89.6_{0.2}$ \\
    \hline
    \end{tabular}
    \caption{For RTE, MRPC, SST-2, and QNLI at their reference sparsity values, we fine-tune using 1) \textcolor{RoyalBlue}{SAM} and 2) \textcolor{YellowOrange}{Adam} optimizers from pre-trained BERT-base initializations using only the remaining weights based on a) a \textcolor{Gray}{Random} mask, b) an \textcolor{YellowOrange}{Adam}-learned mask, and c) a \textcolor{RoyalBlue}{SAM}-learned mask.}
    \label{tab:OptimizerSwapAnalysis}
\end{table}

\subsection{Ticket Transfer Experiments: Comparing Optimization Over Given Tickets}
\label{sec:transfer_appendix}

\begin{figure*}[h]
\centering
    \begin{subfigure}{.495\textwidth}
      \centering
      \includegraphics[width=\textwidth]{"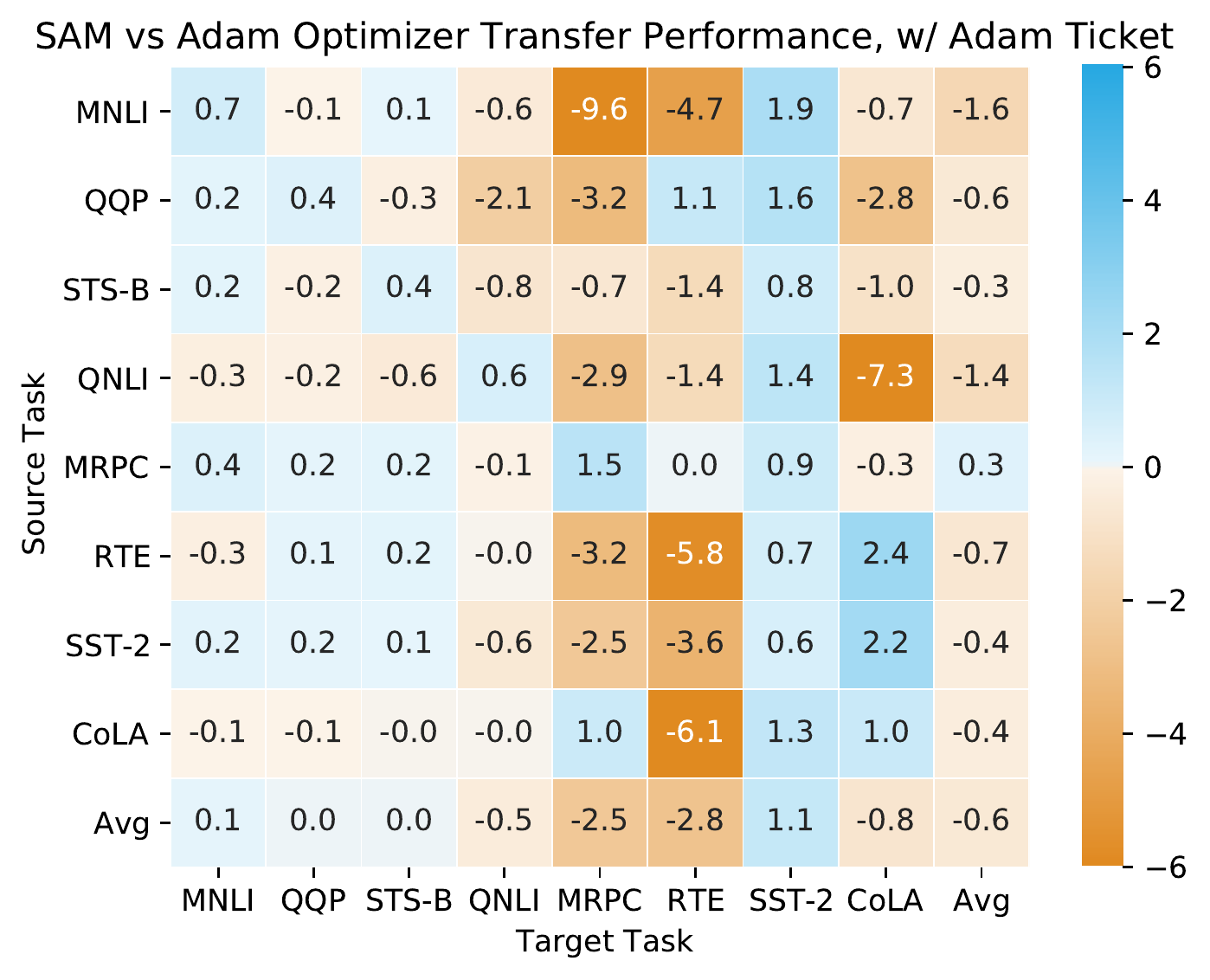"}
    \end{subfigure}\hspace{\fill}%
    \begin{subfigure}{.495\textwidth}
      \centering
      \includegraphics[width=\textwidth]{"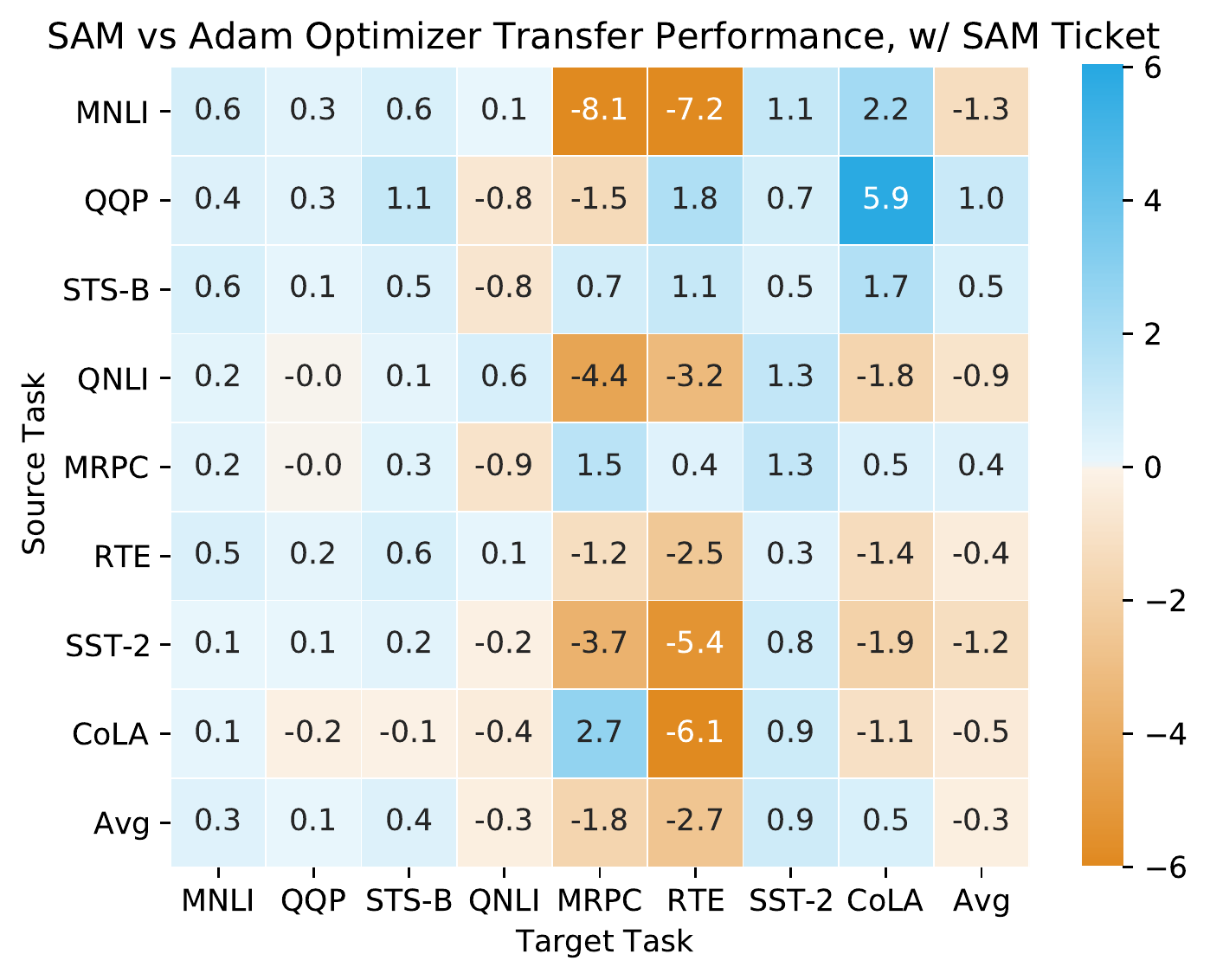"}
    \end{subfigure}\hspace{\fill}%
\caption{Heatmaps indicating the \textit{difference} in target task performance between SAM and Adam optimizers during fine-tuning when transferring tickets across tasks. Values greater than 0 indicate the extent to which \textcolor{RoyalBlue}{SAM optimizer worked better} than Adam; Values less than 0 indicate where \textcolor{YellowOrange}{Adam optimizer worked better} than SAM; Values close to 0 indicate \textcolor{Gray}{little difference} Note that positive values along the diagonal indicate superior SAM ticket performance in the single task setting, even with ``transfer'' between optimizers. \label{fig:optimizer-transfer-heatmaps}}
\end{figure*}

Figure \ref{fig:transfer-heatmaps} in \S\ref{sec:transfer_expts} allows us to evaluate SAM vs. vanilla Adam \textit{ticket} transferability across GLUE tasks. Figure \ref{fig:optimizer-transfer-heatmaps}, on the other hand, allows us to evaluate SAM vs. vanilla Adam \textit{optimization} over given tickets for different GLUE tasks.

\subsection{Structured Pruning Reproducibility and Hyperparameters}
\label{subsec:cofi_appendix}

Following \citet{xia2022structuredCoFi}, we train for 20 epochs each in the pruning and final fine-tuning stages. We use a sparsity epsilon value of $0.01$, meaning that a model can be accepted if its actual sparsity level is within $1\%$ of the target sparsity level of $95\%$.

We pick hyperparameters based on a grid search over \citet{xia2022structuredCoFi}'s baseline implementation (with their Adam-optimized teacher models), reported in Table~\ref{tab:CoFiRefHyperparams}. 

For each task, we use the optimal $\lambda$ and final fine-tuning learning rates found via grid search using \citet{xia2022structuredCoFi}'s implementation, including configurations for finetuning teacher models (which used normal Adam optimizers). 
There are small discrepancies between the reference metric values reported and the values we were able to reproduce, possibly due to variation across random seeds. However, the relative performance of hyperparameter settings seems to be fairly consistent across random seeds.

\begin{table}[h!]
\small
    \centering
    \begin{tabular}{l r | l | l | m{0.06\textwidth} | m{0.05\textwidth}}
    \hline
         Dataset & & $\lambda$ & FT-LR & Teacher Acc. & Pruned Acc. \\
         \hline
         SST-2 & \textit{Ref.} & - & - & 93.1 & 90.6 \\
         (67k) & \textit{Reprod.} & $0.9$ & $3\mathrm{e}{-5}$ & 93.6 & 90.6 \\
    \hline
         QNLI & \textit{Ref.} & - & - & 91.5 & 86.1 \\
         (105k) & \textit{Reprod.} & $0.9$ & $3\mathrm{e}{-5}$ & 91.9 & 86.5 \\
    \hline
         QQP & \textit{Ref.} & - & - & 91.2 & 90.1 \\
         (364k) & \textit{Reprod.} & $0.7$ & $3\mathrm{e}{-5}$ & 91.3 & 89.9\\
    \hline
         MNLI & \textit{Ref.} & - & - & 84.8 & 80.6 \\
         (393k) & \textit{Reprod.} & $0.7$ & $3\mathrm{e}{-5}$ & 85.2 & 80.1 \\
    \hline
    \end{tabular}
    \caption{We report reproduced (\textit{Reprod.}) and reference (\textit{Ref.}) evaluation metrics at $95\%$ sparsity and optimal values for $\lambda$ and fine-tuning learning rate on select tasks from \citet{xia2022structuredCoFi}'s structured pruning setting. We used the same hyperparameters as reported otherwise (distillation temperature $t = 2$, with 20 finetuning epochs after pruning, learning rate $=2e-5$, and batch size$=32$), and conducted our grid search over the same candidate values $\lambda \in \{0.1, 0.3, 0.5\}$ and FT-LR $\in \{1e-5, 2e-5, 3e-5\}$} %
    \label{tab:CoFiRefHyperparams}
\end{table}

Although we do not conduct an additional full grid search for our comparison of compressed models using Adam and SAM-optimized teacher models, we do find that the optimal final fine-tuning learning rates, which are much less computationally expensive to test, transfer to our experimental settings. 

\subsection{Detailed Quantization Results}
\label{subsec:more_quant}
Table~\ref{tab:ptdq_table} contains the numbers used to generate Figure~\ref{fig:ptdq_quant}.
\begin{table}[h!]
\small
    \centering
    \begin{tabular}{l r | r | r r }
    \hline
        Dataset & QAT Ref. & Optim. & Full FT & Quantized \\
        \hline
        MNLI & \multirow{2}{*}{N/A} & Adam & $84.42_{0.37}$ & $78.24_{4.00}$ \\
        Acc. & & SAM & $84.68_{0.13}$ & $83.47_{0.11}$ \\
        \hline
        QQP & \multirow{2}{*}{$87.96$} & Adam & $87.98_{0.12}$ & $85.35_{1.93}$ \\
        F1 & & SAM & $88.20_{0.08}$ & $86.85_{0.28}$ \\
        \hline
        STS-B & \multirow{2}{*}{$89.04$} & Adam & $89.06_{0.11}$ & $86.54_{1.07}$ \\
        Pearson & & SAM & $89.39_{0.07}$ & $87.16_{0.76}$ \\
        \hline
        QNLI & \multirow{2}{*}{$90.62$} & Adam & $91.49_{0.09}$ & $89.05_{0.37}$ \\
        Acc. & & SAM & $91.33_{0.58}$ & $89.83_{0.54}$\\
        \hline
        MRPC & \multirow{2}{*}{$89.56$} & Adam & $89.57_{1.13}$ & $86.81_{1.72}$\\
        F1 & & SAM & $91.24_{0.20}$ & $\mathbf{89.37_{0.83}}$\\
        \hline
        RTE & \multirow{2}{*}{$68.78$} & Adam & $67.63_{1.10}$ & $56.80_{4.51}$ \\
        Acc. & & SAM & $67.87_{0.36}$ & $65.70_{1.65}$ \\
        \hline
        SST-2 & \multirow{2}{*}{$92.24$} & Adam & $92.70_{0.07}$ & $91.17_{0.90}$ \\
        Acc. & & SAM & $93.08_{0.57}$ & $\mathbf{92.39_{0.40}}$ \\
        \hline
        CoLA & \multirow{2}{*}{$58.48$} & Adam & $60.47_{0.55}$ & $55.99_{2.86}$ \\
        Matt. & & SAM & $59.09_{0.72}$ & $54.88_{0.78}$ \\
        \hline
        SQuAD & \multirow{2}{*}{$87.74$} & Adam & $89.20_{0.12}$ & $80.13_{1.85}$ \\
        F1 & & SAM & $89.20_{0.12}$ & $84.92_{0.55}$ \\
    \hline
    \end{tabular}
    \caption{We compare full fine-tuned and quantized BERT$_{base}$ models optimized with SAM and Adam. Notably, applying a simpler post-training dynamic quantization technique on a SAM-optimized model can approach the reported (QAT ref) performance of a model quantized through quantization-aware training \cite{Zafrir2019Q8BERTQ8}. These instances are \textbf{bolded}.}
    \label{tab:ptdq_table}
\end{table}

\subsection{Results for Other BERT Models}
\label{subsec:other_models}
We investigate SAM's influence on amenability to sparsification in both BERT$_{large}$ and RoBERTa$_{base}$ models subject to iterative magnitude pruning (IMP). For consistency, we use the same hyperparameters as for the BERT$_{base}$ set of experiments ($\epsilon=0$ weight decay; $10$ (MRPC, RTE), $3$ (SST-2, QNLI), or $2$ (SQuAD) training epochs for each IMP iteration; linear learning rate decay schedules starting at $2e-5$ (GLUE) or $3e-5$ (SQuAD); batch size of $32$ (GLUE) and $16$ (SQuAD); maximum sequence length of $128$ (GLUE) and $384$ (SQuAD)). It is possible that a different set of hyperparameters would be optimal for these different models, but we also tried different numbers of training epochs for the less stable smaller GLUE tasks ($5$ and $3$), as well as \citet{Liu2019RoBERTaAR}'s learning rate of $1.5e-5$ for RoBERTa, and we generally found that simply matching hyperparameters from BERT$_{base}$ experiments worked well or better.

\begin{figure*}[h]
    \centering
    \begin{subfigure}{.495\textwidth}
      \centering
      \includegraphics[width=\textwidth]{"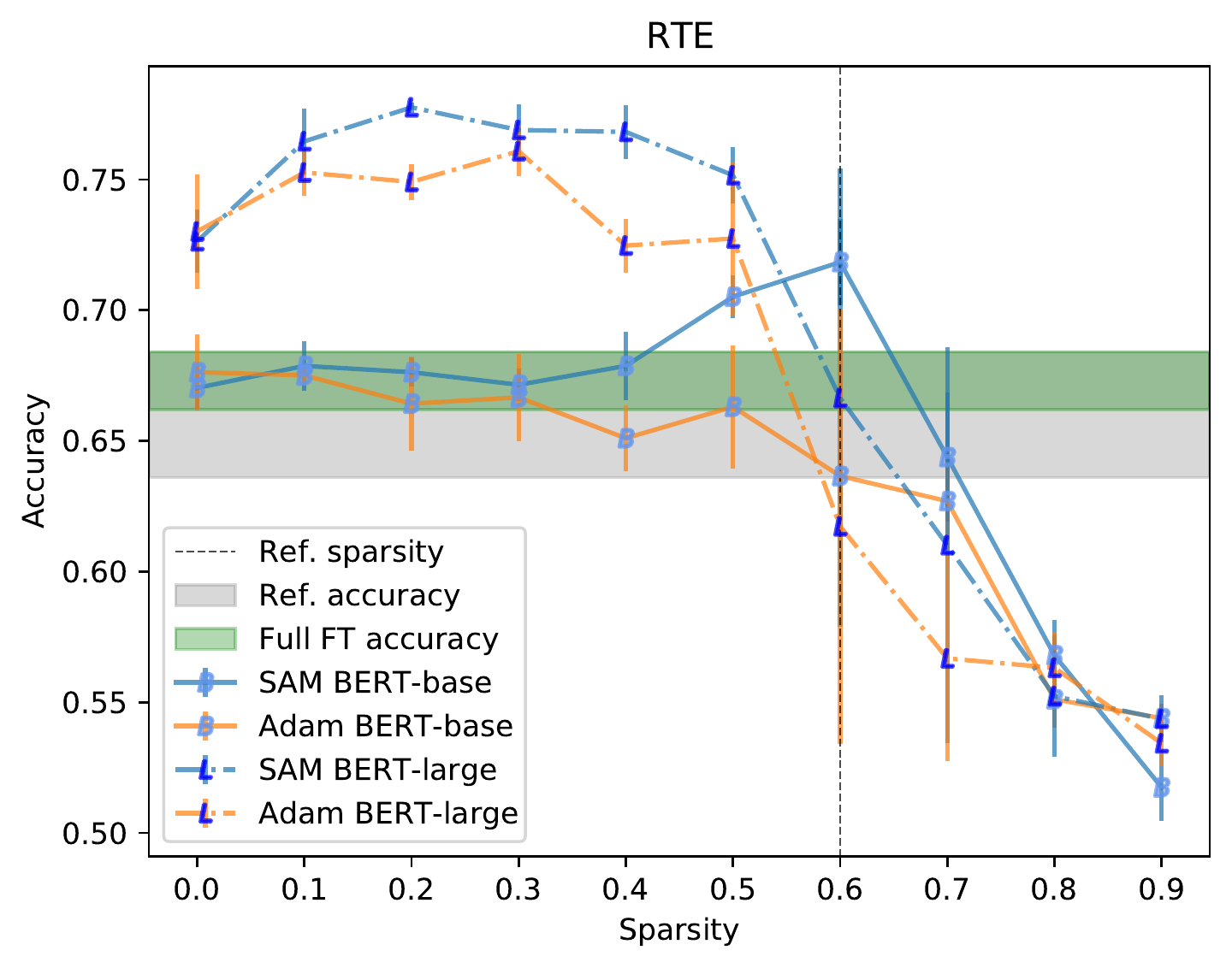"}
      \subcaption{RTE, IMP w/ BERT$_{large}$}
      \label{fig:rte_large}
    \end{subfigure}\hspace{\fill}%
    \begin{subfigure}{.495\textwidth}
      \centering
      \includegraphics[width=\textwidth]{"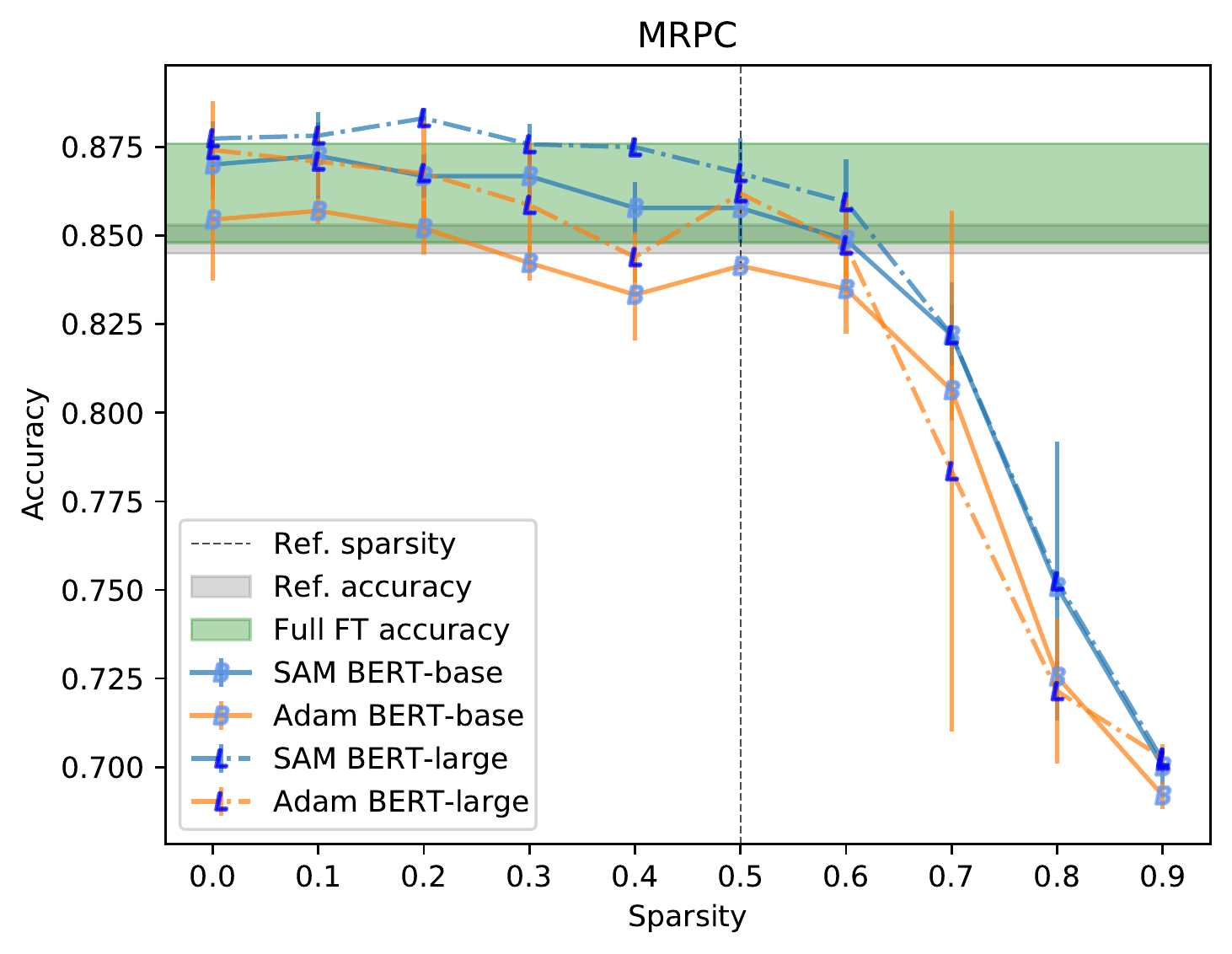"}
      \caption{MRPC, IMP w/ BERT$_{large}$}
      \label{fig:mrpc_large}
    \end{subfigure}\hspace{\fill}%
    \begin{subfigure}{.495\textwidth}
      \centering
      \includegraphics[width=\textwidth]{"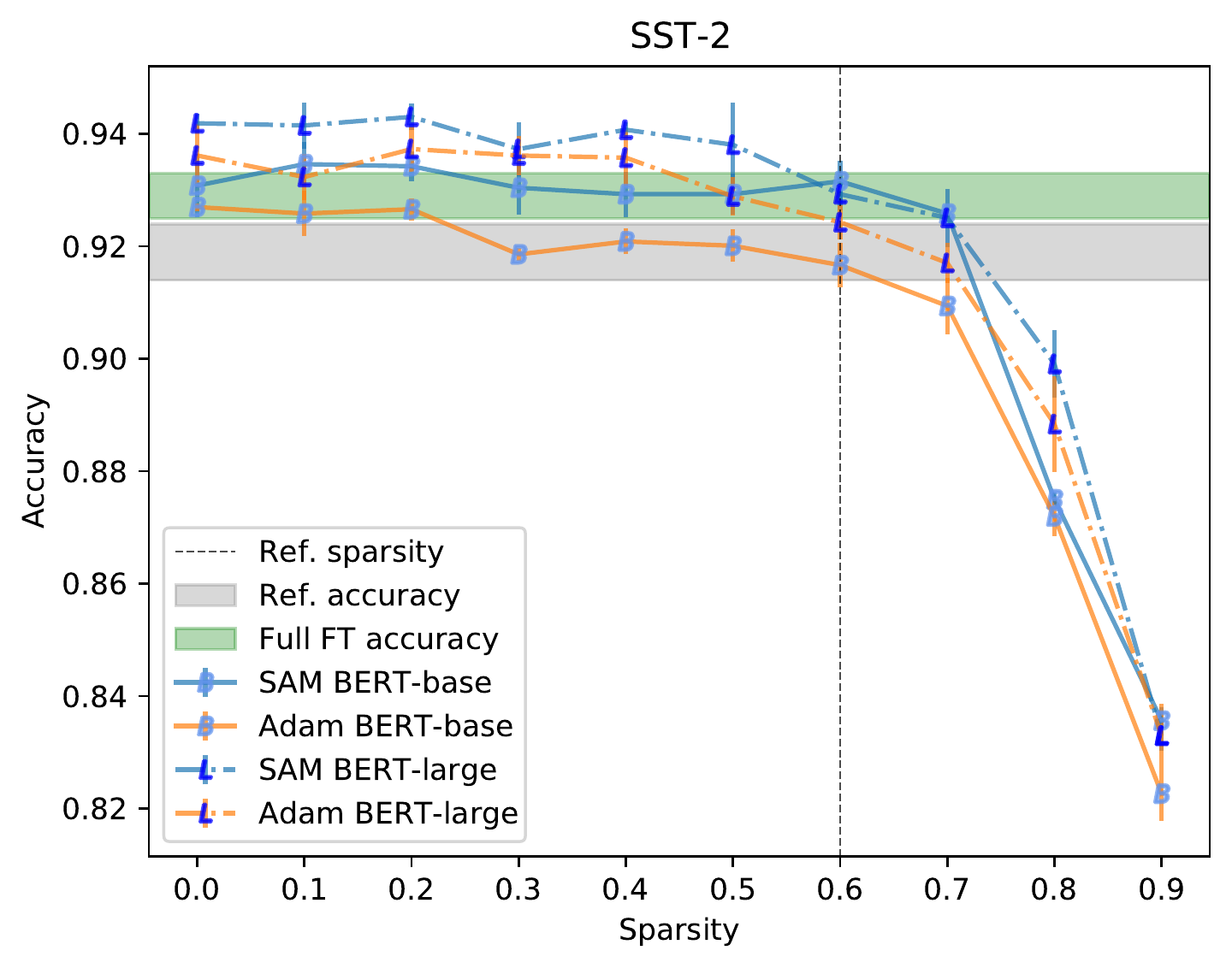"}
      \subcaption{SST-2, IMP w/ BERT$_{large}$}
      \label{fig:sst2_large}
    \end{subfigure}\hspace{\fill}%
    \begin{subfigure}{.495\textwidth}
      \centering
      \includegraphics[width=\textwidth]{"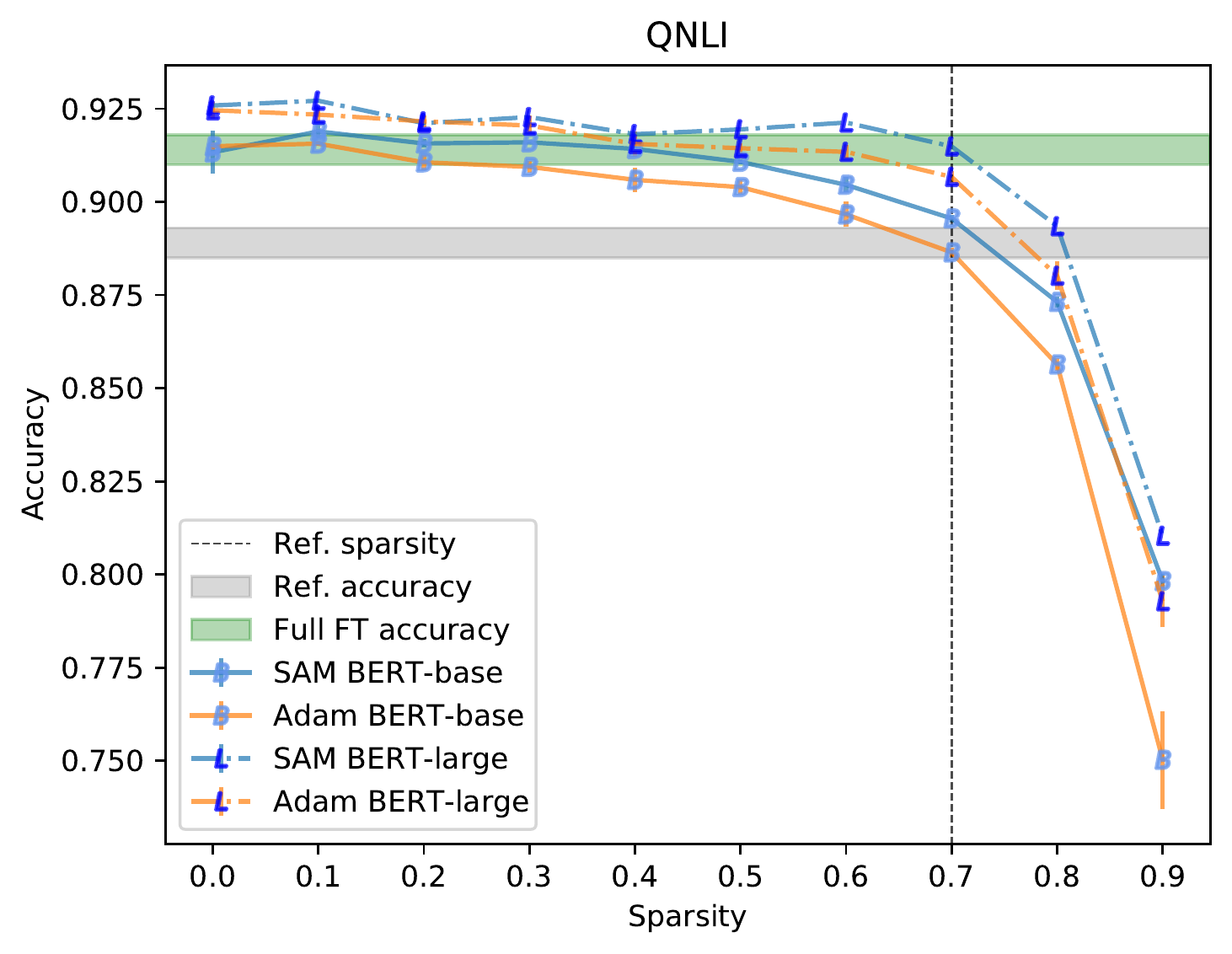"}
      \caption{QNLI, IMP w/ BERT$_{large}$}
      \label{fig:qnli_large}
    \end{subfigure}\hspace{\fill}%
    \begin{subfigure}{.495\textwidth}
      \centering
      \includegraphics[width=\textwidth]{"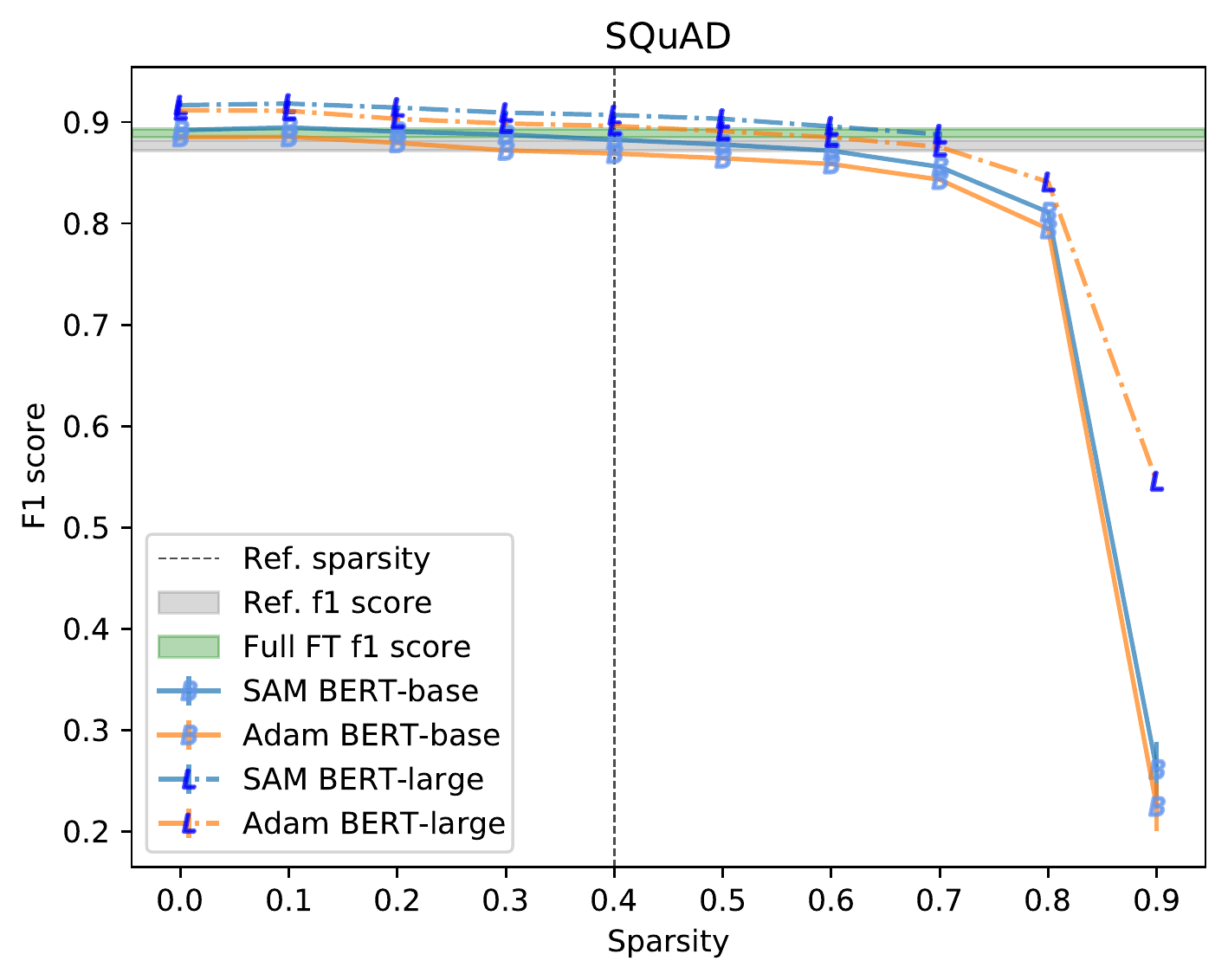"}
      \caption{SQuAD, IMP w/ BERT$_{large}$}
      \label{fig:squad_large}
    \end{subfigure}\hspace{\fill}
    \caption{Individual plots showing sparsity vs. accuracy for GLUE tasks and SQuAD in BERT$_{large}$ models compressed with iterative magnitude pruning (IMP), with BERT$_{base}$ models for comparison. The vertical lines and gray horizontal bands mark reference sparsity and "winning ticket" evaluation metric values that were obtained by \citet{chen2020BertLT}. The green horizontal bands mark the initial performance of our full fine-tuned models.}
    \label{fig:task_bertlarge_plots}
\end{figure*}

\begin{figure*}[h]
    \centering
    \begin{subfigure}{.495\textwidth}
      \centering
      \includegraphics[width=\textwidth]{"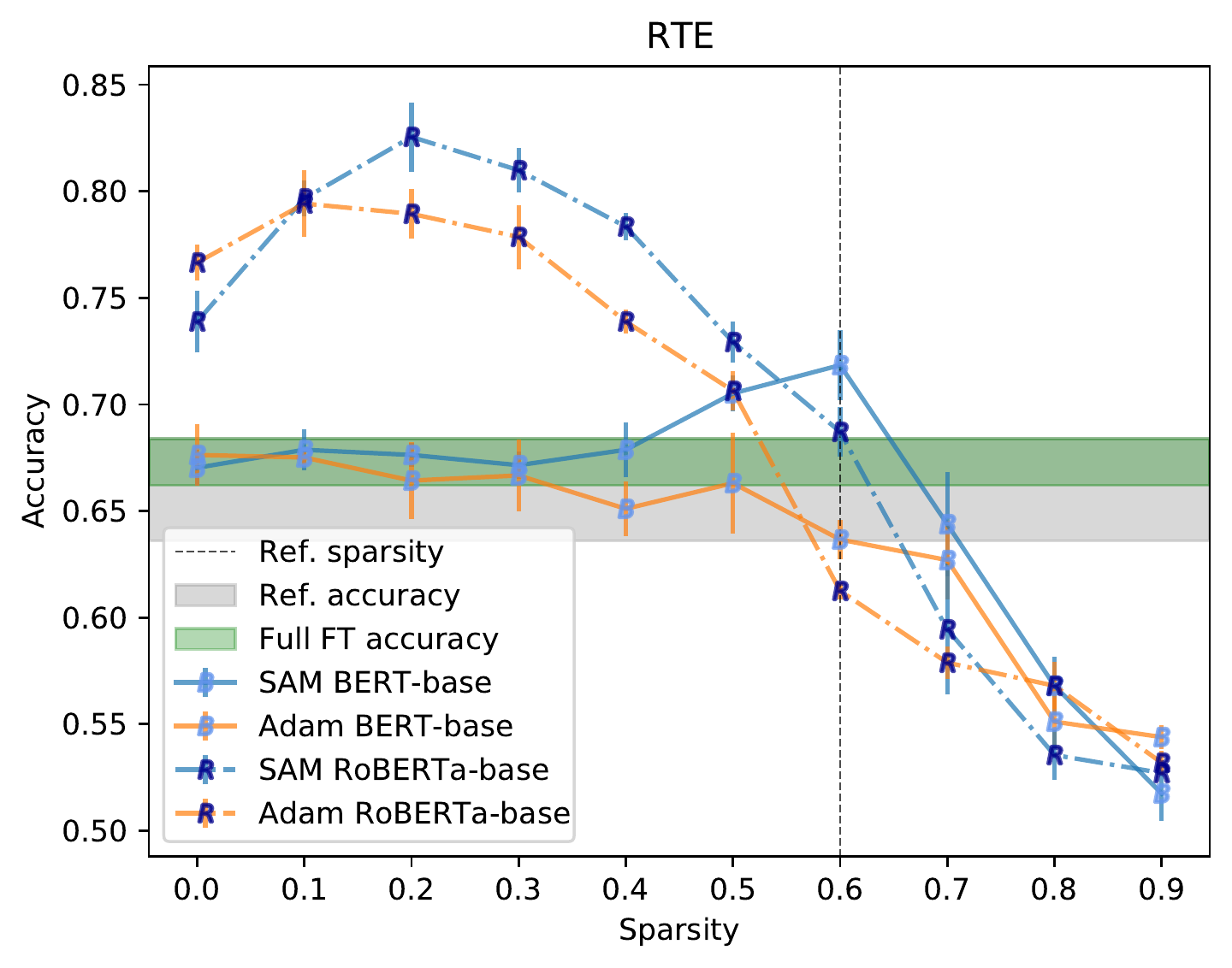"}
      \subcaption{RTE, IMP w/ RoBERTa$_{base}$}
      \label{fig:rte_roberta}
    \end{subfigure}\hspace{\fill}%
    \begin{subfigure}{.495\textwidth}
      \centering
      \includegraphics[width=\textwidth]{"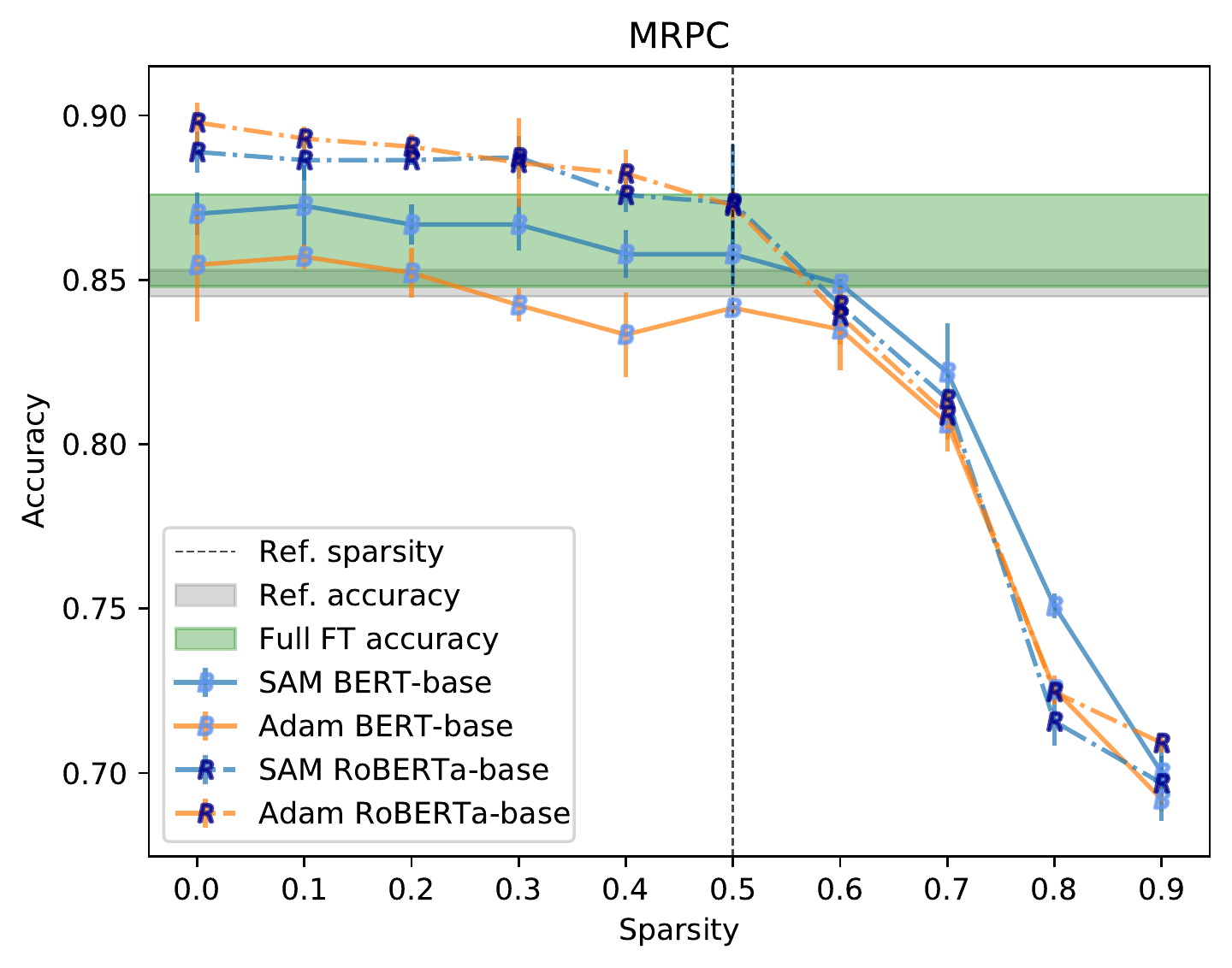"}
      \caption{MRPC, IMP w/ RoBERTa$_{base}$}
      \label{fig:mrpc_roberta}
    \end{subfigure}\hspace{\fill}%
    \begin{subfigure}{.495\textwidth}
      \centering
      \includegraphics[width=\textwidth]{"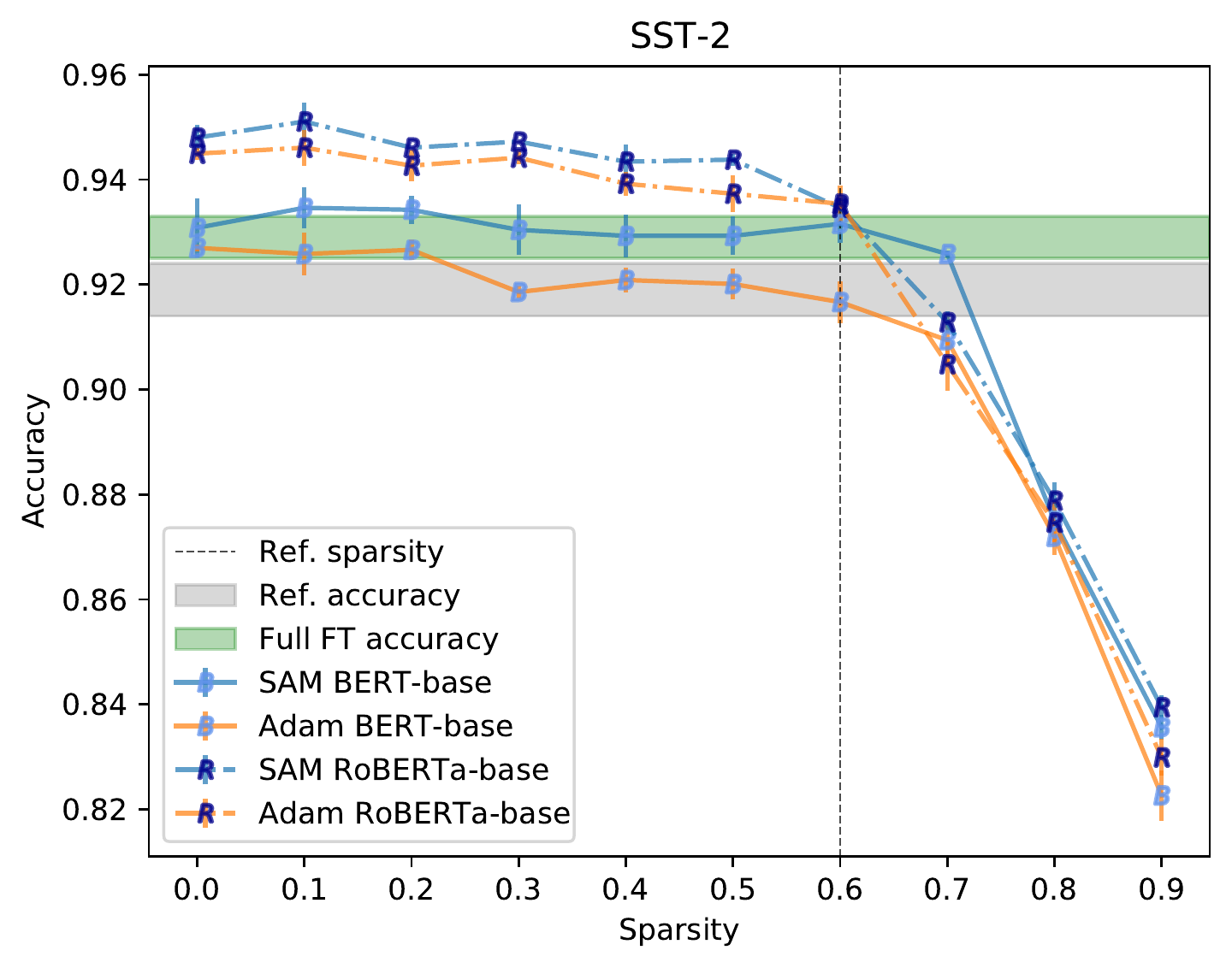"}
      \subcaption{SST-2, IMP w/ RoBERTa$_{base}$}
      \label{fig:sst2_roberta}
    \end{subfigure}\hspace{\fill}%
    \begin{subfigure}{.495\textwidth}
      \centering
      \includegraphics[width=\textwidth]{"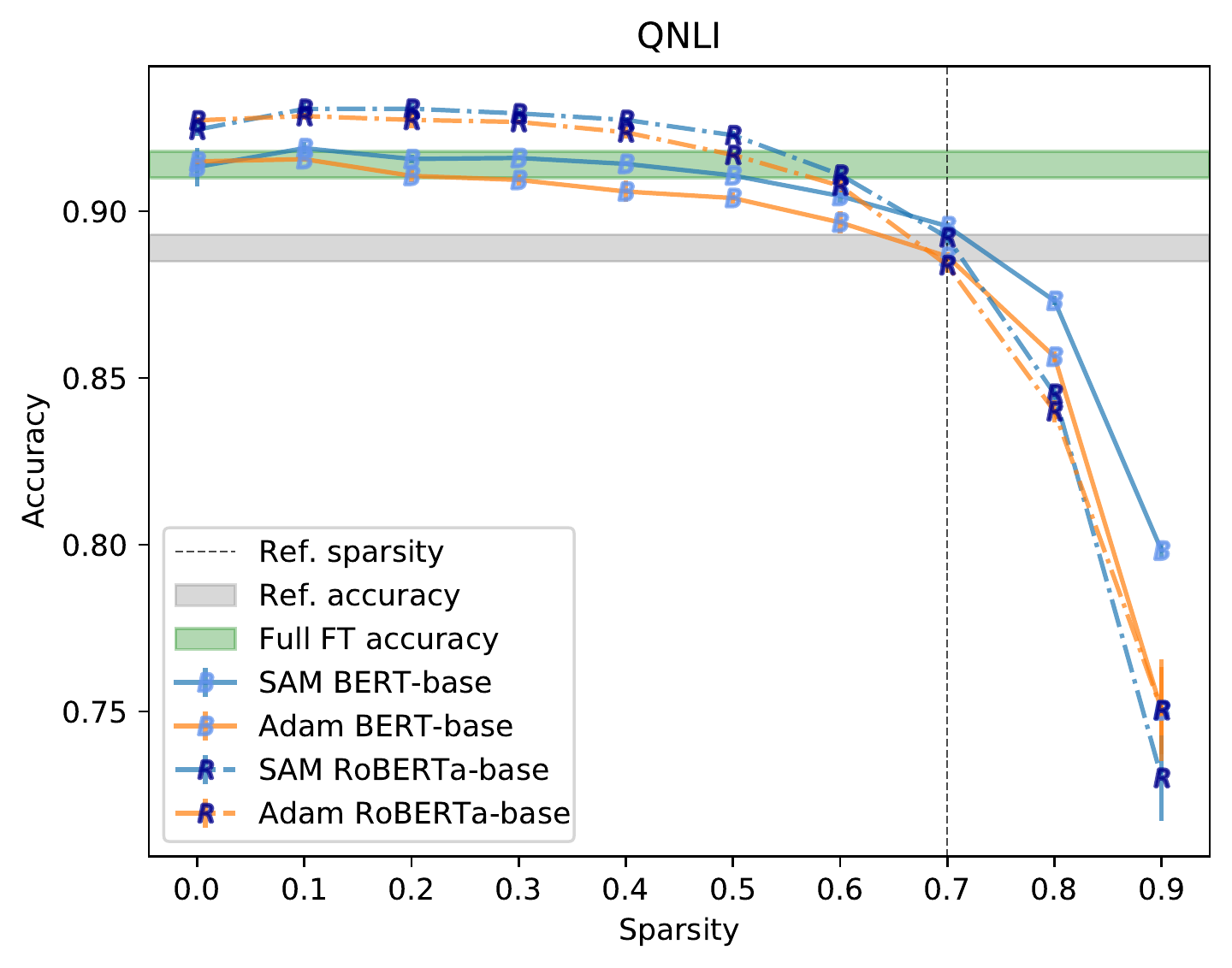"}
      \caption{QNLI, IMP w/ RoBERTa$_{base}$}
      \label{fig:qnli_roberta}
    \end{subfigure}\hspace{\fill}%
    \begin{subfigure}{.495\textwidth}
      \centering
      \includegraphics[width=\textwidth]{"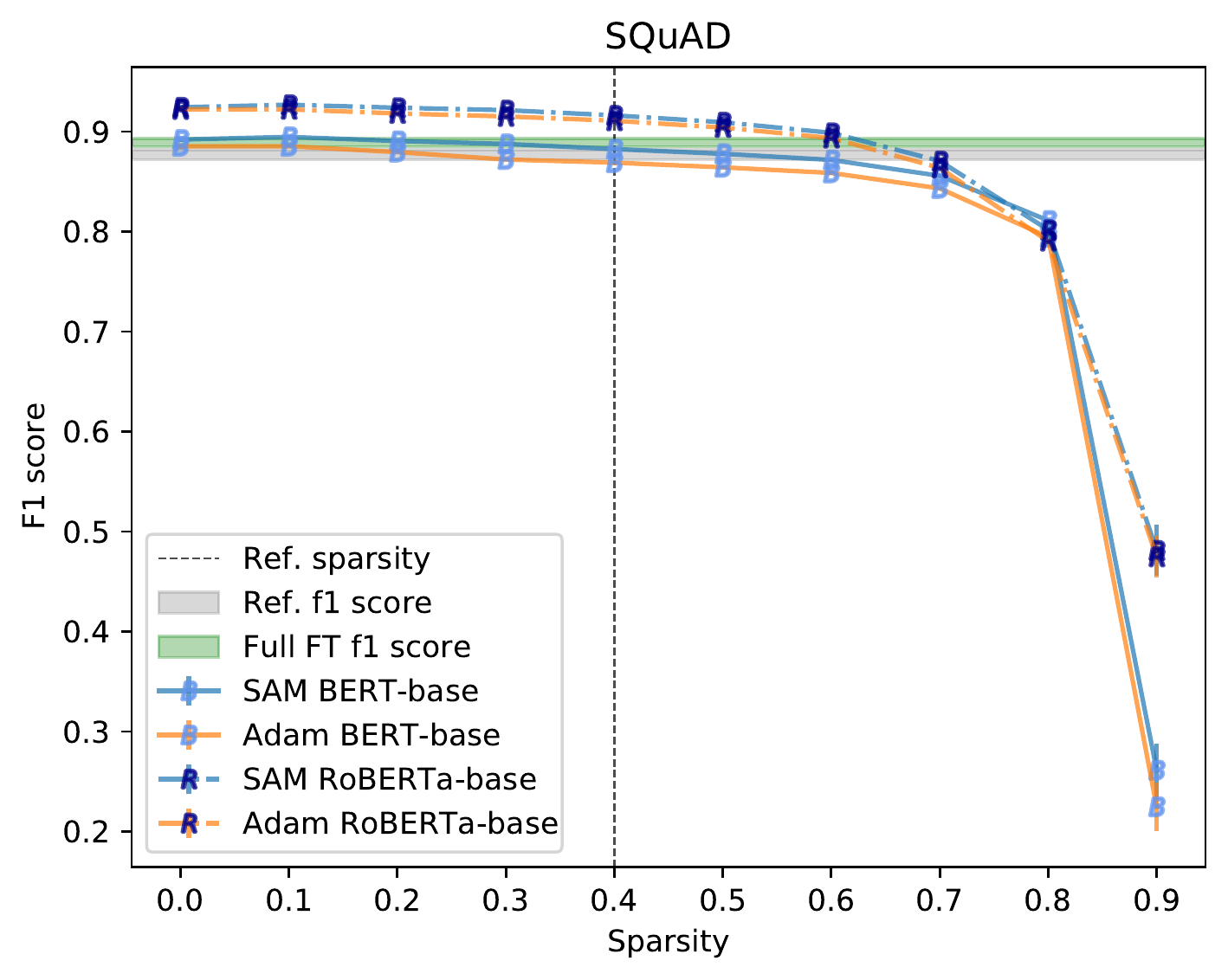"}
      \caption{SQuAD, IMP w/ RoBERTa$_{base}$}
      \label{fig:squad_roberta}
    \end{subfigure}\hspace{\fill}
    \caption{Individual plots showing sparsity vs. accuracy for GLUE tasks and SQuAD in RoBERTa$_{base}$ models compressed with iterative magnitude pruning (IMP), with BERT$_{base}$ models for comparison. The vertical lines and gray horizontal bands mark reference sparsity and "winning ticket" evaluation metric values that were obtained by \citet{chen2020BertLT}. The green horizontal bands mark the initial performance of our full fine-tuned models.}
    \label{fig:task_roberta_plots}
\end{figure*}

Figures~\ref{fig:task_bertlarge_plots} and~\ref{fig:task_roberta_plots} show plots for BERT$_{large}$ and RoBERTa$_{base}$ models compressed with iterative magnitude pruning (IMP). We include our BERT$_{base}$ model results with \citet{chen2020BertLT}'s reference sparsity levels and accuracy ranges (which are likewise for BERT$_{base}$) in the same plots for comparison.

With the exception of RoBERTa$_{base}$ on MRPC, SAM-optimized models consistently fare better than Adam-optimized models in these other BERT variants as well. However, the comparison between BERT variants is more complex. While BERT$_{large}$ and RoBERTa$_{base}$ models generally achieve higher initial performance compared to BERT$_{base}$ and can maintain this higher performance at \citet{chen2020BertLT}'s "winning ticket" sparsity levels (\ref{fig:mrpc_large}, \ref{fig:qnli_large}, \ref{fig:squad_large}, \ref{fig:mrpc_roberta}, \ref{fig:squad_roberta}), the drop-off in performance does not always simply follow a parallel pattern. Initial higher performance tends to decrease more quickly with pruning than in BERT$_{base}$ models, such that BERT$_{large}$ and RoBERTa$_{base}$ performance sometimes falls to near (\ref{fig:sst2_large}, \ref{fig:sst2_roberta}, \ref{fig:qnli_roberta}) or even \textit{below} (\ref{fig:rte_large}, \ref{fig:rte_roberta}) BERT$_{base}$ performance by the time they approach "winning ticket" sparsity levels (which in reality provide an inherent advantage to the larger models that are left with a greater absolute number of parameters at the same sparsity levels).

\subsection{Beyond task accuracy}
\label{sec:checklist_expts}
\begin{figure*}[h]
    \begin{subfigure}{.495\textwidth}
      \centering
      \includegraphics[width=\textwidth]{"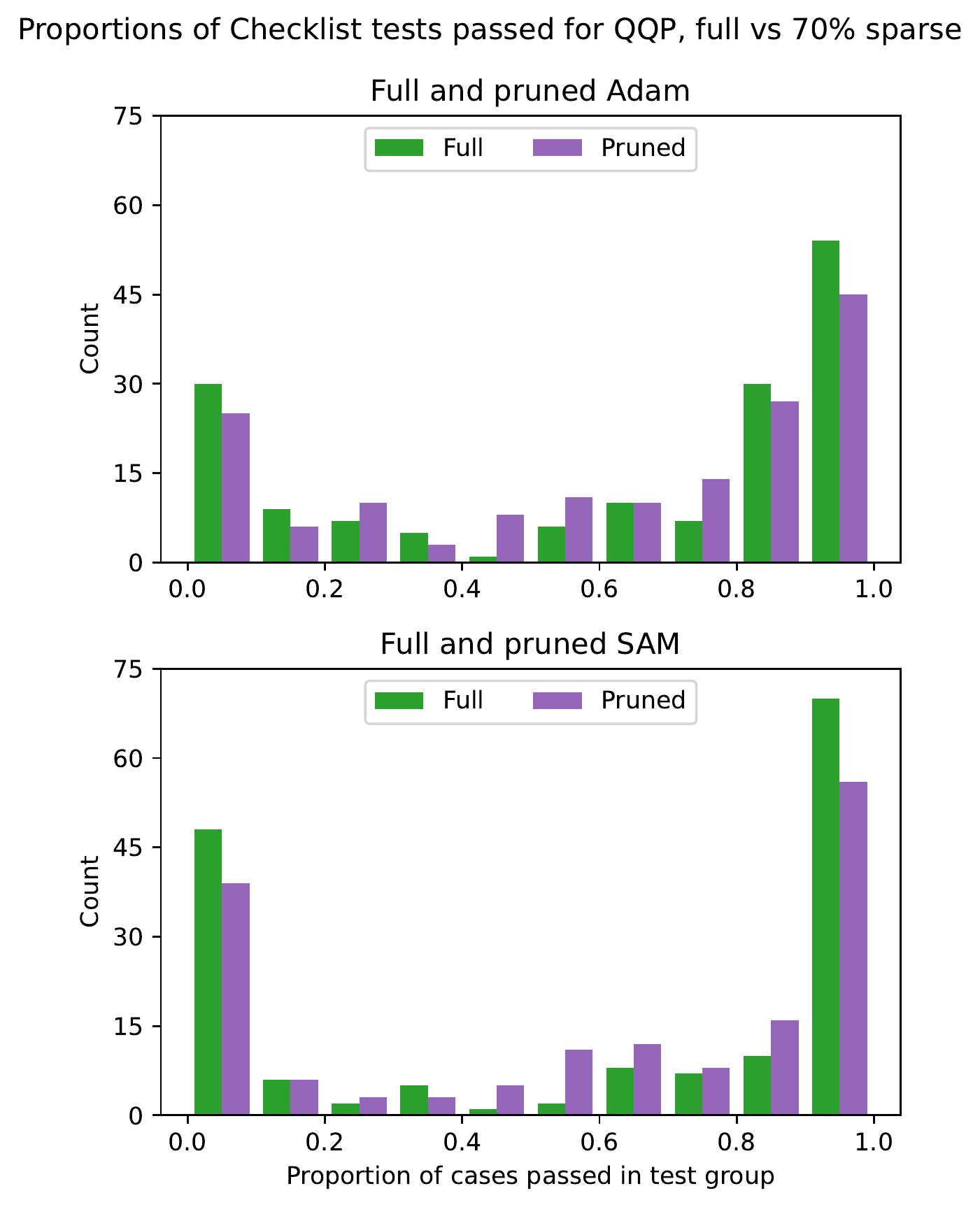"}
    \end{subfigure}\hspace{\fill}%
    \begin{subfigure}{.47\textwidth}
      \centering
      \includegraphics[width=\textwidth]{"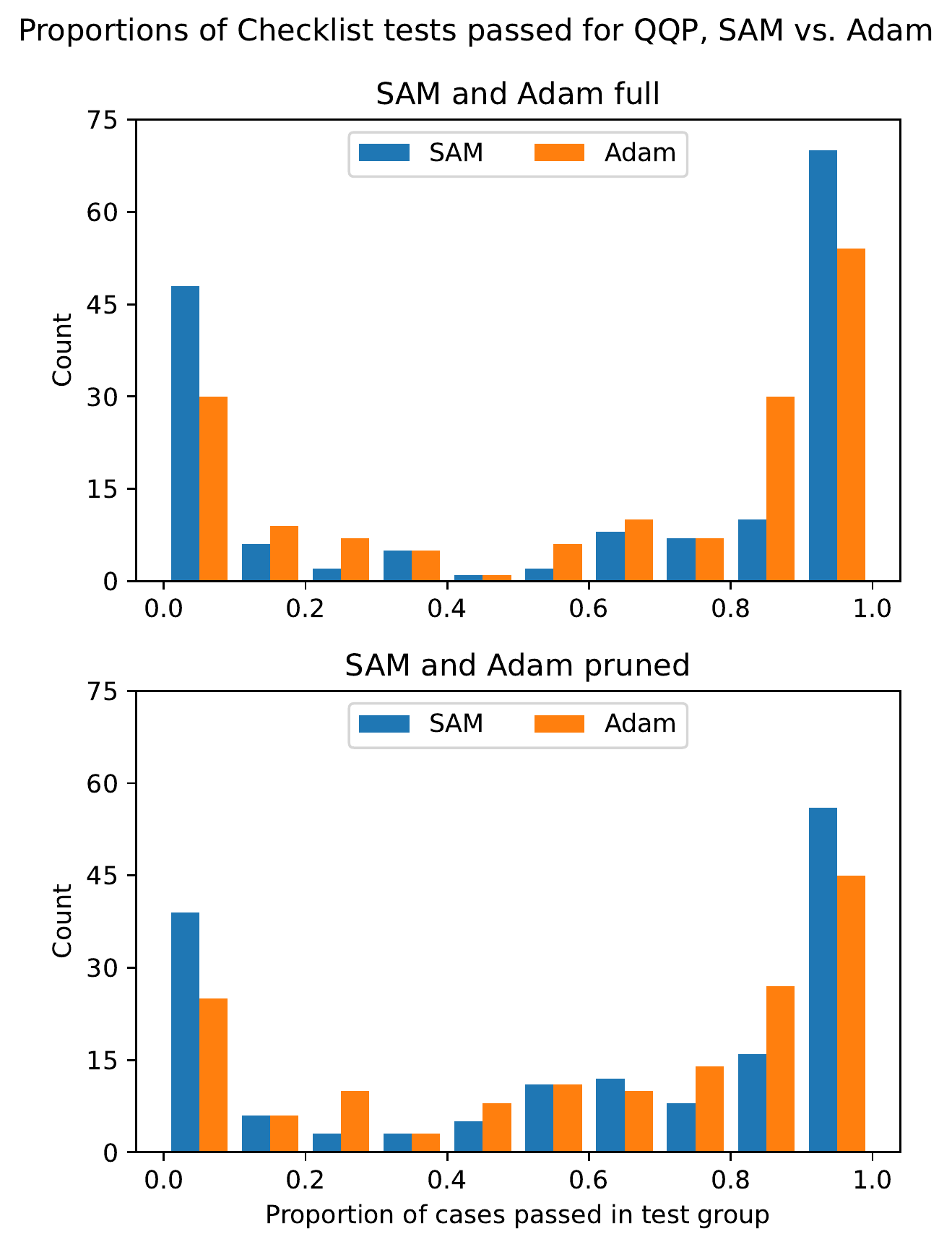"}
    \end{subfigure}\hspace{\fill}%
    \caption{Aggregated Checklist \cite{ribeiro-etal-2020-beyond} results for QQP models. Tests target various capabilities of models such as robustness to typos and simple coference resolution. Note the concentration of frequencies near $0.0$ and $1.0$ for all models, as well as the shifts in frequencies when pruned for both vanilla Adam and SAM models; models can effectively \textit{lose} or even \textit{gain} specific capabilities throughout compression.}
    \label{fig:qqp_checklist_agg}
\end{figure*}

We evaluate full and pruned BERT$_{base}$ models optimized by vanilla Adam and SAM throughout IMP (with rewind) on \citet{ribeiro-etal-2020-beyond}'s pre-curated test suites for sentiment analysis (SST-2), question paraphrase detection (QQP), and question answering (SQuAD). At this time, we do not explicitly make direct comparisons between SAM and vanilla Adam for unpruned and pruned models. %
A single test consists of multiple examples, and the $x$-axes of the histograms in Figure \ref{fig:qqp_checklist_agg} refer to the proportions of examples passed within each test.

\end{document}